\PassOptionsToPackage{most,breakable}{tcolorbox}
\PassOptionsToPackage{dvipsnames,table}{xcolor}
\PassOptionsToPackage{numbers}{natbib}
\documentclass[11pt]{article}

\usepackage[final]{arxiv}

\usepackage[T1]{fontenc}
\usepackage[utf8]{inputenc}
\usepackage{amsmath,amssymb}
\usepackage{mathpazo}
\usepackage[xcharter,bigdelims,vvarbb]{newtxmath}
\usepackage[scaled]{helvet}
\usepackage[scaled=1.1]{zlmtt}
\usepackage{placeins}
\usepackage{float}
\usepackage{booktabs}
\usepackage{multirow}
\usepackage{tabularx}
\usepackage{subcaption}
\usepackage{enumitem}
\usepackage{wrapfig}
\usepackage{microtype}

\usepackage{graphicx}
\usepackage{xspace}

\setlist[itemize]{leftmargin=2em}

\usepackage{tikz}

\newcommand{\benchmark}{FireWorldBench\xspace}

\definecolor{subcol}{RGB}{255,255,255}
\definecolor{bestred}{RGB}{120,170,220}
\definecolor{secondorange}{RGB}{175,205,235}
\definecolor{thirdyellow}{RGB}{215,232,248}
\definecolor{trainmark}{HTML}{B8323E}
\definecolor{oomred}{RGB}{252,228,228}
\definecolor{catgray}{RGB}{230,230,230}
\definecolor{altcolblue}{RGB}{225,237,250}
\definecolor{pretrainmark}{RGB}{197,90,17}
\definecolor{gaingreen}{HTML}{B8323E}
\definecolor{lossred}{RGB}{200,30,30}
\definecolor{navyblue}{HTML}{0071BC}
\definecolor{takeawaybg}{RGB}{239,246,255}
\definecolor{takeawayborder}{RGB}{120,150,200}

\newtcolorbox{takeawaybox}{
  colback=takeawaybg,
  colframe=takeawayborder,
  boxrule=0.4pt,
  arc=1pt,
  left=6pt,
  right=6pt,
  top=5pt,
  bottom=5pt,
  before skip=6pt,
  after skip=6pt
}

\firstpagelogos{}{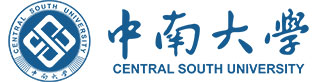}{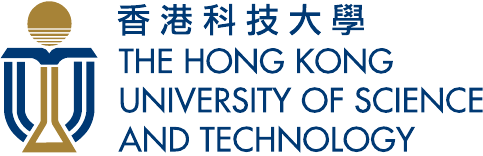}
\abstractlogo{}

\title{\LARGE\mbox{\benchmark: Benchmarking Complex Physical World}\\Intelligence through Coupled-Field Fire Dynamics}

\newcommand{\coremark}{\ensuremath{^{\star}}}
\newcommand{\leadmark}{\ensuremath{^{\dagger}}}
\newcommand{\corrmark}{\ensuremath{^{\ddagger}}}

\newcommand{\authorfootnotes}{%
  \begingroup
  \renewcommand{\thefootnote}{}%
  \footnotetext{$\star$ Core contributors. $\dagger$ Leading project. $\ddagger$ Corresponding authors.}%
  \endgroup
  \addtocounter{footnote}{-1}%
}

\author{
  \AuthorName{Qiang Chen}{1\coremark\leadmark}\quad
  \AuthorName{Hao Guo}{2\coremark}\quad
  \AuthorName{Huatai Zhu}{2\coremark}\quad
  \AuthorName{Tairan Huang}{2\coremark}\quad 
  \AuthorName{Yichao Cao}{2}\quad
  \AuthorName{Hongyan Xu}{2}\quad \\
  \AuthorName{Keke Huang}{2}\quad
  \AuthorName{Haifeng Li}{2}\quad
  \AuthorName{Yi Chen}{1\corrmark}\quad
  \AuthorName{Xiu Su}{2\corrmark}\quad \\
  {\normalfont\footnotesize
  \mbox{\textsuperscript{1}Hong Kong University of Science and Technology}\quad
  \mbox{\textsuperscript{2}Central South University}\quad
}
}

\begin{document}

\maketitle
\authorfootnotes

\begin{paperresources}
\begin{tabularx}{\linewidth}{@{}X@{}}

\paperresourceicon{\resourceprojecticon}
{Project Page}
{https://FireWorldBench.github.io/}
{\nolinkurl{https://FireWorldBench.github.io/}}
\\

\paperresourceicon{\resourcehficon}
{Dataset}
{https://huggingface.co/datasets/FireWorldLab/FireWorldBench}
{\nolinkurl{https://huggingface.co/datasets/FireWorldLab/FireWorldBench}}
\\

\paperresourceicon{\resourcegithubicon}
{Code}
{https://github.com/FireWorldLab/FireWorldBench}
{\nolinkurl{https://github.com/FireWorldLab/FireWorldBench}}
\\

\end{tabularx}
\end{paperresources}

\vspace{1pt}
{\centering
\includegraphics[width=\textwidth]{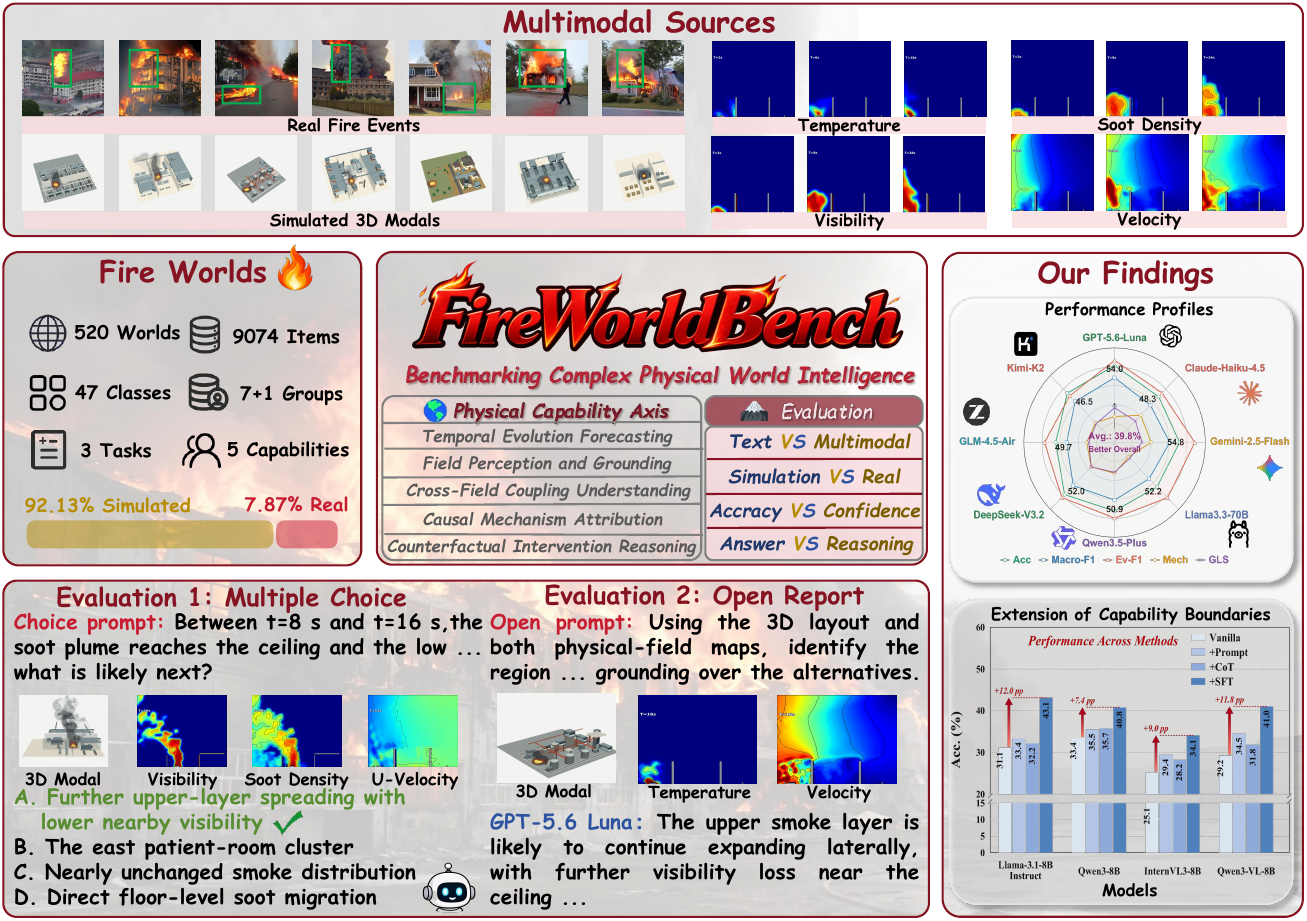}
\par\vspace{3pt}
\begin{minipage}{\textwidth}
\footnotesize
\refstepcounter{figure}\label{fig:teaser}
Figure 1 | \textbf{\benchmark} provides a reproducible, multimodal benchmark spanning 520 coupled-field fire worlds, 47 scene archetypes across 7 environment families, and 9,074 question-answer instances from controlled simulation and real-world-aligned settings, evaluated along complementary physical-capability and fire-scenario task axes. Our analysis reveals current capability boundaries in coupled-field understanding, physical explanation, and intervention reasoning, complemented by FireWorldGPT, which demonstrates that domain adaptation with multi-physics field visualizations can substantially strengthen physical-world understanding.
\end{minipage}
\par}

\vspace{0pt}
\begin{abstract}
\brokenpenalty=10000\relax
Understanding the physical world requires more than recognizing objects, describing scenes, or predicting short-term visual changes. Real-world physical systems often involve multiple continuous fields, latent causal mechanisms, partial observations, and intervention-sensitive dynamics. 
In this work, we propose FireWorldBench, a benchmark for evaluating complex physical world intelligence in multimodal large language models through coupled-field fire dynamics. Fire provides a canonical stress-test environment for complex physical world intelligence, as temperature, soot density, visibility, velocity, carbon-monoxide concentration, combustion and ventilation fields jointly shape observable states and temporal dynamics. 
We organize the benchmark along two complementary axes: a physical capability axis, \textit{progressing from Temporal Evolution Forecasting, Physical Field Perception and Grounding, Cross-Field Coupling Understanding, Causal Mechanism Attribution, to Counterfactual Intervention Reasoning}; and a fire scenario task axis, progressing from Localized Onset, Coupled Propagation, to Critical Transition. 
The benchmark dataset comprises 520 fire-world entries: 494 controlled simulation worlds and 26 real-world-aligned event groups, spanning 47 scene archetypes in 7 environment families. Each entry is represented through structured textual observations, multiple 2D physical-field visualizations, and 3D event-level scene modeling, from which we derive 9,074 text-image interleaved question-answer pairs spanning single-choice, multiple-choice, and open-ended report-generation formats. 
FireWorldBench evaluates whether models can infer latent physical states, explain underlying mechanisms, forecast coupled-field evolution, and assess the consequences of interventions from multimodal partial observations. By bridging physical reasoning benchmarks and scientific simulation-grounded fire dynamics, our benchmark provides a challenging testbed for measuring whether current multimodal large language models possess complex physical world intelligence.

\end{abstract}

\section{Introduction}
Recent progress in multimodal large language models (MLLMs) has enabled systems that can describe images, answer visual questions, interpret videos, and interact with external tools. However, whether these models genuinely understand the physical world remains an open question. Many existing benchmarks evaluate physical reasoning through object motion, spatial relations, intuitive mechanics \citep{riochet2018intphys,bakhtin2019phyre}, or short video prediction \citep{yi2020clevrer,bear2021physion}. These settings are valuable, but they often simplify the physical world into discrete objects, isolated events, or visually recognizable dynamics. In contrast, many real-world physical systems are governed by continuous fields, multi-physics coupling, latent state variables. Understanding such systems requires a model to go beyond perception and language plausibility: it must connect partial observations to physical variables, infer hidden states, reason about causal mechanisms.

We refer to this ability as complex \textbf{physical world intelligence}. Unlike object-centric physical reasoning, complex physical world intelligence concerns environments where observable phenomena arise from \textit{the interaction of multiple physical fields and mechanisms.} 
A physical understanding model need to forecast system evolution, ground heterogeneous observations in latent physical states, interpret coupling across multiple fields, infer the mechanisms behind observed changes, and assess hypothetical interventions under partial observability.
This capability is central to MLLMs intended to support scientific analysis, safety-critical monitoring, operational decision-making, embodied manipulation, and real-world problem solving.

In this work, we use \textbf{coupled-field fire dynamics} as a representative testbed for complex physical world intelligence. Fire serves as a compact and highly expressive testbed for complex physical world intelligence, where temperature, soot density, velocity, visibility, carbon-monoxide concentration, and ventilation state interact continuously\citep{mcgrattan2026fds,overholt2014validation}. 
The physical fields in this system constitute a coupled state: their spatial and temporal agreement provides evidence about latent fire state, while their joint response to a changed opening, source, exhaust, or suppression condition supports intervention reasoning. A local ignition source can produce a thermal plume, evolve into ceiling jet flow, drive smoke-layer formation, alter visibility and toxicity, and respond nonlinearly to ventilation or exhaust interventions. These properties make fire dynamics a natural stress test for evaluating whether MLLMs can understand a physical system as an evolving, coupled, and actionable world.

We introduce FireWorldBench, a benchmark designed to evaluate MLLMs on this form of physical understanding. The benchmark is organized along two complementary axes. The first is a physical capability axis, which defines a progression of increasingly demanding physical-world intelligence capabilities: \textit{Temporal Evolution Forecasting, Physical Field Perception and Grounding, Cross-Field Coupling Understanding, Causal Mechanism Attribution, and Counterfactual Intervention Reasoning}. This axis is designed to make physical understanding measurable by decomposing it into progressively harder capabilities, from field-level observation grounding to counterfactual intervention reasoning. The second is a fire scenario task axis, which operationalizes these capabilities through concrete tasks: Localized Onset, Coupled Propagation and Critical Transition. This dual-axis design allows FireWorldBench to remain grounded in a real physical domain while targeting general capabilities needed for complex world understanding.
The benchmark instantiates this organization across 520 fire-world entries, including 494 controlled simulation worlds and 26 real-world-aligned event groups. The controlled collection spans 47 scene archetypes grouped into 7 environment families. It represents each entry through complementary modalities, including structured textual observations, multiple 2D physical-field visualizations, and 3D event-level scene modeling. These representations support 9,074 text-image interleaved QA instances spanning single-choice, multiple-choice, and open-ended report-generation formats.

Beyond evaluation, we investigate three complementary approaches for improving complex physical-world understanding: \textbf{Physical Field Prompting}, which augments the original input with additional observation-aligned physical-field visualizations; \textbf{Chain-of-Thought Reasoning} (CoT), which prompts an explicit observation--reasoning--answer process without external tools; and \textbf{Supervised Fine-Tuning (SFT)}, which adapts a foundation model on the FireWorldBench with coupled multi-physics field visualizations, enabling it to internalize physical-world understanding. We denote the resulting domain-adapted model as \textbf{FireWorldGPT}. Under the same task definitions and evaluation protocol, FireWorldGPT significantly and consistently improves upon its base model across the benchmark's physical capability and fire-task dimensions, demonstrating that domain-specific supervision over coupled-field observations can substantially strengthen physical-world understanding beyond input-level prompting or inference-time reasoning alone.

Our contributions are summarized as follows:
\begin{itemize}[leftmargin=1.35em,itemsep=0.25em,topsep=0.25em]
    \item \textbf{Problem formulation.} We formulate complex physical world intelligence as a benchmark target beyond object-centric physical reasoning, emphasizing coupled continuous fields, latent mechanisms, and counterfactual interventions.
    \item \textbf{Dual-axis benchmark design.} We introduce a dual-axis evaluation design that connects progressive physical capabilities with concrete fire-dynamics tasks, supported by a comprehensive evaluation protocol across observation modalities and answer formats.
    \item \textbf{Comprehensive empirical analysis.} Extensive experiments on simulated and real-world-aligned data reveal current models' capability boundaries, the gap between physical answer and reasoning / explanation, and the limits of parameter scaling.
    \item \textbf{Strategies for improving physical-world understanding.} We systematically compare physical-field prompting, CoT, and supervised fine-tuning, and show that domain adaptation with coupled multi-physics field visualizations enables FireWorldGPT to substantially strengthen the physical-world understanding of its base model.
\end{itemize}

\section{Related Work}
\label{appendix:related}

\subsection{Physical World Understanding Benchmarks}

Understanding the physical world has long been considered a core component of visual intelligence. Early benchmarks typically evaluate intuitive physics through object-centric settings, such as object permanence, collisions, support relations, and goal-directed physical interactions. IntPhys tests whether models can distinguish physically possible from impossible videos, focusing on basic principles such as object permanence and spatio-temporal continuity \citep{riochet2018intphys}. PHYRE introduces a set of 2D physical reasoning puzzles where agents must infer actions that lead to desired physical outcomes \citep{bakhtin2019phyre}. CLEVRER further extends synthetic physical reasoning to event-level video understanding, evaluating descriptive, explanatory, predictive, and counterfactual questions over collision-based scenes \citep{yi2020clevrer}. Physion moves toward more realistic simulated environments and compares machine physical prediction against human judgments \citep{bear2021physion}.

These benchmarks have established important foundations for evaluating physical reasoning beyond static recognition. However, most of them formulate physical understanding around discrete objects, rigid-body interactions, or short-term event dynamics. Consequently, they provide limited coverage of real-world physical systems where observable phenomena emerge from the interaction of continuous fields, latent state variables, and coupled physical mechanisms. In contrast, our work targets \textit{complex physical world intelligence}: the ability to forecast temporal evolution, ground partial observations into physical fields, understand cross-field couplings, attribute causal mechanisms, and reason about counterfactual interventions in physical systems governed by continuously evolving and causally coupled fields. Fire dynamics serves as a canonical testbed for this setting, as it compactly couples thermal, fluid, chemical, and ventilation-driven processes within a single observable world \citep{mcgrattan2026fds,overholt2014validation}.

\subsection{Physical Reasoning in Multimodal Large Language Models}

Recent work has begun to evaluate physical world understanding in multimodal large language models. PhysBench provides a broad benchmark for vision-language models, covering physical object properties, object relationships, physical scene understanding, and physics-based dynamics \citep{chow2025physbench}. It demonstrates that even strong Vision Language Models (VLMs) remain far from robust physical understanding. SpatialBench studies spatial cognition in MLLMs through a hierarchical taxonomy of spatial capabilities \citep{xu2025spatialbench}, while more recent spatial foundation model benchmarks evaluate whether models can generalize across diverse 3D and embodied spatial settings \citep{peng2026spatialbench}. These efforts show that multimodal models still struggle to form physically grounded representations that support state estimation, causal explanation, temporal prediction, and counterfactual reasoning.

Several recent benchmarks have moved beyond coarse answer-level evaluation toward more structured and fine-grained formulations of physical understanding in multimodal models. QuantiPhy emphasizes quantitative physical reasoning, asking models to infer measurable physical quantities such as distance, velocity, acceleration, and temporal change from videos \citep{li2025quantiphy}. FysicsEval unifies perception, prediction, reasoning, and understanding across rigid, deformable, fluid, and material-centered physical scenarios \citep{han2026fysicseval}. CausalPhys introduces causal scaffolding and graph-based evaluation, testing whether models can explain physical outcomes through correct object-attribute-event dependencies \citep{tang2026causalphys}.

FireWorldBench shifts the unit of evaluation from isolated physical phenomena to coherent coupled-field systems. Prior MLLMs benchmarks mostly test isolated physical phenomena, local quantities, or short causal chains. FireWorldBench instead evaluates complex physical world intelligence in \textit{a coupled-field system}, where observable states emerge from the joint evolution of multiple interacting physical processes. This requires models to build a coherent physical-world representation from partial observations, supporting field grounding, coupling understanding, mechanism attribution, temporal forecasting, and counterfactual intervention.

\subsection{World Models and Physics-Aware Video Evaluation}

A parallel line of work evaluates whether video generation models and world models obey physical principles. Benchmarks such as PhyGenBench \citep{meng2024phygenbench}, VideoPhy \citep{bansal2024videophy}, Physics-IQ \citep{motamed2025physicsiq}, WorldBench \citep{upadhyay2026worldbench}, and WorldModelBench \citep{li2025worldmodelbench} assess whether generated or predicted videos are visually plausible and physically consistent. These benchmarks are motivated by the observation that strong generative models may produce realistic-looking videos while violating basic physical laws, temporal consistency, or commonsense interactions.

This line of research is highly relevant, but its primary evaluation target is often future video generation or physical plausibility in generated trajectories. In contrast, FireWorldBench shifts the focus from frame-level physical plausibility to structured physical-world understanding, asking whether a multimodal model can infer latent physical states from partial observations, explain the mechanisms underlying observed phenomena, forecast future risks, and reason about intervention outcomes. In this sense, FireWorldBench evaluates physical world intelligence as \textit{an observation-to-state-to-mechanism-to-intervention chain}, rather than as video prediction alone.

\subsection{Executable and Simulation-Grounded Physical Reasoning}

Recent studies suggest that physical reasoning should be evaluated through explicit, inspectable, and testable representations. VisPhyWorld \citep{liang2026visphyworld} argues that conventional visual question answering or violation-of-expectation tasks may be solved without constructing a falsifiable physical hypothesis. It therefore asks MLLMs to reconstruct observed physical scenes as executable code, making the model’s physical hypothesis inspectable and verifiable through simulation. Scene Dynamic Field \citep{li2026sdf} studies intuitive physics understanding in MLLMs by using simulator-derived dynamic field representations, showing that explicit dynamic-field cues can improve reasoning over continuous media such as fluids and deformable materials.

FireWorldBench builds on this simulation-grounded philosophy but extends it to a more complex physical system. Fire dynamics naturally involve continuous fields, multi-physics coupling, nonlinear evolution, and intervention-sensitive outcomes. Rather than requiring models to implement a full fire simulator, we use simulation-grounded references to define physically meaningful states, mechanisms, futures, and intervention outcomes. This allows us to evaluate whether multimodal models can construct useful physical interpretations from incomplete observations while maintaining a connection to verifiable physical dynamics.

\subsection{Scientific Machine Learning for Coupled Physical Fields}

Scientific machine learning has developed many benchmarks for modeling physical fields and partial differential equations. PDEBench \citep{takamoto2022pdebench} provides standardized datasets for learning PDE-governed systems such as advection, Burgers’ equation, diffusion-reaction systems, Navier-Stokes flows, and shallow-water dynamics. FluidsBench and related CFD benchmarks \citep{fluidsbench2026} evaluate neural operators and surrogate models for fluid simulation. In combustion modeling, Open-CK \citep{fei2025openck} offers simulation-generated field data for controlled industrial fire scenarios, primarily supporting numerical field prediction and surrogate modeling.

These datasets are important because they provide high-resolution physical fields and rigorous numerical evaluation. However, their primary goal is field prediction, PDE approximation, or simulator acceleration. They typically evaluate whether a machine learning model can approximate future physical fields, not whether a multimodal model can interpret observations as evidence of latent physical states, identify causal mechanisms, or reason about interventions. FireWorldBench bridges this gap by converting coupled-field simulation data into multimodal physical world understanding tasks.
\begin{figure}[!t]
    \centering
    \includegraphics[width=\linewidth]{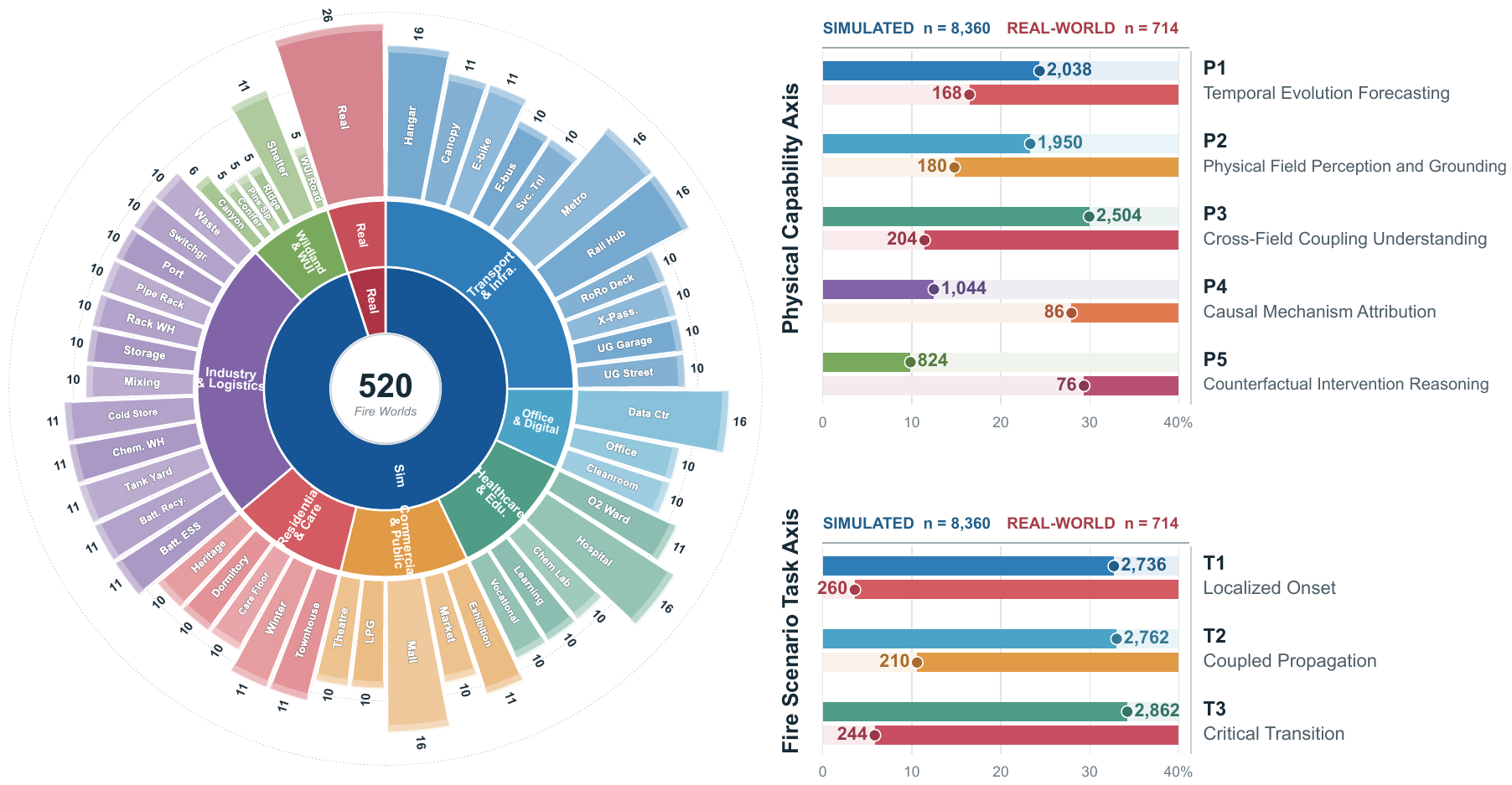}
    \caption{\textbf{Overview of Dataset.}
The left sunburst summarizes the hierarchical composition of 520 fire incidents, while the right panels characterize the coverage of 8,360 simulated and 714 real-world-aligned QA pairs along the five physical capability axis (P1--P5) and fire scenario task axis (T1--T3).}
    \label{fig:dataset}
\end{figure}

\subsection{AI for Fire Dynamics and Smart Firefighting}

Fire safety research has increasingly used machine learning for detection, localization, forecasting, and emergency decision support. Existing studies use visual data \citep{toreyin2006fire,muhammad2018early}, sensor streams, thermal measurements, smoke concentration \citep{jia2026mmodalfire}, and CFD-generated simulations \citep{fang2025fdgen} to detect fire, locate ignition sources, predict smoke movement, estimate temperature fields, and support ventilation control. Tunnel fire datasets and systems such as PolyUFire \citep{zhang2021tunneldb}, intelligent fire location detection for extra-wide immersed tunnels \citep{zhang2024tunnelfire}, AI-driven digital firefighting systems \citep{zhang2022aidfire}, and multimodal indoor fire datasets \citep{jia2026mmodalfire} demonstrate the practical value of data-driven fire modeling.

However, most existing fire AI studies are designed as engineering prediction systems: they optimize classification accuracy, localization error, sensor-based forecasting, or control performance in specific fire-safety scenarios. They do not explicitly evaluate general physical world intelligence in multimodal models. FireWorldBench repositions fire dynamics from an application-specific safety task to a canonical stress test for complex physical world understanding. The benchmark asks not only whether a model can detect a fire, but whether it can understand how coupled physical fields produce the observed state, why the system evolves in a particular way, and how interventions would alter future outcomes.

\subsection{Position of FireWorldBench}

FireWorldBench sits at the intersection of physical reasoning benchmarks, multimodal large language models evaluation, world-model assessment, scientific simulation, and fire-dynamics AI. Existing physical reasoning benchmarks mainly focus on object-level mechanics, spatial relations, or isolated physical events. Video world-model benchmarks evaluate whether generated futures are visually and physically plausible. Scientific ML benchmarks provide high-fidelity physical fields but focus on numerical prediction. Fire AI systems address practical safety tasks but rarely frame fire as a general testbed for physical world intelligence.

In contrast, FireWorldBench uses coupled-field fire dynamics to benchmark complex physical world intelligence. We organize the benchmark along two axes: a physical capability axis, progressing from Temporal Evolution Forecasting, Physical Field Perception and Grounding to Cross-Field Coupling Understanding, Causal Mechanism Attribution, and Counterfactual Intervention Reasoning; and a fire scenario task axis, progressing from Localized Onset, Coupled Propagation, to Critical Transition. This dual-axis design allows fire to serve not as a narrow application domain, but as a structured, simulation-grounded testbed for evaluating whether multimodal models can understand, reason, and intervene in complex physical worlds.

\FloatBarrier

\section{\benchmark}
\label{sec:fireworldbench}

\begin{figure}[!t]
    \centering
    \includegraphics[width=\linewidth]{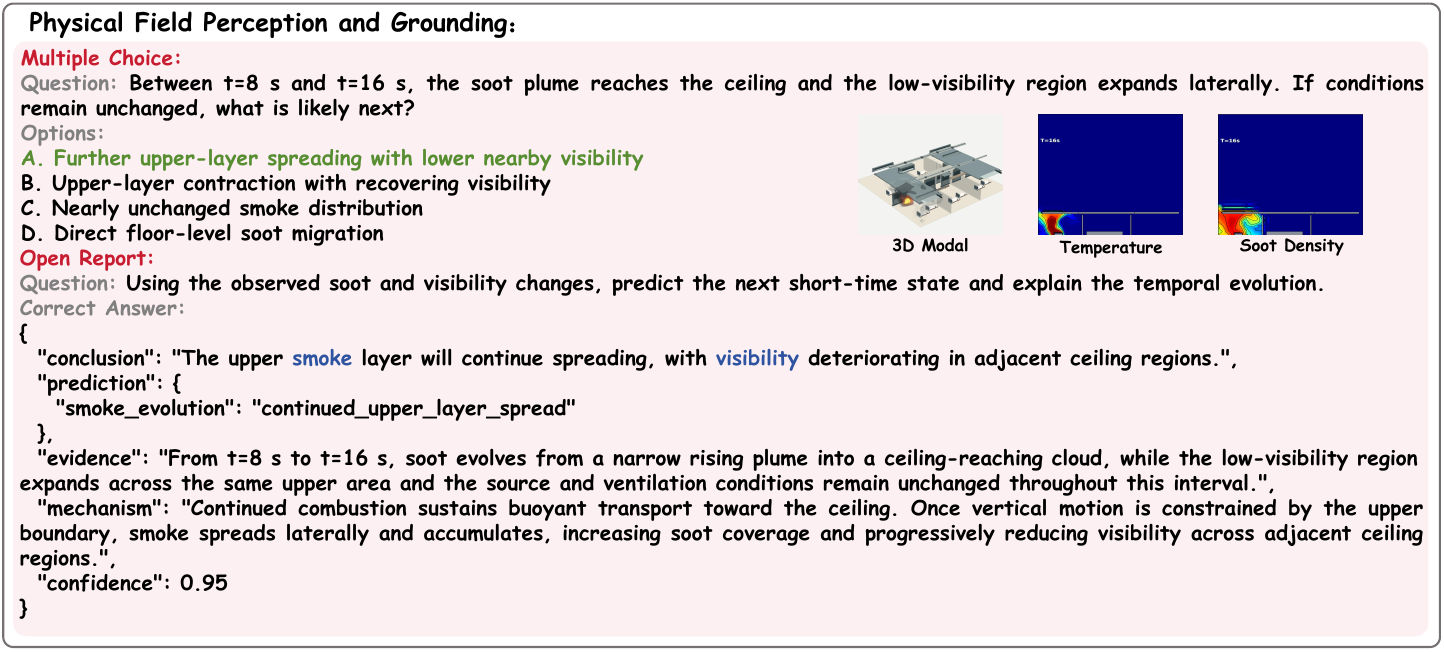}
    \caption{\textbf{Physical field perception and grounding.}
The same fire-world state is queried through choice-format and open-report interfaces using co-registered 3D scene, temperature-field, and soot-density observations. Both interfaces share the same evidence and physical target, while the open report additionally requires an explicit conclusion, supporting evidence, and causal mechanism.}
    \label{fig:capability_examples1}
\end{figure}

FireWorldBench is built upon a diverse collection of coupled-field fire worlds, spanning controlled simulations and real-world-aligned events across heterogeneous environments, fire configurations, observation conditions, and intervention settings. 
Figure~\ref{fig:dataset} provides an overview of the FireWorldBench: the left panel shows scene categories and their fine-grained archetype counts, and the right panel reports instance distributions across both evaluation axes. 
This comprehensive design enables a principled and systematic evaluation of whether models can predict, ground, interpret, explain, and intervene in complex coupled physical worlds.

\subsection{Capabilities, Tasks, and Interfaces}
FireWorldBench organizes every evaluation along two complementary axes. The \emph{physical capability axis} progresses from Temporal Evolution Forecasting (P1), Physical Field Perception and Grounding (P2), Cross-Field Coupling Understanding (P3) to Causal Mechanism Attribution (P4) and Counterfactual Intervention Reasoning (P5). From a fire-evolution perspective, the \emph{fire scenario task axis} spans Localized Onset (T1), Coupled Propagation (T2), Critical Transition (T3). Together, the two axes evaluate models from complementary perspectives of physical-world understanding and fire-dynamics understanding, with tasks designed to span distinct levels of reasoning difficulty.

This design is instantiated across 520 fire-world entries from two complementary sources: 494 controlled simulation worlds and 26 real-world-aligned event groups. The collection spans 47 scene archetypes across 7 environment families: Transportation and Infrastructure, Office and Digital Facilities, Healthcare and Education, Commercial and Public Spaces, Residential and Care Settings, Industry, Energy, and Logistics, and Wildland and Wildland--Urban Interface (WUI). Variations in geometry, confinement, ignition source, combustible materials, ventilation, and propagation conditions produce physically distinct worlds rather than cosmetic scene variants. Each world is represented through structured textual observations, multiple 2D physical-field visualizations, and 3D event-level scene assets.

The benchmark comprises 9,074 QA instances---8,360 from controlled simulations and 714 from real-world-aligned events---across single-choice, multiple-choice, and open-report interfaces. Within the controlled subset, the P1--P5 counts are 2,038, 1,950, 2,504, 1,044, and 824; the corresponding T1--T3 counts are 2,736, 2,762 and 2,862. The remaining 714 instances extend the same capability and task spaces with observed experimental evidence. Matched interfaces hold the physical target fixed: choice-format questions discriminate among plausible hypotheses, whereas open reports connect a conclusion to evidence and mechanism, as illustrated in Figure~\ref{fig:capability_examples1}.

\subsection{Data Collection and Curation}
\begin{wrapfigure}{r}{0.61\textwidth}
    \centering
    \vspace{-3pt}
    \includegraphics[width=\linewidth]{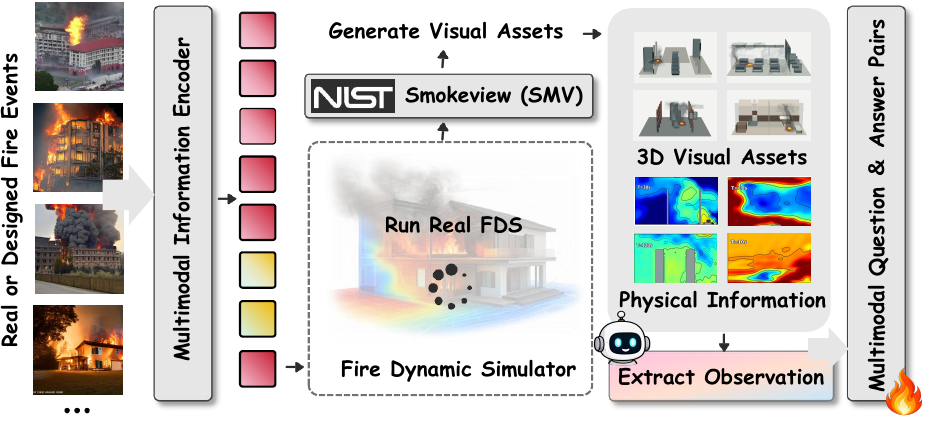}
    \vspace{-7pt}
    \caption{\textbf{Construction pipeline of FireWorldBench.}
Designed scenarios and multimodal records of observed experimental fires are encoded as event constraints and instantiated with FDS. Smokeview produces time-indexed 3D renderings, while the solver state yields aligned physical-field observations. These co-registered outputs are subsequently converted into multimodal question--answer pairs with held-out annotations derived from the underlying physical state.}
    \label{fig:construction_pipeline}
    \vspace{-4pt}
\end{wrapfigure}

Figure~\ref{fig:construction_pipeline} summarizes a world-to-instance pipeline with two complementary starting points. The controlled branch begins from structured specifications of scene geometry, ignition source, combustible materials, ventilation, and intervention variables, making it possible to vary physical causes and compare their consequences. The real-world-aligned branch draws on selected sequences from MmodalFire, which contains 65 indoor-fire videos synchronized with six types of physical sensing data~\cite{jia2026mmodalfire}. A multimodal information encoder synchronizes the recorded streams and translates them into event-level constraints for an aligned Fire Dynamics Simulator (FDS) reconstruction. In this step, the observations anchor the chronology and visible evidence of the recorded event, while the reconstruction supplies the dense latent state required to formulate controlled physical QA instances. Observed and simulation-completed information remain explicitly identified throughout the pipeline.
Each designed or aligned specification is executed with FDS to resolve the spatiotemporal evolution of temperature, soot density, velocity, visibility, and related physical fields. NIST Smokeview converts the solver output into time-indexed 3D scene assets, while an observation extractor derives structured physical records and 2D field visualizations from the same state. Shared geometry, spatial regions, and timestamps keep these representations co-registered, allowing a model's textual and visual evidence to describe the same physical event. In the heptane pool-fire example below, observed frames establish the event chronology, and the aligned 3D reconstruction and FDS-derived fields reveal otherwise latent spatial dynamics. Task-specific temporal cutoffs then define what the model may observe and what it must infer for state reconstruction, forecasting, or intervention reasoning.

Instance construction reverses the simulation process: retained physical states and outcomes define the target, while only task-admissible observations are exposed as evidence. Each logical item is indexed along both benchmark axes and rendered through matched choice-format and open-report interfaces under the same event, modality, and temporal cutoff. Choice alternatives encode physically plausible competing hypotheses; open reports ask the model to connect its prediction to supporting evidence and mechanism. Quality control removes failed or ambiguous simulations and verifies cross-modal alignment, observation sufficiency, answer uniqueness, and consistency between the two answer interfaces. Event-related simulations and paired interventions are grouped within the same split, and construction manifests link every released instance back to its source world and observation configuration.

\begin{figure}[!t]
    \centering
    \includegraphics[width=0.96\linewidth]{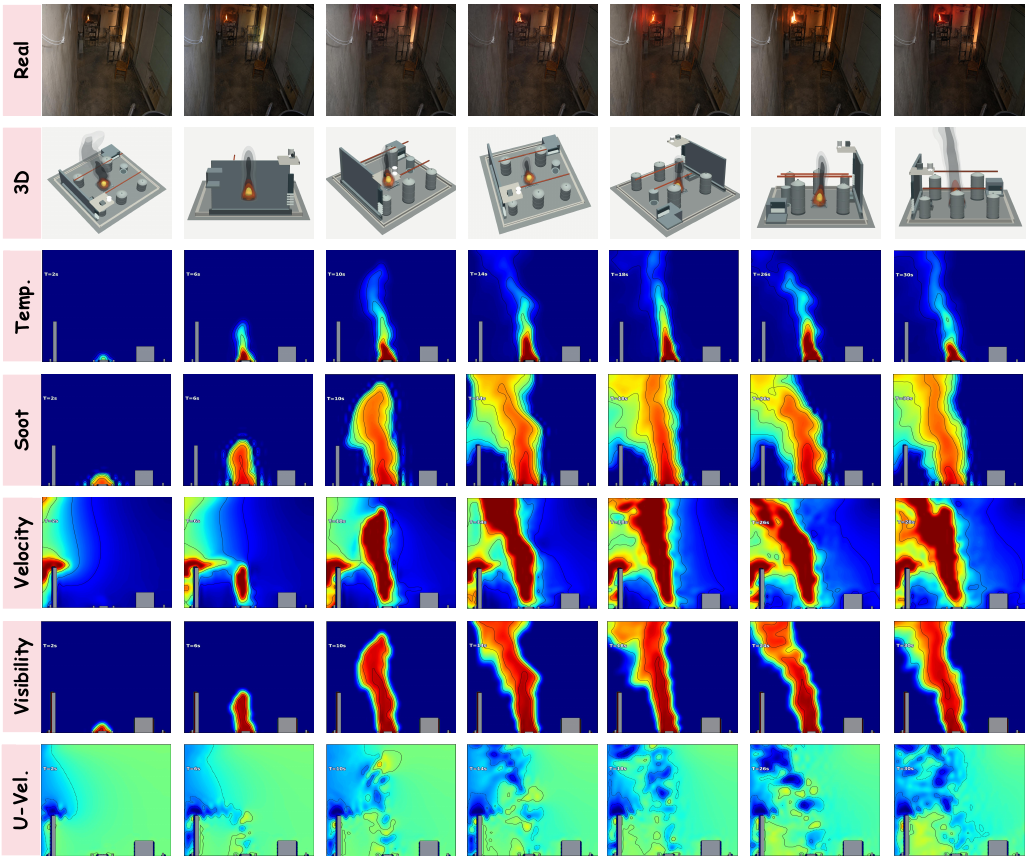}
    \caption{\textbf{Real-world-aligned fire-world reconstruction.}
For a heptane pool-fire experiment, observed video frames anchor the event chronology, while the aligned 3D reconstruction and corresponding FDS realization provide simulation-completed temperature, soot-density, velocity, visibility, and streamwise-velocity fields at matched time points. The sequence preserves the distinction between observed evidence and reconstructed latent physical state.}
    \label{fig:assets}
\end{figure}

\subsection{Evaluation Protocol}
In this work, we evaluate 25 representative models under a unified protocol that fixes the dataset splits, matched task instances, observation policies, response schemas, and deterministic scoring and aggregation procedures across models. We categorize the evaluated models into three groups: 1) \textbf{Frontier closed-source models}, including GPT-5.6-Luna~\citep{openai2025gpt5}, Gemini-2.5-Flash~\citep{comanici2025gemini25}, Claude-Haiku-4.5~\citep{anthropic2025haiku45}, and Qwen3.5-Plus-02-15~\citep{qwen2026qwen35}; 2) \textbf{Open-weight text-only LLMs}, including Llama-3.1-8B-Instruct~\citep{grattafiori2024llama3}, Qwen3-8B~\citep{yang2025qwen3}, Gemma-3-12B-IT and Gemma-3-27B-IT~\citep{gemmateam2025gemma3}, Qwen3-32B~\citep{yang2025qwen3}, Llama-3.3-70B-Instruct~\citep{grattafiori2024llama3}, Qwen3-Next-80B-A3B-Instruct~\citep{qwen2025qwen3next}, GLM-4.5-Air (106B-A12B)~\citep{zeng2025glm45}, Qwen3-235B-A22B~\citep{yang2025qwen3}, DeepSeek-V3.2~\citep{liu2025deepseekv32}, and Kimi-K2-Instruct (1T-A32B)~\citep{kimiteam2025kimik2}; 3) \textbf{Open-weight multimodal LLMs}, comprising InternVL3-8B, InternVL3-38B, and InternVL3-78B~\citep{zhu2025internvl3}, together with Qwen3-VL-8B-Instruct, Qwen3-VL-32B-Instruct, and Qwen3-VL-235B-A22B-Instruct~\citep{bai2025qwen3vl}.

To comprehensively assess the capabilities of each model, we design six general evaluation metrics across different settings. 
\emph{Completion Accuracy} (Acc) measures the fraction of the required gold-standard answer recovered by a prediction, using Jaccard overlap for set-valued choice responses and average slot-wise exact agreement for structured open reports. \textit{Macro-F1} (F1) complements Acc by assessing balanced performance across frequent and infrequent answer outcomes, reducing the dominance of majority outputs in aggregate results. \textit{Brier score} measures the squared discrepancy between a model's reported confidence and exact whole-item correctness, with lower values indicating better confidence--completion consistency.
Beyond these general-purpose metrics, open reports are further evaluated using three explanation-specific measures: \textit{Evidence-F1} (Evi-F1) measures how precisely and completely a generated explanation identifies the spatial regions and physical variables annotated in the reference answer as supporting evidence.; \emph{Mechanism Alignment} (Mech) measures recall of the normalized lexical content carried by the reference causal mechanism; and \emph{Gold-Linked Support} (GLS) measures the fraction of required prediction slots that are both correct and explicitly supported by task-aware evidence or mechanism anchors. 
All metrics except Brier are higher-is-better, and no metric relies on an LLM judge.
Appendix B reports complete metric definitions.

\subsection{Why Fire Dynamics Is a Testbed for Physical-World Intelligence?}
Fire dynamics brings multiple dimensions of physical-world intelligence into a single evolving environment. Heat transfer, buoyancy-driven flow, combustion, smoke and toxic-gas transport, visibility, and ventilation interact continuously, and their spatial distributions jointly determine how hazards emerge and spread. Understanding such a system requires connecting observations to the underlying physical state, identifying the mechanisms that govern its evolution, forecasting future conditions, and assessing the consequences of possible actions. Fire scenarios therefore provide a compact setting for complex physical world intelligence.

Three characteristics make this setting especially demanding. 
First, observations are partial and heterogeneous: sensors, field maps, scene assets, and textual summaries each reveal different aspects of the same event. A model must locate relevant evidence and integrate it across modalities, regions, and time. 
Second, the system state is jointly constrained by \textit{coupled multi-physics fields}. Interpreting any field depends on its spatial and temporal relations with the others. 
Third, physical systems are intervention-sensitive: changes to boundary conditions, source terms, control inputs, or environmental constraints can redirect transport pathways and produce nonlinear downstream effects throughout the system.
These properties directly instantiate the requirements of temporal forecasting, evidence grounding, cross-field understanding, causal attribution, and counterfactual intervention reasoning for physical world intelligence.

\FloatBarrier

\definecolor{fwgroup}{RGB}{242,242,242}
\definecolor{ttgroup}{RGB}{232,232,232}
\definecolor{fwsubgroup}{RGB}{246,246,246}
\definecolor{fwgain}{HTML}{16845B}
\definecolor{fwloss}{HTML}{B8323E}
\definecolor{fwbestblue}{RGB}{145,190,225}
\definecolor{fwsecondblue}{RGB}{213,231,246}
\newcommand{\fwTBD}{\textcolor{gray}{\textemdash}}
\newcommand{\fwup}[1]{\textcolor{fwgain}{\scriptsize$\uparrow$#1}}
\newcommand{\fwdown}[1]{\textcolor{fwloss}{\scriptsize$\downarrow$#1}}
\newcommand{\fwbest}[1]{\cellcolor{fwbestblue}\textbf{#1}}
\newcommand{\fwsecond}[1]{\cellcolor{fwsecondblue}#1}
\newcommand{\fwNineTBD}{\fwTBD&\fwTBD&\fwTBD&\fwTBD&\fwTBD&\fwTBD&\fwTBD&\fwTBD&\fwTBD&\fwTBD&\fwTBD&\fwTBD&\fwTBD&\fwTBD&\fwTBD&\fwTBD&\fwTBD&\fwTBD}
\newcommand{\fwEightTBD}{\fwTBD&\fwTBD&\fwTBD&\fwTBD&\fwTBD&\fwTBD&\fwTBD&\fwTBD}
\newcommand{\fwSixTBD}{\fwTBD&\fwTBD&\fwTBD&\fwTBD&\fwTBD&\fwTBD}

\section{FireWorldGPT}
\label{sec:fireworldgpt}

FireWorldBench identifies whether a model fails to structure physical evidence, organize it into a causal inference, or internalize the reasoning pattern required by the task. 
In this work, we investigate three controlled enhancement strategies targeting these complementary bottlenecks: physics-aware prompting for input-level evidence structuring, chain-of-thought reasoning for physical inference, and supervised fine-tuning for parameter-level domain adaptation. 
The three strategies isolate the contribution of representation, reasoning, and learning to physical world understanding.

\subsection{Physics-Aware Prompting: Coupled-Field Atlas}
\label{sec:physics_aware_prompting}

Physical-world observations are often fragmented across variables, locations, and time, leaving their underlying correspondence implicit and encouraging models to process them as unrelated visual or numerical information. The core idea of the \textit{Coupled-Field Atlas} is to transform these heterogeneous observations into a coordinate-consistent representation by registering physical variables to a shared spatial partition and temporal index while preserving extrema, gradients, flow directions, cross-field co-location, and temporal changes. 
The Prompt-based method explicitly organizes multi-physics evidence to strengthen the model’s unified representation and understanding of complex physical worlds.

\subsection{Chain-of-Thought Reasoning: Constructing Traceable Physical Inference Chains}
\label{sec:chain_of_thought}

Complex physical judgments generally cannot be derived from a single observation. They require a sequence of operations, including evidence identification, state estimation, cross-field association, mechanism attribution, and consequence assessment; models that generate answers directly often bypass these critical steps. The central idea of \textit{Constructing Traceable Physical Inference Chains} is to organize physical reasoning as an executable inference program progressively constrained by observational evidence. Without introducing additional observations, the method guides a frozen model through the evidence-grounded physical inference sequence, requiring each intermediate conclusion to be explicitly linked to an observed physical variable, spatial region, and temporal index before being projected into a unified structured output. In essence, the method transforms implicit physical intuition into an evidence-grounded, auditable, and physically constrained inference chain, thereby strengthening the model’s ability to reason about complex physical systems from partial observations.

\subsection{Supervised Fine-Tuning: Internalizing Observation-to-Representation Mapping}
\label{sec:supervised_fine_tuning}

\begin{table}[t!]
\centering
\caption{\textbf{Overall evaluation along the physical capability axis on the controlled simulation subset.} Models are evaluated by averaging the corresponding choice-format and open-report scores under the same observation track, capability dimension, and metric. Each capability dimension reports accuracy (Acc), macro-F1 (F1), and Brier score, with the Average column summarizing performance across P1--P5.}
\label{tab:main_avg_physical_axis}
\scriptsize
\setlength{\tabcolsep}{2.15pt}
\resizebox{\textwidth}{!}{%
\begin{tabular}{l*{18}{c}}
\toprule
\multirow{3}{*}{\textbf{Model}}
& \multicolumn{3}{c}{\textbf{P1}}
& \multicolumn{3}{c}{\textbf{P2}}
& \multicolumn{3}{c}{\textbf{P3}}
& \multicolumn{3}{c}{\textbf{P4}}
& \multicolumn{3}{c}{\textbf{P5}}
& \multicolumn{3}{c}{\textbf{Average}} \\
\cmidrule(lr){2-4}\cmidrule(lr){5-7}\cmidrule(lr){8-10}
\cmidrule(lr){11-13}\cmidrule(lr){14-16}\cmidrule(lr){17-19}
& Acc & F1 & Brier & Acc & F1 & Brier & Acc & F1 & Brier
& Acc & F1 & Brier & Acc & F1 & Brier & Acc & F1 & Brier \\
\midrule
\rowcolor{ttgroup}\multicolumn{19}{c}{S-track: structured sensor and physical records}\\
\rowcolor{fwgroup}\multicolumn{19}{l}{\textit{Open-weight models: fewer than 20B total parameters}}\\
Llama-3.1-8B-Instruct~\citep{grattafiori2024llama3} & 31.93 & 42.55 & 0.44 & 31.35 & 39.79 & 0.49 & 28.38 & 39.19 & 0.44 & 37.16 & 40.09 & 0.42 & 29.38 & 38.81 & 0.42 & 31.64 & 40.09 & 0.44\\
Qwen3-8B~\citep{yang2025qwen3} & 46.55 & 56.68 & 0.31 & 35.34 & 42.53 & 0.48 & 33.88 & 45.05 & 0.38 & 42.98 & 45.58 & 0.37 & 31.58 & 39.89 & 0.44 & 38.07 & 45.95 & 0.40\\
Gemma-3-12B-IT~\citep{gemmateam2025gemma3} & 34.67 & 44.06 & 0.45 & 38.57 & 48.96 & 0.42 & 33.30 & 44.88 & 0.42 & 34.03 & 36.60 & 0.47 & 27.89 & 36.01 & 0.51 & 33.70 & 42.10 & 0.46\\
\rowcolor{fwgroup}\multicolumn{19}{l}{\textit{Open-weight models: 20--50B total parameters}}\\
Gemma-3-27B-IT~\citep{gemmateam2025gemma3} & 59.04 & 67.47 & 0.23 & 53.40 & \fwsecond{60.59} & 0.31 & \fwsecond{46.56} & \fwsecond{57.64} & 0.31 & 37.08 & 37.95 & 0.47 & \fwsecond{46.85} & \fwsecond{54.32} & 0.33 & 48.59 & 55.59 & 0.33\\
Qwen3-32B~\citep{yang2025qwen3} & 58.30 & 67.64 & 0.21 & 45.76 & 53.59 & 0.34 & 40.16 & 51.84 & 0.33 & 42.67 & 44.80 & 0.37 & 39.61 & 48.33 & 0.35 & 45.30 & 53.24 & 0.32\\
\rowcolor{fwgroup}\multicolumn{19}{l}{\textit{Open-weight models: 50--200B total parameters}}\\
Llama-3.3-70B-Instruct~\citep{grattafiori2024llama3} & 47.08 & 56.89 & 0.22 & 44.00 & 51.33 & 0.32 & 37.11 & 49.34 & 0.26 & 38.02 & 40.23 & 0.33 & 27.80 & 36.88 & 0.34 & 38.80 & 46.94 & 0.29\\
Qwen3-Next-80B-A3B-Instruct~\citep{qwen2025qwen3next} & 48.41 & 58.75 & 0.33 & 42.61 & 51.66 & 0.41 & 39.53 & 51.06 & 0.39 & 32.91 & 34.67 & 0.53 & 33.95 & 42.23 & 0.48 & 39.48 & 47.67 & 0.43\\
GLM-4.5-Air (106B-A12B)~\citep{zeng2025glm45} & 49.53 & 59.97 & 0.21 & 39.76 & 47.24 & 0.36 & 39.84 & 51.64 & 0.28 & 37.49 & 39.95 & 0.36 & 29.56 & 38.02 & 0.35 & 39.23 & 47.36 & 0.31\\
\rowcolor{fwgroup}\multicolumn{19}{l}{\textit{Open-weight models: more than 200B total parameters}}\\
Qwen3-235B-A22B~\citep{yang2025qwen3} & 55.38 & 65.12 & 0.20 & 42.76 & 49.66 & 0.39 & 43.73 & 55.41 & 0.29 & 40.56 & 42.06 & 0.37 & 36.47 & 44.12 & 0.35 & 43.78 & 51.27 & 0.32\\
DeepSeek-V3.2~\citep{liu2025deepseekv32} & 60.70 & 68.97 & 0.18 & 49.23 & 58.04 & \fwsecond{0.28} & 45.12 & 56.37 & 0.24 & \fwsecond{46.91} & \fwsecond{48.36} & \fwsecond{0.31} & 44.45 & 52.86 & \fwbest{0.21} & \fwsecond{49.28} & \fwsecond{56.92} & 0.24\\
Kimi-K2-Instruct (1T-A32B)~\citep{kimiteam2025kimik2} & \fwsecond{61.28} & \fwsecond{70.30} & 0.16 & 51.80 & 59.80 & 0.29 & 44.53 & 55.98 & 0.26 & 40.98 & 43.70 & 0.38 & 38.00 & 45.91 & 0.28 & 47.32 & 55.14 & 0.27\\
\rowcolor{fwgroup}\multicolumn{19}{l}{\textit{Closed-source models}}\\
GPT-5.6-Luna~\citep{openai2025gpt5} & \fwbest{66.59} & \fwbest{74.62} & \fwbest{0.13} & \fwbest{54.44} & 59.98 & 0.33 & \fwbest{55.49} & \fwbest{66.57} & \fwsecond{0.20} & \fwbest{52.14} & \fwbest{53.70} & \fwbest{0.30} & \fwbest{48.07} & \fwbest{54.94} & \fwsecond{0.23} & \fwbest{55.34} & \fwbest{61.96} & \fwsecond{0.24}\\
Gemini-2.5-Flash~\citep{comanici2025gemini25} & 59.92 & 68.51 & 0.23 & 50.81 & 59.02 & 0.34 & 42.52 & 54.16 & 0.34 & 37.32 & 40.62 & 0.47 & 40.41 & 48.51 & 0.34 & 46.20 & 54.16 & 0.35\\
Claude-Haiku-4.5~\citep{anthropic2025haiku45} & 56.56 & 66.30 & \fwsecond{0.14} & \fwsecond{53.95} & \fwbest{62.28} & \fwbest{0.22} & 39.92 & 51.64 & \fwbest{0.20} & 42.03 & 45.77 & 0.31 & 31.45 & 39.12 & 0.24 & 44.78 & 53.02 & \fwbest{0.22}\\
Qwen3.5-Plus-02-15~\citep{qwen2026qwen35} & 60.77 & 70.07 & 0.22 & 47.20 & 53.97 & 0.38 & 42.76 & 53.98 & 0.35 & 40.61 & 42.89 & 0.43 & 44.92 & 53.66 & 0.31 & 47.25 & 54.91 & 0.34\\
\midrule
\rowcolor{ttgroup}\multicolumn{19}{c}{I-track: multimodal physics-rendered visual evidence}\\
\rowcolor{fwgroup}\multicolumn{19}{l}{\textit{Open-weight multimodal models: 7--8B}}\\
InternVL3-8B~\citep{zhu2025internvl3} & 26.86 & 28.87 & 0.46 & 25.08 & 28.82 & 0.41 & 34.08 & 36.55 & 0.43 & 26.59 & 28.96 & 0.42 & 26.27 & 31.93 & 0.45 & 27.77 & 31.03 & 0.43\\
Qwen3-VL-8B-Instruct~\citep{bai2025qwen3vl} & 31.92 & 40.28 & 0.51 & 21.47 & 27.38 & 0.66 & 32.45 & 39.98 & 0.53 & 24.80 & 28.70 & 0.63 & 36.42 & 41.94 & 0.47 & 29.41 & 35.66 & 0.56\\
\rowcolor{fwgroup}\multicolumn{19}{l}{\textit{Open-weight multimodal models: 20--50B}}\\
InternVL3-38B~\citep{zhu2025internvl3} & 41.94 & 40.20 & 0.33 & 26.66 & 30.31 & 0.58 & 37.95 & 41.52 & 0.43 & 35.55 & 34.52 & 0.50 & 31.94 & 37.33 & 0.47 & 34.81 & 36.77 & 0.46\\
Qwen3-VL-32B-Instruct~\citep{bai2025qwen3vl} & \fwsecond{58.28} & \fwsecond{66.45} & \fwsecond{0.18} & 40.23 & \fwbest{46.74} & 0.42 & 41.50 & 49.61 & 0.36 & \fwsecond{36.91} & \fwsecond{39.08} & 0.42 & 35.82 & 40.73 & 0.38 & 42.55 & 48.52 & 0.35\\
\rowcolor{fwgroup}\multicolumn{19}{l}{\textit{Open-weight multimodal models: more than 50B total parameters}}\\
InternVL3-78B~\citep{zhu2025internvl3} & 45.96 & 43.66 & 0.30 & 37.17 & 40.59 & 0.49 & 35.84 & 39.34 & 0.46 & 33.42 & 34.56 & 0.52 & \fwsecond{41.22} & \fwsecond{46.72} & 0.45 & 38.72 & 40.98 & 0.44\\
Qwen3-VL-235B-A22B-Instruct~\citep{bai2025qwen3vl} & 38.02 & 44.83 & 0.34 & 40.42 & 46.46 & 0.42 & 36.66 & 44.05 & 0.41 & 34.38 & 37.81 & 0.42 & 20.62 & 23.96 & 0.36 & 34.02 & 39.42 & 0.39\\
\rowcolor{fwgroup}\multicolumn{19}{l}{\textit{Closed-source multimodal models}}\\
GPT-5.6-Luna~\citep{openai2025gpt5} & \fwbest{60.74} & \fwbest{68.18} & 0.19 & \fwbest{41.11} & 46.69 & \fwbest{0.39} & \fwsecond{44.09} & \fwsecond{51.18} & 0.34 & 34.40 & 38.13 & \fwsecond{0.41} & 34.61 & 39.48 & \fwsecond{0.32} & \fwsecond{42.99} & \fwsecond{48.73} & \fwsecond{0.33}\\
Gemini-2.5-Flash~\citep{comanici2025gemini25} & 52.84 & 59.95 & 0.28 & \fwsecond{40.84} & 46.52 & 0.46 & \fwbest{45.86} & \fwbest{53.70} & \fwsecond{0.34} & \fwbest{41.12} & \fwbest{45.48} & 0.42 & \fwbest{43.46} & \fwbest{48.59} & 0.36 & \fwbest{44.83} & \fwbest{50.85} & 0.37\\
Claude-Haiku-4.5~\citep{anthropic2025haiku45} & 56.04 & 63.92 & \fwbest{0.17} & 34.52 & 40.72 & \fwsecond{0.40} & 40.49 & 48.33 & \fwbest{0.30} & 34.03 & 38.33 & \fwbest{0.38} & 32.47 & 35.98 & \fwbest{0.21} & 39.51 & 45.46 & \fwbest{0.29}\\
Qwen3.5-Plus-02-15~\citep{qwen2026qwen35} & 44.92 & 52.06 & 0.37 & 40.77 & \fwsecond{46.73} & 0.44 & 40.72 & 48.23 & 0.42 & 32.75 & 35.84 & 0.49 & 37.83 & 42.77 & 0.40 & 39.40 & 45.12 & 0.42\\
\bottomrule
\end{tabular}}
\end{table}

Even when physical evidence is explicitly organized or the reasoning process is guided step by step, a foundation model may still lack parameterized representations that reliably transform multi-physics observations into latent states and coupling mechanisms. The central idea of domain-specific fine-tuning is to \textit{internalize the mapping from physical observations to physical states directly into model parameters} through structured evidence. Specifically, FireWorldGPT undergoes 1-epoch autoregressive LoRA fine-tuning on valid observations and standardized structured answers from non-evaluation data, with training samples organized according to a capability progression. At inference time, it uses the standard prompt of the foundation model without any explicit reasoning program or external tool. Without relying on state graphs, relational edges, or intermediate-process supervision, this parameter-level intervention, isolated from Prompt and CoT, examines whether domain adaptation enables more stable representations and reasoning over complex physical worlds.

\section{Experiments}
\label{sec:experiments}

\begin{table}[t!]
\centering
\caption{\textbf{Overall evaluation along the fire scenario task axis on the controlled simulation subset.} Models are evaluated by averaging matched choice-format and open-report scores under the same simulated fire-world entry, observation track, task dimension, and metric. Each task reports accuracy (Acc), macro-F1 (F1), and Brier score, with Average summarizing performance across T1--T3.}
\label{tab:main_avg_fire_axis}
\scriptsize
\setlength{\tabcolsep}{2.15pt}
\resizebox{\textwidth}{!}{%
\begin{tabular}{l*{12}{c}}
\toprule
\multirow{2}{*}{\textbf{Model}}
& \multicolumn{3}{c}{\textbf{T1}}
& \multicolumn{3}{c}{\textbf{T2}}
& \multicolumn{3}{c}{\textbf{T3}}
& \multicolumn{3}{c}{\textbf{Average}} \\
\cmidrule(lr){2-4}\cmidrule(lr){5-7}\cmidrule(lr){8-10}\cmidrule(lr){11-13}
& Acc & F1 & Brier & Acc & F1 & Brier & Acc & F1 & Brier & Acc & F1 & Brier \\
\midrule
\rowcolor{ttgroup}\multicolumn{13}{c}{S-track: structured sensor and physical records}\\
\rowcolor{fwgroup}\multicolumn{13}{l}{\textit{Open-weight models: fewer than 20B total parameters}}\\
Llama-3.1-8B-Instruct~\citep{grattafiori2024llama3} & 32.55 & 40.39 & 0.46 & 29.81 & 39.07 & 0.45 & 32.57 & 41.68 & 0.43 & 31.64 & 40.38 & 0.45\\
Qwen3-8B~\citep{yang2025qwen3} & 37.53 & 43.84 & 0.44 & 35.61 & 45.19 & 0.39 & 44.50 & 52.78 & 0.33 & 39.21 & 47.27 & 0.39\\
Gemma-3-12B-IT~\citep{gemmateam2025gemma3} & 37.82 & 46.69 & 0.43 & 29.27 & 38.84 & 0.47 & 34.58 & 42.19 & 0.46 & 33.89 & 42.57 & 0.46\\
\rowcolor{fwgroup}\multicolumn{13}{l}{\textit{Open-weight models: 20--50B total parameters}}\\
Gemma-3-27B-IT~\citep{gemmateam2025gemma3} & 46.67 & 52.04 & 0.39 & 39.57 & 49.21 & 0.38 & 57.87 & 64.41 & 0.24 & 48.04 & 55.22 & 0.34\\
Qwen3-32B~\citep{yang2025qwen3} & 43.62 & 50.38 & 0.36 & 38.83 & 48.69 & 0.36 & 55.51 & 63.15 & 0.23 & 45.99 & 54.07 & 0.32\\
\rowcolor{fwgroup}\multicolumn{13}{l}{\textit{Open-weight models: 50--200B total parameters}}\\
Llama-3.3-70B-Instruct~\citep{grattafiori2024llama3} & 41.75 & 48.70 & 0.32 & 34.07 & 43.85 & 0.31 & 44.15 & 52.24 & 0.24 & 39.99 & 48.26 & 0.29\\
Qwen3-Next-80B-A3B-Instruct~\citep{qwen2025qwen3next} & 41.19 & 48.28 & 0.43 & 31.81 & 41.05 & 0.50 & 46.86 & 54.91 & 0.35 & 39.95 & 48.08 & 0.43\\
GLM-4.5-Air (106B-A12B)~\citep{zeng2025glm45} & 38.74 & 45.40 & 0.37 & 35.97 & 45.91 & 0.32 & 46.47 & 54.87 & 0.23 & 40.39 & 48.73 & 0.31\\
\rowcolor{fwgroup}\multicolumn{13}{l}{\textit{Open-weight models: more than 200B total parameters}}\\
Qwen3-235B-A22B~\citep{yang2025qwen3} & 41.45 & 47.47 & 0.39 & 39.72 & 49.26 & 0.34 & 52.60 & 60.24 & 0.22 & 44.59 & 52.32 & 0.32\\
DeepSeek-V3.2~\citep{liu2025deepseekv32} & 46.32 & 52.69 & \fwsecond{0.31} & \fwsecond{45.43} & \fwsecond{54.62} & 0.26 & 58.47 & 65.22 & 0.18 & \fwsecond{50.07} & \fwsecond{57.51} & 0.25\\
Kimi-K2-Instruct (1T-A32B)~\citep{kimiteam2025kimik2} & 47.46 & 53.20 & 0.35 & 41.50 & 51.17 & 0.30 & 57.55 & 64.63 & 0.18 & 48.83 & 56.33 & 0.27\\
\rowcolor{fwgroup}\multicolumn{13}{l}{\textit{Closed-source models}}\\
GPT-5.6-Luna~\citep{openai2025gpt5} & \fwbest{49.62} & \fwsecond{54.05} & 0.35 & \fwbest{56.51} & \fwbest{64.91} & \fwsecond{0.23} & \fwbest{63.90} & \fwbest{70.05} & \fwbest{0.14} & \fwbest{56.68} & \fwbest{63.00} & \fwsecond{0.24}\\
Gemini-2.5-Flash~\citep{comanici2025gemini25} & 46.64 & 52.80 & 0.40 & 37.95 & 47.53 & 0.40 & 56.96 & 63.87 & 0.24 & 47.18 & 54.73 & 0.35\\
Claude-Haiku-4.5~\citep{anthropic2025haiku45} & \fwsecond{47.53} & \fwbest{54.08} & \fwbest{0.30} & 40.93 & 50.77 & \fwbest{0.22} & 51.96 & 59.99 & \fwsecond{0.15} & 46.80 & 54.94 & \fwbest{0.22}\\
Qwen3.5-Plus-02-15~\citep{qwen2026qwen35} & 44.96 & 49.98 & 0.40 & 39.36 & 49.08 & 0.39 & \fwsecond{58.84} & \fwsecond{66.26} & 0.23 & 47.72 & 55.11 & 0.34\\
\midrule
\rowcolor{ttgroup}\multicolumn{13}{c}{I-track: multimodal physics-rendered visual evidence}\\
\rowcolor{fwgroup}\multicolumn{13}{l}{\textit{Open-weight multimodal models: 7--8B}}\\
InternVL3-8B~\citep{zhu2025internvl3} & 26.91 & 37.90 & \fwbest{0.36} & 29.93 & 42.42 & 0.47 & 26.66 & 36.83 & 0.46 & 27.83 & 39.05 & 0.43\\
Qwen3-VL-8B-Instruct~\citep{bai2025qwen3vl} & 23.34 & 28.32 & 0.66 & 28.15 & 35.27 & 0.56 & 34.35 & 40.86 & 0.49 & 28.61 & 34.82 & 0.57\\
\rowcolor{fwgroup}\multicolumn{13}{l}{\textit{Open-weight multimodal models: 20--50B}}\\
InternVL3-38B~\citep{zhu2025internvl3} & 30.47 & 39.48 & 0.56 & 35.59 & \fwsecond{48.39} & 0.46 & 38.46 & 50.12 & 0.38 & 34.84 & 45.99 & 0.46\\
Qwen3-VL-32B-Instruct~\citep{bai2025qwen3vl} & 40.53 & 46.13 & 0.42 & 36.96 & 44.45 & 0.39 & \fwsecond{50.86} & \fwsecond{57.52} & 0.25 & 42.78 & 49.37 & 0.35\\
\rowcolor{fwgroup}\multicolumn{13}{l}{\textit{Open-weight multimodal models: more than 50B total parameters}}\\
InternVL3-78B~\citep{zhu2025internvl3} & 37.70 & \fwbest{49.35} & 0.50 & 33.51 & 46.06 & 0.48 & 44.32 & 55.89 & 0.35 & 38.51 & \fwsecond{50.43} & 0.44\\
Qwen3-VL-235B-A22B-Instruct~\citep{bai2025qwen3vl} & 38.63 & 43.18 & 0.46 & 35.58 & 42.82 & 0.38 & 32.93 & 37.58 & 0.35 & 35.71 & 41.19 & 0.40\\
\rowcolor{fwgroup}\multicolumn{13}{l}{\textit{Closed-source multimodal models}}\\
GPT-5.6-Luna~\citep{openai2025gpt5} & 40.77 & 44.86 & 0.42 & \fwsecond{39.21} & 46.07 & \fwsecond{0.34} & \fwbest{51.67} & \fwbest{58.21} & \fwsecond{0.24} & \fwsecond{43.88} & 49.71 & \fwsecond{0.33}\\
Gemini-2.5-Flash~\citep{comanici2025gemini25} & \fwbest{42.42} & \fwsecond{46.45} & 0.45 & \fwbest{42.67} & \fwbest{50.52} & 0.37 & 49.59 & 56.00 & 0.31 & \fwbest{44.89} & \fwbest{50.99} & 0.37\\
Claude-Haiku-4.5~\citep{anthropic2025haiku45} & 36.62 & 41.29 & \fwsecond{0.41} & 35.96 & 43.54 & \fwbest{0.32} & 47.86 & 54.22 & \fwbest{0.18} & 40.15 & 46.35 & \fwbest{0.30}\\
Qwen3.5-Plus-02-15~\citep{qwen2026qwen35} & \fwsecond{41.99} & 45.81 & 0.43 & 34.62 & 41.89 & 0.47 & 42.46 & 48.83 & 0.38 & 39.69 & 45.51 & 0.42\\
\bottomrule
\end{tabular}}
\end{table}

\subsection{Controlled Simulation Results}
\label{sec:main_results}

Table~\ref{tab:main_avg_physical_axis} provides an integrated evaluation of complex physical world understanding on the controlled simulation subset along the physical capability axis. It aggregates choice-format and open-report results under matched observation tracks, capability dimensions, and metrics, thereby comparing models beyond a single response interface. 
The S-track evaluates reasoning from structured sensor and physical records, while the I-track evaluates reasoning from multimodal physics-rendered visual evidence. \textit{Acc and Macro-F1 measure predictive quality, while lower Brier scores indicate better calibration.} 
The aggregated results show that current models still have clear capability boundaries in complex physical world understanding. GPT-5.6-Luna leads the S-track (55.34 Acc / 61.96 F1), whereas Gemini-2.5-Flash leads the I-track (44.83 Acc / 50.85 F1); Claude-Haiku-4.5 achieves the lowest average Brier score on both tracks. 

Table~\ref{tab:main_avg_fire_axis} provides an integrated evaluation of complex physical world understanding on the controlled simulation subset along the fire scenario task axis. It aggregates choice-format and open-report results under matched simulated fire-world entries, observation tracks, task dimensions, and metrics, thereby reducing interface-specific variance and focusing on task-level physical competence. 
The results show that task-level physical understanding remains strongly observation-track dependent, even under controlled conditions. GPT-5.6-Luna leads the S-track (56.68 Acc / 63.00 F1), whereas Gemini-2.5-Flash leads the I-track (44.89 Acc / 50.99 F1); Claude-Haiku-4.5 achieves the lowest average Brier score on both tracks. 
   
\begin{takeawaybox}
\textbf{Takeaway}: Even in clean, controlled fire worlds, current multimodal large language models do not exhibit robust complex physical world intelligence: clear limitations persist in both physical-capability reasoning and task-level problem solving.
\end{takeawaybox}

\begin{table}[t]
\centering
\caption{\textbf{Overall evaluation along the physical capability axis on the real-world-aligned subset.} Models are evaluated by aggregating the corresponding choice-format and open-report results under the same real-world-aligned event groups, observation track, capability dimension, and metric. Each capability dimension reports accuracy (Acc), macro-F1 (F1), and Brier score, with the Average column summarizing performance across P1--P5.}
\label{tab:rwa_avg_physical_axis}
\scriptsize
\setlength{\tabcolsep}{2.15pt}
\resizebox{\textwidth}{!}{%
\begin{tabular}{l*{18}{c}}
\toprule
\multirow{3}{*}{\textbf{Model}}
& \multicolumn{3}{c}{\textbf{P1}}
& \multicolumn{3}{c}{\textbf{P2}}
& \multicolumn{3}{c}{\textbf{P3}}
& \multicolumn{3}{c}{\textbf{P4}}
& \multicolumn{3}{c}{\textbf{P5}}
& \multicolumn{3}{c}{\textbf{Average}} \\
\cmidrule(lr){2-4}\cmidrule(lr){5-7}\cmidrule(lr){8-10}
\cmidrule(lr){11-13}\cmidrule(lr){14-16}\cmidrule(lr){17-19}
& Acc & F1 & Brier & Acc & F1 & Brier & Acc & F1 & Brier
& Acc & F1 & Brier & Acc & F1 & Brier & Acc & F1 & Brier \\
\midrule
\rowcolor{ttgroup}\multicolumn{19}{c}{S-track: structured sensor and physical records}\\
\rowcolor{fwgroup}\multicolumn{19}{l}{\textit{Open-weight models: fewer than 20B total parameters}}\\
Llama-3.1-8B-Instruct~\citep{grattafiori2024llama3} & 35.16 & 43.98 & 0.46 & 30.85 & 39.77 & 0.48 & 32.95 & 41.36 & 0.43 & 29.25 & 37.84 & 0.46 & 34.49 & 44.73 & 0.36 & 32.54 & 41.53 & 0.44\\
Qwen3-8B~\citep{yang2025qwen3} & 37.95 & 45.41 & 0.49 & 35.06 & 43.30 & 0.49 & 48.86 & 57.95 & 0.31 & 39.06 & 48.25 & 0.42 & 34.49 & 45.00 & 0.41 & 39.09 & 47.98 & 0.42\\
Gemma-3-12B-IT~\citep{gemmateam2025gemma3} & 43.73 & 52.13 & 0.35 & 44.34 & 55.99 & 0.34 & 39.43 & 47.23 & 0.38 & 40.54 & 49.12 & 0.41 & 38.20 & 47.50 & 0.40 & 41.25 & 50.40 & 0.38\\
\rowcolor{fwgroup}\multicolumn{19}{l}{\textit{Open-weight models: 20--50B total parameters}}\\
Gemma-3-27B-IT~\citep{gemmateam2025gemma3} & 56.94 & 65.09 & 0.27 & 43.15 & 54.25 & 0.35 & 44.13 & 49.62 & 0.43 & 43.64 & 53.41 & 0.37 & 36.30 & 46.36 & 0.40 & 44.83 & 53.75 & 0.37\\
Qwen3-32B~\citep{yang2025qwen3} & 62.71 & 69.47 & 0.19 & 45.58 & 54.70 & 0.34 & \fwsecond{52.12} & \fwsecond{60.53} & 0.27 & 39.14 & 45.63 & 0.43 & 38.66 & 49.26 & 0.36 & 47.64 & 55.92 & 0.32\\

\rowcolor{fwgroup}\multicolumn{19}{l}{\textit{Open-weight models: 50--200B total parameters}}\\
Llama-3.3-70B-Instruct~\citep{grattafiori2024llama3} & 66.12 & 72.94 & \fwsecond{0.15} & 48.53 & 58.81 & 0.23 & 47.35 & 54.02 & 0.26 & 49.02 & 58.66 & \fwbest{0.23} & 45.00 & 55.09 & \fwsecond{0.22} & 51.20 & 59.91 & \fwsecond{0.22}\\
Qwen3-Next-80B-A3B-Instruct~\citep{qwen2025qwen3next} & 52.98 & 60.05 & 0.31 & 56.27 & 65.96 & 0.26 & 46.44 & 54.85 & 0.36 & 46.14 & 54.91 & 0.37 & 43.05 & 52.97 & 0.38 & 48.98 & 57.75 & 0.34\\
GLM-4.5-Air (106B-A12B)~\citep{zeng2025glm45} & 66.66 & 72.41 & 0.17 & 53.31 & 61.62 & 0.25 & 51.33 & 57.50 & 0.28 & 45.61 & 53.28 & 0.31 & 35.84 & 47.31 & 0.28 & 50.55 & 58.43 & 0.26\\
\rowcolor{fwgroup}\multicolumn{19}{l}{\textit{Open-weight models: more than 200B total parameters}}\\
Qwen3-235B-A22B~\citep{yang2025qwen3} & 53.80 & 60.64 & 0.24 & 47.33 & 57.06 & 0.32 & 46.41 & 54.39 & 0.33 & 41.86 & 50.41 & 0.36 & 32.96 & 43.89 & 0.36 & 44.47 & 53.28 & 0.32\\
DeepSeek-V3.2~\citep{liu2025deepseekv32} & 63.36 & 68.94 & 0.17 & 59.50 & \fwsecond{69.38} & \fwsecond{0.17} & 47.23 & 53.03 & 0.26 & 40.52 & 46.41 & 0.38 & 35.60 & 44.73 & 0.31 & 49.24 & 56.50 & 0.26\\
Kimi-K2-Instruct (1T-A32B)~\citep{kimiteam2025kimik2} & 61.17 & 66.02 & 0.15 & 60.23 & 69.06 & 0.18 & 46.21 & 52.58 & 0.33 & 41.46 & 47.22 & 0.38 & 38.61 & 47.59 & 0.29 & 49.54 & 56.49 & 0.27\\
\rowcolor{fwgroup}\multicolumn{19}{l}{\textit{Closed-source models}}\\
GPT-5.6-Luna~\citep{openai2025gpt5} & \fwbest{75.29} & \fwbest{78.97} & \fwbest{0.13} & \fwsecond{61.60} & 69.16 & 0.22 & \fwbest{53.64} & \fwbest{62.05} & \fwbest{0.25} & \fwbest{60.34} & \fwbest{63.34} & 0.29 & \fwsecond{49.95} & \fwsecond{58.80} & \fwbest{0.21} & \fwbest{60.16} & \fwbest{66.46} & \fwbest{0.22}\\
Gemini-2.5-Flash~\citep{comanici2025gemini25} & \fwsecond{69.79} & \fwsecond{75.14} & 0.15 & 57.55 & 67.14 & 0.23 & 48.48 & 55.08 & 0.34 & 48.62 & 56.25 & 0.35 & 47.31 & 56.30 & 0.29 & 54.35 & 61.98 & 0.27\\
Claude-Haiku-4.5~\citep{anthropic2025haiku45} & 63.97 & 70.40 & 0.16 & \fwbest{65.12} & \fwbest{73.31} & \fwbest{0.15} & 50.53 & 58.03 & \fwsecond{0.26} & \fwsecond{55.27} & \fwsecond{61.21} & \fwsecond{0.27} & 40.37 & 48.34 & 0.29 & 55.05 & 62.26 & 0.22\\
Qwen3.5-Plus-02-15~\citep{qwen2026qwen35} & 64.53 & 70.50 & 0.22 & 57.45 & 66.52 & 0.25 & 50.76 & 58.26 & 0.33 & 53.44 & 60.29 & 0.31 & \fwbest{50.55} & \fwbest{60.83} & 0.28 & \fwsecond{55.35} & \fwsecond{63.28} & 0.28\\
\midrule
\rowcolor{ttgroup}\multicolumn{19}{c}{I-track: multimodal physics-rendered visual evidence}\\
\rowcolor{fwgroup}\multicolumn{19}{l}{\textit{Open-weight multimodal models: 7--8B}}\\
InternVL3-8B~\citep{zhu2025internvl3} & 19.20 & 19.93 & 0.60 & 28.80 & 29.70 & 0.32 & 26.58 & 28.41 & 0.40 & 29.64 & 32.66 & 0.32 & 30.41 & 36.25 & 0.42 & 26.93 & 29.39 & 0.41\\
Qwen3-VL-8B-Instruct~\citep{bai2025qwen3vl} & 42.36 & 45.89 & 0.44 & 41.76 & 48.93 & 0.45 & 29.00 & 35.59 & 0.58 & 30.50 & 36.14 & 0.57 & 39.79 & 45.25 & 0.47 & 36.68 & 42.36 & 0.50\\
\rowcolor{fwgroup}\multicolumn{19}{l}{\textit{Open-weight multimodal models: 20--50B}}\\
InternVL3-38B~\citep{zhu2025internvl3} & 31.03 & 29.77 & 0.45 & 38.70 & 39.48 & 0.43 & 35.95 & 37.06 & 0.46 & 46.25 & 45.95 & 0.40 & 31.16 & 36.50 & 0.48 & 36.62 & 37.75 & 0.44\\
Qwen3-VL-32B-Instruct~\citep{bai2025qwen3vl} & 44.73 & 49.08 & 0.33 & \fwsecond{53.59} & \fwsecond{59.92} & 0.31 & 39.00 & 46.09 & 0.40 & \fwsecond{50.97} & \fwsecond{58.03} & \fwsecond{0.31} & \fwbest{48.62} & \fwbest{53.16} & 0.31 & 47.38 & 53.26 & 0.33\\
\rowcolor{fwgroup}\multicolumn{19}{l}{\textit{Open-weight multimodal models: more than 50B total parameters}}\\
InternVL3-78B~\citep{zhu2025internvl3} & 27.56 & 25.01 & 0.54 & 37.85 & 39.70 & 0.43 & 34.11 & 34.62 & 0.47 & 46.22 & 45.98 & 0.40 & \fwsecond{46.62} & \fwsecond{52.16} & 0.42 & 38.47 & 39.50 & 0.45\\
Qwen3-VL-235B-A22B-Instruct~\citep{bai2025qwen3vl} & 57.70 & 59.80 & 0.28 & 41.50 & 47.89 & 0.43 & 30.77 & 38.84 & 0.48 & 44.39 & 49.28 & 0.42 & 32.96 & 38.34 & 0.41 & 41.47 & 46.83 & 0.40\\
\rowcolor{fwgroup}\multicolumn{19}{l}{\textit{Closed-source multimodal models}}\\
GPT-5.6-Luna~\citep{openai2025gpt5} & \fwbest{59.37} & \fwbest{61.94} & \fwsecond{0.20} & \fwbest{60.50} & \fwbest{65.92} & \fwbest{0.20} & \fwbest{41.30} & \fwsecond{47.33} & \fwsecond{0.31} & \fwbest{53.48} & \fwbest{58.18} & \fwbest{0.30} & 42.91 & 46.50 & \fwsecond{0.27} & \fwbest{51.51} & \fwbest{55.97} & \fwbest{0.26}\\
Gemini-2.5-Flash~\citep{comanici2025gemini25} & \fwsecond{58.96} & 60.70 & 0.30 & 49.81 & 55.00 & 0.39 & 37.69 & 46.88 & 0.40 & 47.23 & 53.86 & 0.39 & 44.54 & 50.66 & 0.35 & \fwsecond{47.65} & \fwsecond{53.42} & 0.37\\
Claude-Haiku-4.5~\citep{anthropic2025haiku45} & 58.35 & \fwsecond{61.68} & \fwbest{0.19} & 51.64 & 57.26 & \fwsecond{0.28} & \fwsecond{41.25} & \fwbest{48.80} & \fwbest{0.28} & 43.58 & 49.34 & 0.33 & 34.91 & 40.00 & \fwbest{0.24} & 45.95 & 51.41 & \fwsecond{0.26}\\
Qwen3.5-Plus-02-15~\citep{qwen2026qwen35} & 54.39 & 56.34 & 0.35 & 52.41 & 58.85 & 0.34 & 39.86 & 47.09 & 0.43 & 40.70 & 47.08 & 0.42 & 30.00 & 36.16 & 0.41 & 43.47 & 49.11 & 0.39\\
\bottomrule
\end{tabular}}
\end{table}

\FloatBarrier
\subsection{Real-World Alignment Results}
\label{sec:rwa_results}

Table~\ref{tab:rwa_avg_physical_axis} provides an integrated evaluation of complex physical-world understanding on the real-world-aligned subset along the physical capability axis.
Even in the comparatively simpler real-world-aligned setting, the results expose persistent boundaries in physical-world understanding and remain strongly observation-track dependent.
GPT-5.6-Luna leads both the S-track (60.16 Acc / 66.46 F1 / 0.22 Brier) and the I-track (51.51 Acc / 55.97 F1 / 0.26 Brier). Across the four frontier closed models evaluated on both tracks, structured records retain an average advantage of 9.09 Acc / 11.02 F1 over physics-rendered visual evidence. This gap indicates that real-world-aligned field visualizations remain difficult to map into stable latent-state, mechanism, and intervention judgments.

\begin{table}[t!]
\centering
\caption{\textbf{Overall evaluation along the fire scenario task axis on the real-world-aligned subset.} Models are evaluated by averaging matched choice-format and open-report scores under the same real-world-aligned event group, observation track, task dimension, and metric. Each task reports accuracy (Acc), macro-F1 (F1), and Brier score, with Average summarizing performance across T1--T3.}
\label{tab:rwa_avg_fire_axis}
\scriptsize
\setlength{\tabcolsep}{2.15pt}
\resizebox{\textwidth}{!}{%
\begin{tabular}{l*{12}{c}}
\toprule
\multirow{2}{*}{\textbf{Model}}
& \multicolumn{3}{c}{\textbf{T1}}
& \multicolumn{3}{c}{\textbf{T2}}
& \multicolumn{3}{c}{\textbf{T3}}
& \multicolumn{3}{c}{\textbf{Average}} \\
\cmidrule(lr){2-4}\cmidrule(lr){5-7}\cmidrule(lr){8-10}\cmidrule(lr){11-13}
& Acc & F1 & Brier & Acc & F1 & Brier & Acc & F1 & Brier & Acc & F1 & Brier \\
\midrule
\rowcolor{ttgroup}\multicolumn{13}{c}{S-track: structured sensor and physical records}\\
\rowcolor{fwgroup}\multicolumn{13}{l}{\textit{Open-weight models: fewer than 20B total parameters}}\\
Llama-3.1-8B-Instruct~\citep{grattafiori2024llama3} & 33.84 & 34.05 & 0.44 & 25.69 & 30.04 & 0.50 & 34.96 & 37.76 & 0.43 & 31.50 & 33.95 & 0.46\\
Qwen3-8B~\citep{yang2025qwen3} & 39.12 & 40.15 & 0.45 & 39.94 & 41.73 & 0.39 & 36.93 & 41.02 & 0.47 & 38.66 & 40.97 & 0.44\\
Gemma-3-12B-IT~\citep{gemmateam2025gemma3} & 45.82 & 46.33 & 0.34 & 35.67 & 39.71 & 0.43 & 42.10 & 43.52 & 0.37 & 41.20 & 43.18 & 0.38\\
\rowcolor{fwgroup}\multicolumn{13}{l}{\textit{Open-weight models: 20--50B total parameters}}\\
Gemma-3-27B-IT~\citep{gemmateam2025gemma3} & 43.95 & 44.05 & 0.36 & 42.89 & 43.80 & 0.40 & 50.85 & 52.98 & 0.31 & 45.89 & 46.94 & 0.36\\
Qwen3-32B~\citep{yang2025qwen3} & 41.95 & 42.86 & 0.40 & 48.58 & 49.92 & 0.30 & 55.61 & 56.71 & 0.24 & 48.71 & 49.83 & 0.31\\
\rowcolor{fwgroup}\multicolumn{13}{l}{\textit{Open-weight models: 50--200B total parameters}}\\
Llama-3.3-70B-Instruct~\citep{grattafiori2024llama3} & 49.77 & 51.74 & \fwsecond{0.24} & 46.42 & 46.78 & 0.24 & 59.90 & 61.69 & \fwsecond{0.17} & 52.03 & 53.40 & 0.22\\
Qwen3-Next-80B-A3B-Instruct~\citep{qwen2025qwen3next} & 51.17 & 53.27 & 0.32 & 49.21 & 51.72 & 0.32 & 50.06 & 51.65 & 0.33 & 50.14 & 52.21 & 0.32\\
GLM-4.5-Air (106B-A12B)~\citep{zeng2025glm45} & 49.17 & 50.86 & 0.30 & 51.31 & 52.90 & 0.25 & 57.56 & 59.20 & 0.20 & 52.68 & 54.32 & 0.25\\
\rowcolor{fwgroup}\multicolumn{13}{l}{\textit{Open-weight models: more than 200B total parameters}}\\
Qwen3-235B-A22B~\citep{yang2025qwen3} & 45.42 & 44.88 & 0.35 & 44.52 & 46.75 & 0.33 & 47.66 & 49.45 & 0.28 & 45.86 & 47.03 & 0.32\\
DeepSeek-V3.2~\citep{liu2025deepseekv32} & 51.30 & 52.01 & 0.27 & 47.40 & 48.16 & 0.27 & 55.17 & 56.68 & 0.21 & 51.29 & 52.28 & 0.25\\
Kimi-K2-Instruct (1T-A32B)~\citep{kimiteam2025kimik2} & 51.98 & 52.99 & 0.29 & 47.48 & 48.59 & 0.27 & 54.51 & 54.46 & 0.19 & 51.32 & 52.01 & 0.25\\
\rowcolor{fwgroup}\multicolumn{13}{l}{\textit{Closed-source models}}\\
GPT-5.6-Luna~\citep{openai2025gpt5} & \fwsecond{56.24} & \fwsecond{56.44} & 0.29 & \fwbest{64.60} & \fwbest{63.22} & \fwbest{0.19} & \fwbest{67.82} & \fwbest{67.90} & \fwbest{0.15} & \fwbest{62.88} & \fwbest{62.52} & \fwsecond{0.21}\\
Gemini-2.5-Flash~\citep{comanici2025gemini25} & 52.48 & 52.96 & 0.32 & 52.04 & 52.61 & 0.27 & \fwsecond{63.16} & \fwsecond{62.79} & 0.20 & 55.89 & 56.12 & 0.26\\
Claude-Haiku-4.5~\citep{anthropic2025haiku45} & \fwbest{58.65} & \fwbest{58.91} & \fwbest{0.23} & 57.95 & 57.56 & \fwsecond{0.19} & 57.01 & 59.29 & 0.20 & \fwsecond{57.87} & \fwsecond{58.58} & \fwbest{0.21}\\
Qwen3.5-Plus-02-15~\citep{qwen2026qwen35} & 52.10 & 52.60 & 0.33 & \fwsecond{58.44} & \fwsecond{60.00} & 0.24 & 60.41 & 62.29 & 0.24 & 56.98 & 58.30 & 0.27\\
\midrule
\rowcolor{ttgroup}\multicolumn{13}{c}{I-track: multimodal physics-rendered visual evidence}\\
\rowcolor{fwgroup}\multicolumn{13}{l}{\textit{Open-weight multimodal models: 7--8B}}\\
InternVL3-8B~\citep{zhu2025internvl3} & 28.14 & 39.70 & \fwsecond{0.30} & 28.66 & 40.63 & 0.38 & 22.88 & 28.35 & 0.54 & 26.56 & 36.23 & 0.41\\
Qwen3-VL-8B-Instruct~\citep{bai2025qwen3vl} & 37.31 & 39.62 & 0.52 & 30.56 & 31.11 & 0.55 & 41.52 & 43.70 & 0.45 & 36.46 & 38.14 & 0.50\\
\rowcolor{fwgroup}\multicolumn{13}{l}{\textit{Open-weight multimodal models: 20--50B}}\\
InternVL3-38B~\citep{zhu2025internvl3} & 35.19 & 45.65 & 0.51 & \fwsecond{45.75} & \fwbest{56.25} & 0.35 & 31.08 & 38.86 & 0.46 & 37.34 & 46.92 & 0.44\\
Qwen3-VL-32B-Instruct~\citep{bai2025qwen3vl} & \fwsecond{53.06} & \fwsecond{52.88} & 0.32 & 43.22 & 43.25 & 0.36 & 46.01 & 46.17 & 0.32 & 47.43 & 47.43 & 0.33\\
\rowcolor{fwgroup}\multicolumn{13}{l}{\textit{Open-weight multimodal models: more than 50B total parameters}}\\
InternVL3-78B~\citep{zhu2025internvl3} & 37.49 & 49.96 & 0.47 & 41.63 & \fwsecond{52.44} & 0.39 & 33.82 & 42.26 & 0.50 & 37.64 & 48.22 & 0.45\\
Qwen3-VL-235B-A22B-Instruct~\citep{bai2025qwen3vl} & 38.73 & 38.82 & 0.48 & 39.60 & 40.56 & 0.40 & 49.59 & 50.84 & 0.32 & 42.64 & 43.40 & 0.40\\
\rowcolor{fwgroup}\multicolumn{13}{l}{\textit{Closed-source multimodal models}}\\
GPT-5.6-Luna~\citep{openai2025gpt5} & \fwbest{55.50} & \fwbest{55.36} & \fwbest{0.28} & \fwbest{48.75} & 48.96 & \fwsecond{0.26} & \fwsecond{53.98} & \fwsecond{53.48} & \fwsecond{0.22} & \fwbest{52.74} & \fwbest{52.60} & \fwbest{0.25}\\
Gemini-2.5-Flash~\citep{comanici2025gemini25} & 49.43 & 49.87 & 0.40 & 40.86 & 40.63 & 0.39 & \fwbest{54.24} & \fwbest{54.46} & 0.32 & \fwsecond{48.17} & \fwsecond{48.32} & 0.37\\
Claude-Haiku-4.5~\citep{anthropic2025haiku45} & 47.86 & 48.26 & 0.34 & 43.45 & 43.89 & \fwbest{0.26} & 50.67 & 50.92 & \fwbest{0.20} & 47.33 & 47.69 & \fwsecond{0.27}\\
Qwen3.5-Plus-02-15~\citep{qwen2026qwen35} & 49.74 & 50.16 & 0.37 & 39.24 & 39.38 & 0.41 & 46.39 & 48.07 & 0.37 & 45.12 & 45.87 & 0.38\\
\bottomrule
\end{tabular}}
\end{table}

Table~\ref{tab:rwa_avg_fire_axis} summarizes fire-world task performance on the real-world-aligned subset. Performance remains markedly dependent on the observation track: GPT-5.6-Luna attains the highest average predictive scores on both structured records (62.88 Acc / 62.52 F1) and physics-rendered visual evidence (52.74 Acc / 52.60 F1). For the four frontier closed models assessed under both tracks, structured records yield an average gain of 10.06 Acc and 10.26 F1. The resulting gap shows that converting visualized multi-field observations into dependable fire-task decisions remains a central limitation.

\begin{takeawaybox}
\textbf{Takeaway}: Even on relatively simple real-world-aligned events, current models retain clear boundaries in physical-world understanding: they reason more reliably from structured records than from visualized fields, which remain difficult to map into stable latent-state, mechanism, and intervention judgments.
\end{takeawaybox}

\subsection{FireWorldGPT Results}

Table~\ref{tab:fireworldgpt_main} examines whether complex physical world understanding can be internalized rather than elicited only at inference time by comparing Vanilla models with physics-aware prompting, chain-of-thought reasoning, and supervised fine-tuning. Prompting and CoT yield moderate improvements, whereas SFT produces the most consistent gains in accuracy and Macro-F1 across the evaluated foundation models, although confidence calibration remains an independent challenge. These results indicate that FireWorldGPT uses structured supervision over physical evidence and latent states to internalize the mapping from multi-physics observations to physically grounded representations, thereby substantially strengthening the base model’s physical world understanding.

\begin{table*}[t]
\centering
\caption{\textbf{FireWorldGPT results on controlled simulations.}
Rows compare Vanilla, physics-aware prompting, chain-of-thought reasoning, and
supervised fine-tuning across four foundation models.
Colored annotations report changes relative to Vanilla:
green denotes performance improvement and red denotes degradation, while arrows
indicate the numerical direction of change.
Acc. and F1 are higher-is-better, whereas lower Brier is better.}
\label{tab:fireworldgpt_main}
\vspace{2pt}

\scriptsize
\setlength{\tabcolsep}{3.2pt}
\renewcommand{\arraystretch}{1.12}

\resizebox{\textwidth}{!}{%
\begin{tabular}{l*{12}{c}}
\toprule
\multirow{2}{*}{\textbf{Method}}
& \multicolumn{3}{c}{\textbf{Llama-3.1-8B-Instruct}}
& \multicolumn{3}{c}{\textbf{Qwen3-8B}}
& \multicolumn{3}{c}{\textbf{InternVL3-8B}}
& \multicolumn{3}{c}{\textbf{Qwen3-VL-8B}} \\
\cmidrule(lr){2-4}
\cmidrule(lr){5-7}
\cmidrule(lr){8-10}
\cmidrule(lr){11-13}
& Acc & F1 & Brier
& Acc & F1 & Brier
& Acc & F1 & Brier
& Acc & F1 & Brier \\
\midrule

\textbf{Vanilla}
& 31.1 & 32.9 & 0.77
& 33.4 & 35.7 & 0.80
& 25.1 & 26.4 & 0.59
& 29.2 & 29.5 & 0.79 \\

\textbf{+Prompt}
& 33.4 {\scriptsize\textcolor{green!50!black}{$\uparrow$2.3}}
& 34.5 {\scriptsize\textcolor{green!50!black}{$\uparrow$1.6}}
& 0.80 {\scriptsize\textcolor{red}{$\uparrow$0.03}}
& 35.5 {\scriptsize\textcolor{green!50!black}{$\uparrow$2.1}}
& 37.0 {\scriptsize\textcolor{green!50!black}{$\uparrow$1.3}}
& 0.82 {\scriptsize\textcolor{red}{$\uparrow$0.02}}
& 29.4 {\scriptsize\textcolor{green!50!black}{$\uparrow$4.3}}
& 29.4 {\scriptsize\textcolor{green!50!black}{$\uparrow$3.0}}
& 0.56 {\scriptsize\textcolor{green!50!black}{$\downarrow$0.03}}
& 34.5 {\scriptsize\textcolor{green!50!black}{$\uparrow$5.3}}
& 33.7 {\scriptsize\textcolor{green!50!black}{$\uparrow$4.2}}
& \textbf{0.78} {\scriptsize\textcolor{green!50!black}{$\downarrow$0.01}} \\

\textbf{+CoT}
& 32.2 {\scriptsize\textcolor{green!50!black}{$\uparrow$1.1}}
& 33.6 {\scriptsize\textcolor{green!50!black}{$\uparrow$0.7}}
& \textbf{0.70} {\scriptsize\textcolor{green!50!black}{$\downarrow$0.07}}
& 35.7 {\scriptsize\textcolor{green!50!black}{$\uparrow$2.3}}
& 37.1 {\scriptsize\textcolor{green!50!black}{$\uparrow$1.4}}
& \textbf{0.79} {\scriptsize\textcolor{green!50!black}{$\downarrow$0.01}}
& 28.2 {\scriptsize\textcolor{green!50!black}{$\uparrow$3.1}}
& 30.0 {\scriptsize\textcolor{green!50!black}{$\uparrow$3.6}}
& \textbf{0.45} {\scriptsize\textcolor{green!50!black}{$\downarrow$0.14}}
& 31.8 {\scriptsize\textcolor{green!50!black}{$\uparrow$2.6}}
& 33.9 {\scriptsize\textcolor{green!50!black}{$\uparrow$4.4}}
& 0.82 {\scriptsize\textcolor{red}{$\uparrow$0.03}} \\

\textbf{+SFT}
& \textbf{43.1} {\scriptsize\textcolor{green!50!black}{$\uparrow$12.0}}
& \textbf{44.6} {\scriptsize\textcolor{green!50!black}{$\uparrow$11.7}}
& 0.89 {\scriptsize\textcolor{red}{$\uparrow$0.12}}
& \textbf{40.8} {\scriptsize\textcolor{green!50!black}{$\uparrow$7.4}}
& \textbf{40.8} {\scriptsize\textcolor{green!50!black}{$\uparrow$5.1}}
& 0.84 {\scriptsize\textcolor{red}{$\uparrow$0.04}}
& \textbf{34.1} {\scriptsize\textcolor{green!50!black}{$\uparrow$9.0}}
& \textbf{37.3} {\scriptsize\textcolor{green!50!black}{$\uparrow$10.9}}
& 0.87 {\scriptsize\textcolor{red}{$\uparrow$0.28}}
& \textbf{41.0} {\scriptsize\textcolor{green!50!black}{$\uparrow$11.8}}
& \textbf{39.3} {\scriptsize\textcolor{green!50!black}{$\uparrow$9.8}}
& 0.82 {\scriptsize\textcolor{red}{$\uparrow$0.03}} \\

\bottomrule
\end{tabular}}
\end{table*}

\section{In-Depth Analysis}
\label{sec:analysis}

\begin{figure}[!t]
    \centering
    \includegraphics[width=0.96\linewidth]{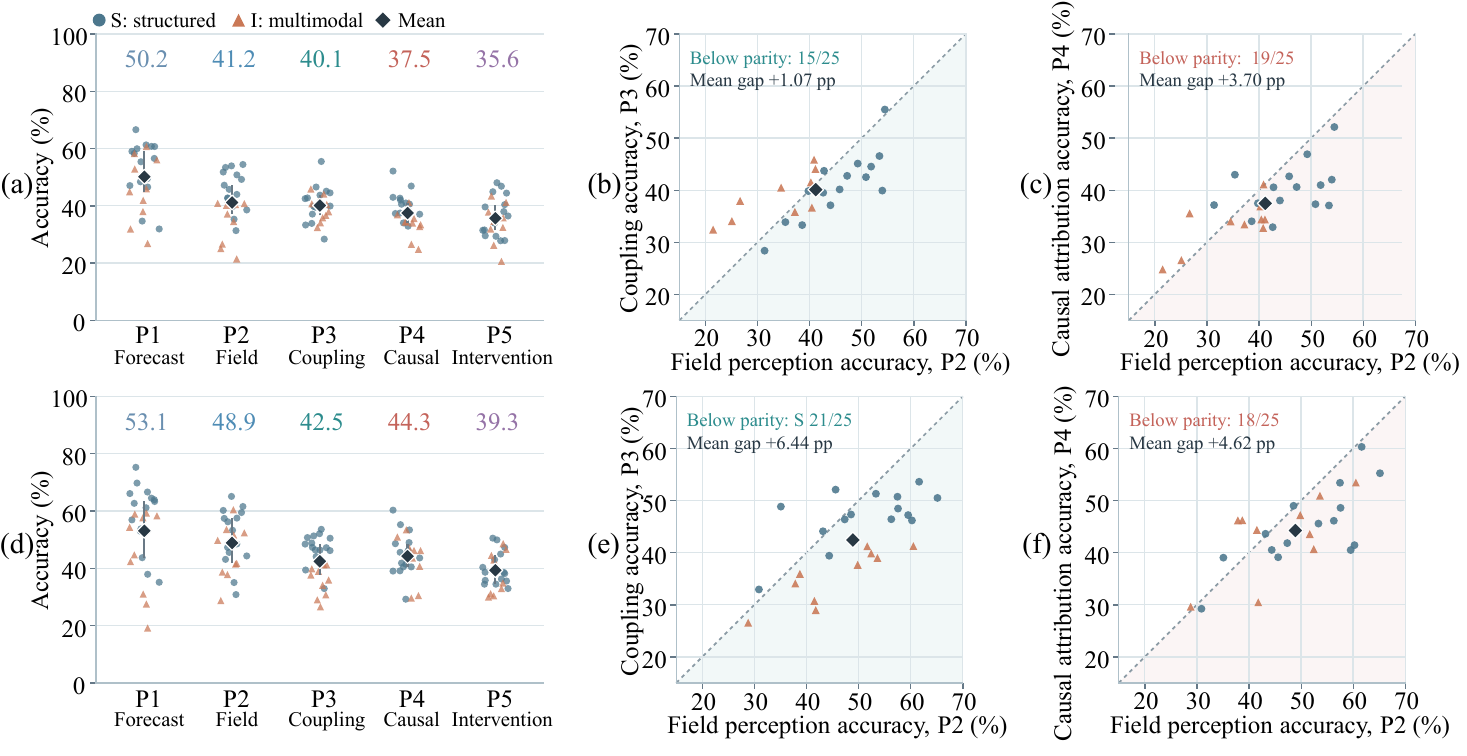}
    \caption{\textbf{Physical capability profiles and perception--mechanism gaps.}
Results on the controlled-simulation (top) and real-world-aligned (bottom) subsets, each comprising 25 equally weighted model--track results.
(a,d) Performance across P1--P5;(b,e) Cross-field coupling and (c,f) mechanism attribution versus field perception.
Mean gaps measure field perception minus coupling or mechanism accuracy in percentage points, circles and triangles denote S-track and I-track results, respectively.}
    \label{fig:physical_capability_gaps}
\end{figure}

\paragraph{Field Perception Outpaces Mechanistic Understanding.}
Figure~\ref{fig:physical_capability_gaps} reports format-averaged accuracy across P1--P5 and contrasts physical field perception (P2) with cross-field coupling understanding (P3) and causal mechanism attribution (P4). Mechanism attribution trails field perception by 3.70 and 4.62 percentage points on the controlled and real-world-aligned subsets, respectively, and falls below field perception in 19/25 and 18/25 model--track results. Cross-field coupling shows a smaller gap on controlled simulations (1.07 points; 15/25 below parity) but a substantially larger gap on real-world-aligned data (6.44 points; 21/25 below parity). These results indicate that recognizing individual field patterns does not reliably translate into coherent cross-field relationships or causal mechanisms, with this limitation becoming more pronounced under real-world-aligned evidence.

\begin{figure}[!t]
    \centering
    \includegraphics[width=0.96\linewidth]{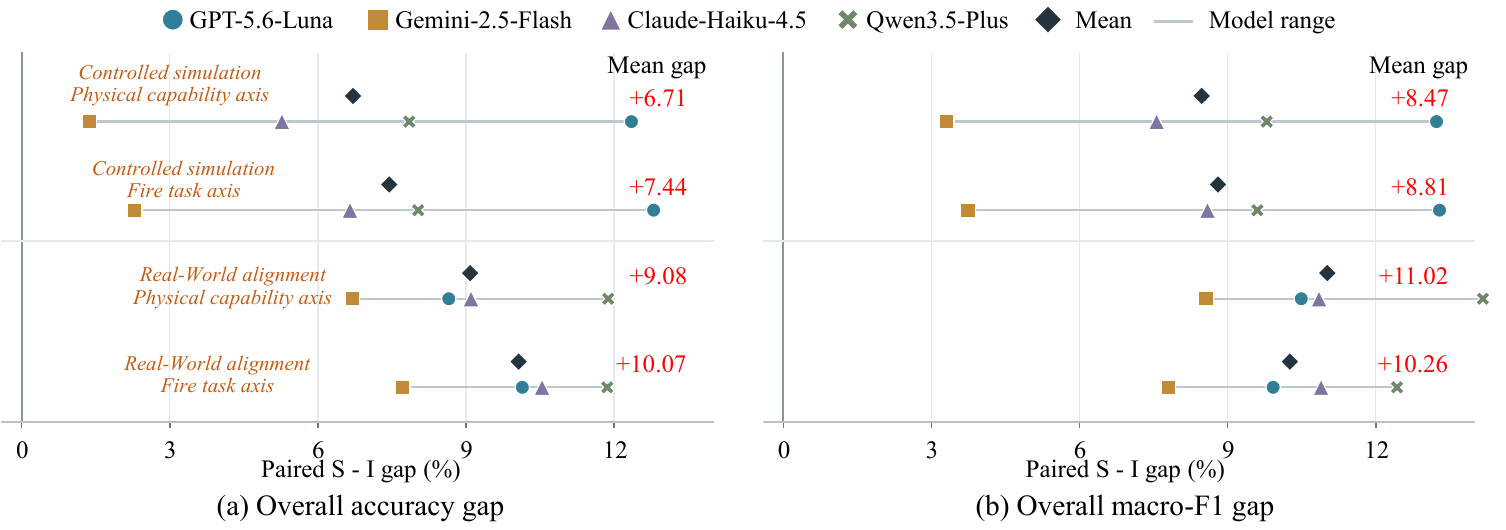}
    \caption{\textbf{Paired performance gaps between structured and multimodal observations.}
(a) Accuracy and (b) Macro-F1 differences across the physical capability and fire-task axes on the controlled and real-world-aligned subsets.
Gaps are computed as S-track minus I-track in percentage points using format-averaged overall scores; positive values favor structured observations.}
    \label{fig:observation_track_gaps}
\end{figure}

\begin{figure}[!t]
    \centering
    \includegraphics[width=0.96\linewidth]{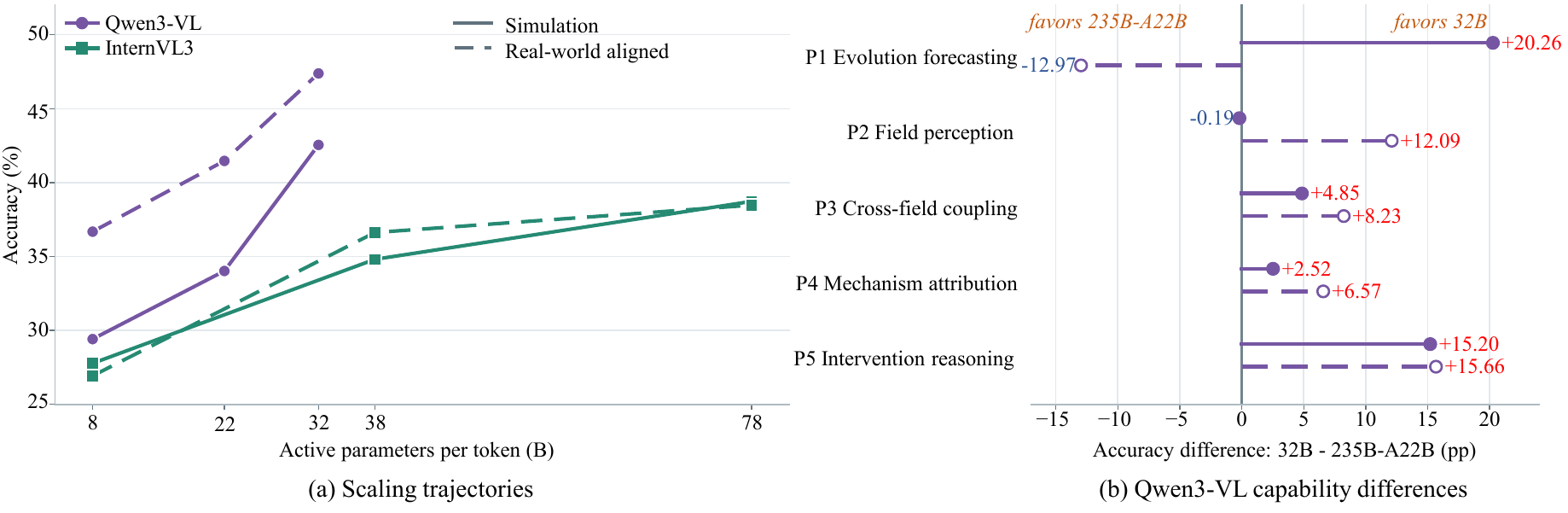}
    \caption{\textbf{Active-parameter scaling yields limited returns in complex physical-world understanding.}
    (a) Format-averaged I-track accuracy across Qwen3-VL and InternVL3 checkpoints, with solid and dashed lines denoting controlled and real-world-aligned settings, respectively.
    (b) Capability-wise differences show a broad advantage of the 32B model over the 22B-active checkpoint.}
    \label{fig:parameter_count}
\end{figure}

\paragraph{A Persistent Gap under Multimodal Physical Understanding.}
Figure~\ref{fig:observation_track_gaps} reports paired S--I performance differences for the same four frontier closed-source models, with choice and open-report scores averaged within each evaluation. Structured observations consistently outperform multimodal physics-rendered evidence across both benchmark axes and both subsets. On controlled simulations, the mean gaps are 6.71--7.44 percentage points in accuracy and 8.47--8.81 in macro-F1; on real-world-aligned data, they increase to 9.08--10.07 and 10.26--11.02, respectively. These results suggest a persistent challenge in interpreting and integrating multimodal physical evidence, although the aggregate comparisons do not isolate visual processing from other differences between observation tracks.

\paragraph{Scaling Without Multi-Physics Supervision Is Insufficient.}

Figure~\ref{fig:parameter_count} shows that increasing active parameter count improves aggregate performance, but the marginal benefit becomes limited once models enter the tens-of-billions regime. 
Capability-wise accuracy differences further shows that the 32B model achieves broad but non-uniform gains over the 22B-active checkpoint across the physical capability spectrum.
However, scaling from 38B to 78B yields only marginal improvements, indicating rapidly diminishing returns at larger active parameter scales.
3 frontier closed-source models reported earlier further support this observation.
These results suggest that, \textbf{without targeted training on coupled multi-physics data, model scaling alone provides limited returns for physical-world understanding}; progress likely requires learning signals that explicitly encode physical fields, their interactions, and the mechanisms governing their evolution.

\begin{figure}[!t]
    \centering
    \includegraphics[width=0.96\linewidth]{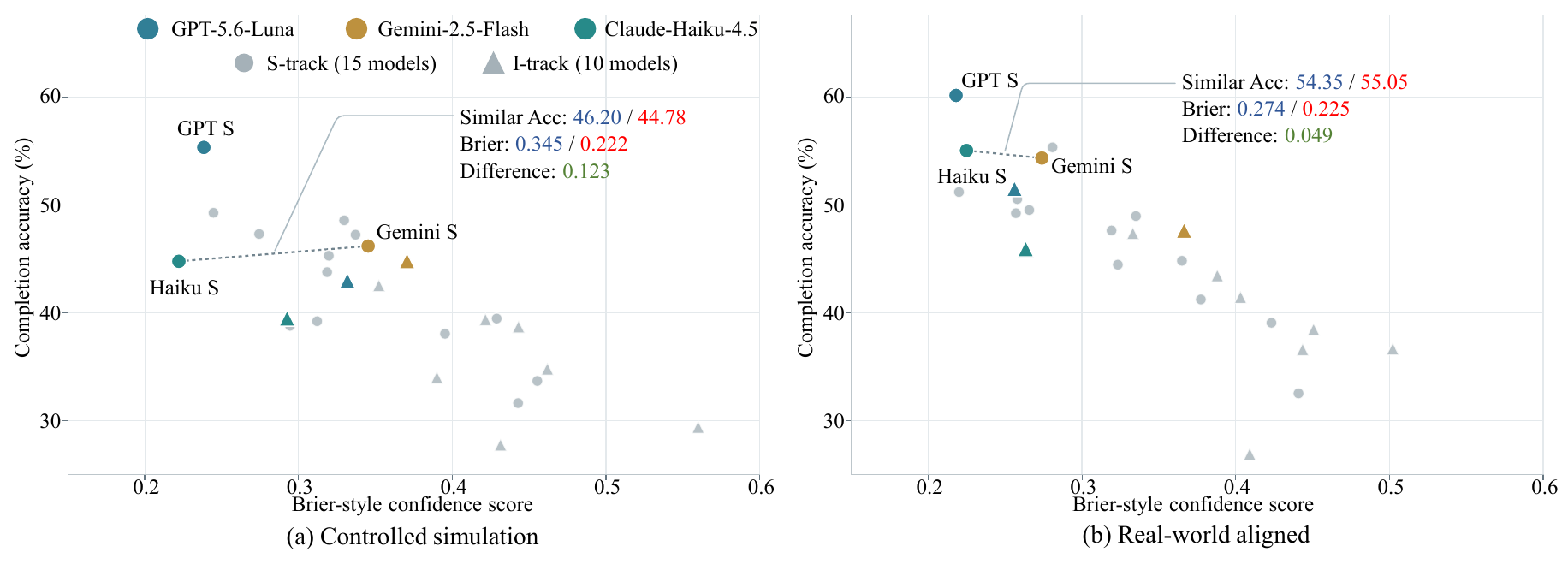}
    \caption{\textbf{Physical correctness and confidence quality.} Completion accuracy versus the benchmark’s Brier-style score on the physical capability axis for (a) controlled simulation and (b) real-world-aligned data. Circles and triangles denote S-track and I-track results. Dashed segments connect illustrative similar-accuracy contrasts.}
    \label{fig:correctness_confidence}
\end{figure}

\paragraph{Accuracy Does Not Determine Confidence Quality.}
Figure~\ref{fig:correctness_confidence} shows that similar physical-answer accuracy can coexist with substantially different confidence errors. On controlled simulations, Gemini-2.5-Flash and Claude-Haiku-4.5 achieve accuracies of 46.20\% and 44.78\%, respectively, while their Brier-style scores differ markedly at 0.345 and 0.222. The pattern persists on real-world-aligned data: their accuracies remain close at 54.35\% and 55.05\%, yet their Brier-style scores are 0.274 and 0.225. Although this quadratic score does not isolate calibration, it reveals variation in confidence--completion consistency that accuracy alone cannot capture. Complex physical-world understanding should therefore be evaluated through both the correctness of physical judgments and the reliability of the confidence assigned to them.

\paragraph{Expert-Validated Questions and Answers.}

\begin{figure}[!t]
    \centering
    \includegraphics[width=0.96\linewidth]{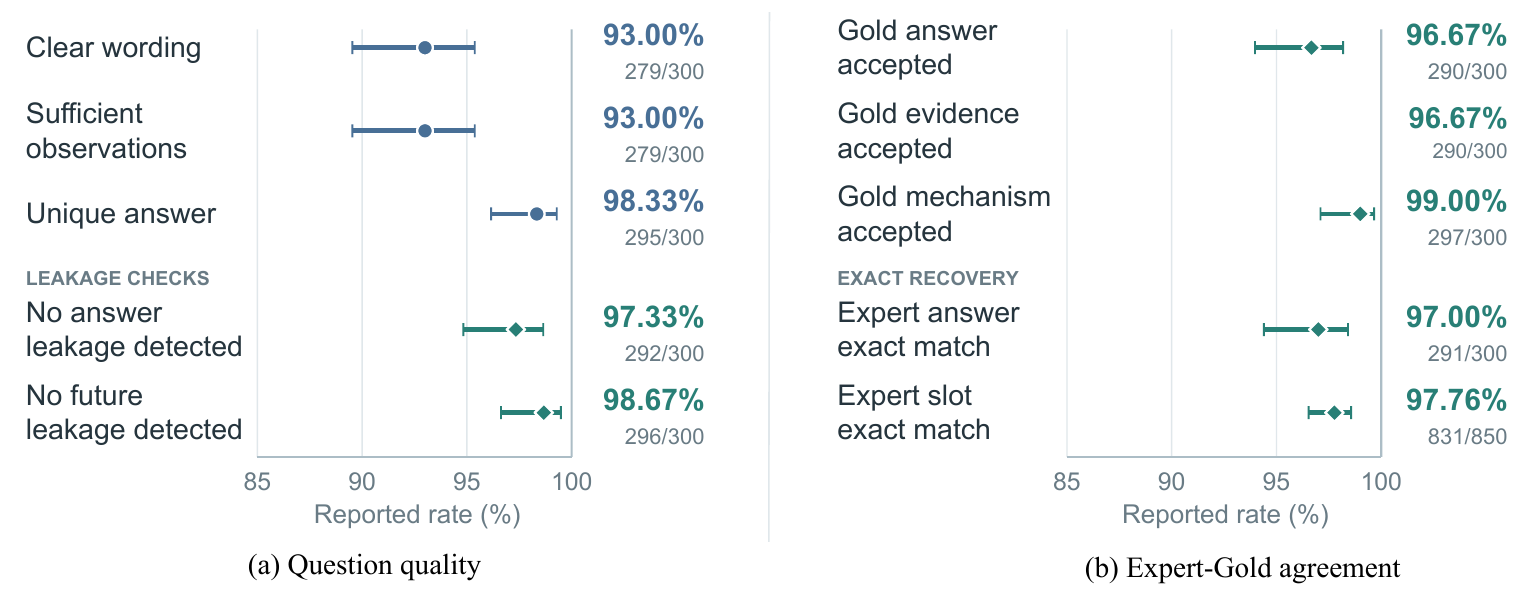}
  \caption{\textbf{Question quality and expert--Gold agreement.}
    (a) Expert-assessed question-quality and leakage-check rates on 300 controlled-simulation QA instances.
    (b) Expert acceptance of Gold answers and mechanisms, alongside exact agreement between expert responses and Gold at the answer and output-slot levels. 
    Circles denote rates from the original audit, while diamonds denote expert-reported recheck rates. Horizontal error bars show nominal 95\% Wilson confidence intervals.}
    \label{fig:expert_audit_statistics}
\end{figure}

Figure~\ref{fig:expert_audit_statistics} provides evidence of task answerability and reference reliability within the audited subset. Question clarity and observation sufficiency each receive 93.0\% fully positive ratings, although rechecks flag answer leakage in eight items and future-information leakage in four. Gold-answer and mechanism acceptance reach 96.67\% and 99.0\%, respectively, while reported expert exact agreement reaches 97.0\% across answers and 97.76\% across output slots. 
These findings indicate that most audited questions are clearly formulated, provide sufficient observations, and exhibit no detected answer or future-information leakage, while strong agreement with expert responses supports the validity and accuracy of the Gold answers.

\FloatBarrier

\section{Conclusion}
\label{sec:conclusion}

We introduced FireWorldBench, a benchmark for complex physical-world intelligence that uses coupled-field fire dynamics as a canonical stress test. By combining controlled simulations and real-world-aligned events, structured records and multi-physics visualizations, and matched choice and open-report interfaces, FireWorldBench evaluates models along complementary physical-capability and fire-task axes. Our results expose persistent capability boundaries: current models struggle to transform observable fields into coupled states and causal mechanisms, remain sensitive to observation modality, and exhibit limitations in explanation, confidence calibration, and parameter scaling. Controlled enhancement studies further show that supervised adaptation is most effective, enabling FireWorldGPT to internalize the mapping from multi-physics observations to physically grounded states and mechanisms. FireWorldBench thus provides a reproducible and capability-resolved foundation for measuring and advancing multimodal models toward reliable understanding of complex physical worlds.

\FloatBarrier
\clearpage
\begingroup
\footnotesize
\setlength{\bibsep}{2pt plus 0.2ex}
\bibliography{custom}
\endgroup

\clearpage
\appendix

\thispagestyle{plain}

\begingroup
\makeatletter

\footnotesize
\setlength{\parindent}{0pt}
\setcounter{tocdepth}{1}

\newcommand*\l@apsection[2]{%
  \begingroup
  \color{trainmark}%
  \bfseries
  \@dottedtocline{0}{0em}{2.4em}{#1}{#2}%
  \endgroup
}

\newcommand*\l@apsubsection[2]{%
  \begingroup
  \color{trainmark}%
  \normalfont
  \@dottedtocline{1}{1.6em}{3.4em}{#1}{#2}%
  \endgroup
}

\noindent
{\LARGE\bfseries Appendix Contents}

\vspace{0.8em}

\@starttoc{apc}

\makeatother
\endgroup

\clearpage

\let\FWBOriginalSection\section
\let\FWBOriginalSubsection\subsection

\renewcommand{\section}[1]{%
  \FWBOriginalSection{#1}%
  \addcontentsline{apc}{apsection}%
    {\protect\numberline{\thesection}#1}%
}

\renewcommand{\subsection}[1]{%
  \FWBOriginalSubsection{#1}%
  \addcontentsline{apc}{apsubsection}%
    {\protect\numberline{\thesubsection}#1}%
}


\section{Benchmark Construction Details}
\label{app:construction}

\subsection{Construction Scope and Provenance}

FireWorldBench is constructed through two provenance-preserving branches that share a common representation and question-generation protocol, as summarized in Figure~\ref{fig:construction_pipeline}. The controlled branch contains 494 designed simulation worlds, while the real-world-aligned branch contains 26 event groups reconstructed from observed experimental fires. Together, they cover 47 scene archetypes across the seven environment families summarized in Figure~\ref{fig:dataset} and yield 9,074 question--answer instances: 8,360 from controlled simulations and 714 from real-world-aligned events. The two branches differ in how their initial conditions are obtained, but converge on the same construction invariant: every released instance is traceable to a versioned fire-world specification, a retained physical state, an explicitly delimited observation, and a deterministic Gold target. This invariant allows the benchmark to compare controlled and observed settings without conflating recorded evidence with simulation-completed latent state.

Each instance is annotated along the two complementary axes introduced in Section~\ref{sec:fireworldbench}. The physical capability axis comprises Temporal Evolution Forecasting (P1), Physical Field Perception and Grounding (P2), Cross-Field Coupling Understanding (P3), Causal Mechanism Attribution (P4), and Counterfactual Intervention Reasoning (P5). The fire scenario task axis comprises Localized Onset (T1), Coupled Propagation (T2), and Critical Transition (T3). These labels are assigned to the same logical target and provide two cross-cutting views of the resulting dataset rather than two independently generated collections.

\subsection{Controlled Fire-World Specification}

Each controlled world begins from a structured, versioned specification that defines scene geometry, materials, ignition source and location, heat-release-rate behavior, openings, ventilation, obstacles, sensors, cameras, event timing, and, where applicable, a designated intervention variable. The 47 scene archetypes instantiate the seven environment families through physically meaningful variation in confinement, connectivity, combustible loading, source placement, ambient or mechanical flow, and propagation conditions. Within an archetype, alternative topologies, source configurations, and ventilation regimes produce distinct physical systems rather than cosmetic reskins of a fixed scene.

Objects that provide semantic evidence are registered to their physical counterparts in the simulation. Furniture, vehicles, equipment, storage racks, vegetation, and structural elements therefore enter the solver as flow obstacles, thermal boundaries, radiative occluders, combustible components, or ventilation and control boundaries as appropriate; assets that are purely decorative are excluded from benchmark evidence. This binding prevents the visual scene from implying physical structure that is absent from the underlying world state.

Counterfactual worlds are generated from a shared base specification. A paired intervention preserves the geometry, source world, observation configuration, and all nondesignated conditions while changing only the declared control variable, such as an opening, exhaust setting, or suppression condition. Differences between the resulting trajectories can therefore be attributed to the intervention rather than to an uncontrolled scene change.

\subsection{Real-World-Aligned Event Reconstruction}

The real-world-aligned branch begins from selected sequences in MmodalFire, a collection of 65 indoor-fire videos synchronized with six types of physical sensing data~\citep{jia2026mmodalfire}. A multimodal information encoder synchronizes the recorded streams and converts the observed chronology, scene configuration, and sensor evidence into event-level constraints for an aligned Fire Dynamics Simulator (FDS) reconstruction. The recorded frames and measurements remain the source of observed evidence; the aligned reconstruction supplies the dense spatial fields and intervention-sensitive state needed to instantiate the same diagnostic, forecasting, and intervention tasks used for controlled worlds.

We preserve this provenance distinction throughout construction. Every retained asset is marked as either observed or simulation-completed, and reconstructed fields are not presented as direct measurements. The observed sequence anchors event timing and appearance, whereas the FDS realization provides a co-registered latent representation of temperature, soot density, velocity, visibility, streamwise flow, and related fields, as illustrated in Figure~\ref{fig:assets}. This separation is essential to real-world alignment: it permits physically structured evaluation around recorded events without treating simulator-derived quantities as independently observed ground truth.

\subsection{Simulation and Multimodal Observation Construction}

Designed and aligned specifications are executed with FDS~\citep{mcgrattan2026fds,overholt2014validation} to resolve the spatiotemporal evolution of the coupled fire state. The construction targets a nominal mesh spacing of $0.5\,\mathrm{m}$ or finer, with local refinement around ignition regions and narrow openings when required by geometric resolution or numerical stability. Because the scene families differ substantially in scale and topology, the exact mesh, solver configuration, event duration, and output cadence are stored with each world rather than imposed as a single global setting. Failed, unstable, or incomplete executions are rejected before any question is generated.

Accepted solver outputs are converted into three co-registered representations: structured textual observations, two-dimensional physical-field visualizations, and three-dimensional event-level scene assets. NIST Smokeview produces the time-indexed three-dimensional views, while an observation extractor derives sensor traces, region-level summaries, and two-dimensional fields from the same retained solver state. Shared world coordinates, region identifiers, and timestamps align all representations. Rendering manifests fix the camera, color mapping, temporal sampling, obstacle mask, and source assets; an accompanying sidecar records the configuration required to reconstruct each visual observation.

The benchmark exposes these representations through two observation tracks. The S-track provides only task-authorized sensor traces, field summaries, spatial regions, and timestamps. The I-track provides the corresponding physics-rendered visual evidence and neutral scene context. The two tracks are generated from the same underlying state wherever both are available, but their evidence contracts remain separate: S-track serialization cannot state the target label, and I-track images cannot expose privileged identifiers, filenames, captions, or hidden geometry. Task-specific temporal cutoffs further determine which timestamps are observable. Forecasting instances reveal no post-cutoff state, state-reconstruction instances withhold the queried variables, and intervention instances expose only the factual or counterfactual branch permitted by the task contract.

\subsection{Gold-First Logical-Item Construction}

Question construction proceeds from the retained physical state to the target and only then to the observable prompt. For each eligible world and cutoff, deterministic task rules first derive a complete, normalized prediction from the solver record. This private construction record includes the required answer fields, the physical evidence supporting them, and the corresponding mechanism or intervention anchors. The builder then selects only task-admissible observations and assigns the resulting logical item to one physical-capability label and one fire-task label. This target-first order prevents answers from being inferred from rendering artifacts, scene names, or manually written descriptions.

Each track-conditioned logical item is instantiated once in a choice format and once as an open report, with the world, observation, temporal cutoff, target, and split held fixed. Choice questions may require either one option or a set of options. Their alternatives encode competing but physically plausible hypotheses at the same semantic granularity as the Gold answer; option positions, wording, and lengths are audited for template and frequency shortcuts. Open reports expose no candidates and require a structured \texttt{prediction} together with a concise \texttt{conclusion}, supporting \texttt{evidence}, a causal \texttt{mechanism}, and \texttt{confidence}. Both interfaces are therefore grounded in the same physical target while testing recognition and explicit physical commitment under different answer constraints.

\subsection{Quality Control and Split Isolation}

Quality control is applied before an item enters the released benchmark. At the world level, we reject incomplete runs, unstable numerical outputs, missing source fields, and construction records that cannot be traced to their specifications. At the item level, we remove cases with insufficient admissible evidence, tied or nonunique physical targets, nonidentifiable intervention effects, or visual observations that do not support the requested inference. Automated consistency checks verify field and timestamp alignment, target derivation, answer uniqueness, option validity, open-report schema completeness, and agreement between the paired answer interfaces. When one member of an answer-format pair fails validation, the complete pair is removed so that format comparisons retain a matched population.

Splits are frozen at the event-group level rather than at the rendered-instance level. All temporal windows, observation variants, answer formats, and counterfactual siblings derived from the same source world or event group remain in the same partition. This prevents a model from encountering the same base dynamics or an alternative intervention branch across training and test sets. Difficulty and template checks use only training-side event groups; revisions may alter question sampling or presentation but never the retained FDS state or physical target used by the sealed evaluation set.

\subsection{Traceability and Reproducibility}

Every world and instance carries persistent identifiers that link the released question to its source specification and construction record. For controlled worlds, the retained bundle includes the FDS input, execution log, device traces, event records, planar fields, three-dimensional outputs, rendering metadata, and quality-control status. For real-world-aligned events, the manifest additionally records the observed source sequence, synchronization metadata, event constraints, and the boundary between observed and simulation-completed assets. Instance-level records bind the world identifier, split, axis labels, task contract, observation track, temporal cutoff, answer format, Gold fields, and evidence provenance.

The construction release freezes builder versions, asset inventories, configuration files, and hashes of core artifacts. The rebuild procedure begins from the versioned specifications, regenerates solver-derived observations and question records, and compares the resulting manifests and core data files with the frozen release. Because Gold targets are derived from retained physical states and evaluation is performed offline, the benchmark remains auditable at the level of worlds, observations, logical items, and final scores. Detailed metric definitions and invalid-response handling are provided separately in Appendix~\ref{app:protocol}.
\section{Evaluation Protocol and Metrics}
\label{app:protocol}

\subsection{Deterministic Scoring Protocol}

All FireWorldBench metrics are computed by a frozen deterministic evaluator, without an LLM judge. Let $\mathcal{D}$ denote an evaluation set and $i\in\mathcal{D}$ an item. Each item has a response type $q_i\in\{\mathrm{choice},\mathrm{open}\}$, a gold target $y_i$, a model prediction $\hat{y}_i$, and an optional confidence $c_i\in[0,1]$. A choice-format response specifies a set of selected options. An open report contains a structured \texttt{prediction} object together with \texttt{conclusion}, \texttt{evidence}, \texttt{mechanism}, and \texttt{confidence}. When the Gold record provides \texttt{answer\_fields}, only those required paths are scored; otherwise, all recursively flattened Gold paths under \texttt{prediction} are used. Paths outside this task-specific scoring schema are ignored.

Importantly, the structured fields used by the evaluator are not arbitrary formatting tokens introduced for convenient string matching. They are determined during Gold-first construction from the retained physical state and the task contract, before model responses are observed. Required fields encode the physical commitments that are necessary to solve the corresponding task, such as the relevant spatial region, physical variable, temporal trend, causal driver, risk state, prediction target, intervention outcome, or forecast horizon. Thus, matching a required field means recovering a task-defined physical state or relation under a frozen output ontology. Schema compliance alone receives no correctness credit: once a response is parseable, its score is determined by whether these physically meaningful target values are recovered.

The evaluator first normalizes the response and then applies the same parsing and scoring rules to every model. Unparseable responses, missing required content, illegal choice identifiers, and responses whose status is not \texttt{ok} receive deterministic failure scores rather than being repaired, re-prompted, or judged semantically. We report six metrics: Completion Accuracy (Acc), Macro-F1 (F1), Brier score, Evidence-F1 (Evi-F1), Mechanism Alignment (Mech), and Gold-Linked Support (GLS). Acc, F1, and Brier apply to both response interfaces; Evi-F1, Mech, and GLS apply only to open reports. Higher is better for all metrics except Brier.

Acc and Macro-F1 measure whether the required physical target is recovered; Evi-F1 examines whether the model identifies the physical regions and variables that support that target; Mech examines whether the generated explanation recovers the annotated mechanism content; GLS explicitly couples target correctness with Gold-linked physical support; and Brier measures whether confidence is consistent with complete physical correctness. No single metric is interpreted as sufficient evidence of physical understanding in isolation. Their joint use is designed to distinguish a plausible final answer from an answer that is physically grounded, mechanistically supported, and reliably expressed.

Throughout this appendix, $\mathbb{1}[\cdot]$ denotes the indicator function. Any metric ratio with a zero denominator is defined as zero unless stated otherwise.

\subsection{Completion Accuracy}

Completion Accuracy measures the fraction of required answer content recovered by the prediction and permits partial credit. For a choice-format item, let $\mathcal{C}_i$ and $\widehat{\mathcal{C}}_i$ denote the Gold and predicted option sets. The item-level completion ratio is their Jaccard overlap:

\begin{equation}
r_i^{\mathrm{choice}}
=
\frac{|\widehat{\mathcal{C}}_i\cap\mathcal{C}_i|}
{|\widehat{\mathcal{C}}_i\cup\mathcal{C}_i|}.
\label{eq:choice-completion}
\end{equation}

Choice identifiers are normalized to uppercase letters. Duplicate or nonalphabetic identifiers invalidate the response, and an empty union yields zero. This formulation penalizes both omitted and extraneous selections without rewarding the large number of unselected false options.

For the choice interface, the selected option set represents a commitment among physically competing hypotheses constructed at the same semantic granularity. Completion therefore measures how much of the required physical conclusion is recovered while penalizing both missing and unsupported physical claims. It is not computed from superficial overlap with the question text.

For an open report, let $\mathcal{S}_i$ denote the set of required flattened output slots, and let $y_{is}$ and $\hat{y}_{is}$ be the Gold and predicted values for slot $s$. A slot is correct only when it is present and its value exactly matches the Gold value under the evaluator's frozen normalization:

\begin{equation}
r_i^{\mathrm{open}}
=
\frac{1}{|\mathcal{S}_i|}
\sum_{s\in\mathcal{S}_i}
\mathbb{1}
[s\in\widehat{\mathcal{S}}_i
\land
\hat{y}_{is}=y_{is}].
\label{eq:open-completion}
\end{equation}

Here, $\widehat{\mathcal{S}}_i$ is the set of predicted flattened paths. A missing required slot is incorrect. Nested outputs, such as horizon-specific temperature or visibility trends, are scored at their flattened paths. Composite slot values are canonicalized and then treated as atomic values for exact comparison.

The exact slot comparison should therefore be interpreted as an exact comparison of \emph{physical commitments under a controlled ontology}, rather than as a generic natural-language exact-match metric. For example, correctly identifying a region while assigning the wrong physical trend, causal driver, intervention consequence, or temporal horizon does not constitute the same physical state and is intentionally scored as incorrect. Conversely, linguistic material outside the required prediction paths does not improve Completion Accuracy. This design makes the structured report a deterministic test of whether the model has recovered the physical quantities and relations required by the task, while leaving explanation quality to the complementary metrics defined below.

The interface-appropriate ratio is denoted by $r_i$:

\begin{equation}
r_i
=
\mathbb{1}[q_i=\mathrm{choice}]r_i^{\mathrm{choice}}
+
\mathbb{1}[q_i=\mathrm{open}]r_i^{\mathrm{open}}.
\label{eq:completion-ratio}
\end{equation}

Completion Accuracy is the item average:

\begin{equation}
\mathrm{Acc}(\mathcal{D})
=
\frac{1}{|\mathcal{D}|}
\sum_{i\in\mathcal{D}}r_i.
\label{eq:completion-accuracy}
\end{equation}

Acc is therefore not whole-item exact match. Exact completion is used only as the binary event underlying the Brier score.

Accordingly, Acc answers the first and most direct evaluation question in FireWorldBench: \emph{does the model recover the physical state, outcome, or intervention consequence that the available observations support?} It deliberately permits partial credit because many benchmark targets contain multiple jointly required physical commitments, while the subsequent metrics determine whether those commitments are balanced across target types and grounded in the appropriate evidence and mechanisms.

\subsection{Interface-Specific Macro-F1}

Macro-F1 complements item-averaged Acc, but its construction follows the semantics of each response interface. Choice-format F1 balances complete answer-set outcomes, whereas open-report F1 balances required structured output slots. The F1 columns in all tables use the corresponding definition below.

The role of Macro-F1 is distinct from Acc. Physical-world tasks contain heterogeneous states and outcomes whose empirical frequencies need not be balanced. A model that repeatedly predicts a common fire state, dominant trend, or frequent answer structure may obtain nontrivial item-averaged completion without demonstrating uniformly reliable physical reasoning. Macro-F1 therefore asks whether performance remains balanced across distinct physical answer outcomes or structured physical fields, reducing the extent to which frequent or easier outcomes dominate the reported score.

\paragraph{Choice-format Macro-F1.}
For each choice item, the sorted Gold option set is serialized as one categorical label $u_i$, and the predicted set is serialized as $\hat{u}_i$. Thus, two option sets that differ by even one option are distinct labels. Let $\mathcal{L}_{\mathcal{D}}$ be the union of Gold and predicted exact-set labels in the choice-format evaluation cell. For each label $\ell\in\mathcal{L}_{\mathcal{D}}$, the confusion counts are

\begin{equation}
\mathrm{TP}_{\ell}
=
\sum_{i\in\mathcal{D}}
\mathbb{1}[u_i=\ell\land\hat{u}_i=\ell].
\label{eq:choice-tp}
\end{equation}

\begin{equation}
\mathrm{FP}_{\ell}
=
\sum_{i\in\mathcal{D}}
\mathbb{1}[u_i\neq\ell\land\hat{u}_i=\ell].
\label{eq:choice-fp}
\end{equation}

\begin{equation}
\mathrm{FN}_{\ell}
=
\sum_{i\in\mathcal{D}}
\mathbb{1}[u_i=\ell\land\hat{u}_i\neq\ell].
\label{eq:choice-fn}
\end{equation}

The F1 score for an exact-set label is

\begin{equation}
F1_{\ell}
=
\frac{2\mathrm{TP}_{\ell}}
{2\mathrm{TP}_{\ell}+\mathrm{FP}_{\ell}+\mathrm{FN}_{\ell}}.
\label{eq:choice-label-f1}
\end{equation}

Choice-format Macro-F1 gives equal weight to all exact-set labels:

\begin{equation}
\mathrm{MacroF1}_{\mathrm{choice}}(\mathcal{D})
=
\frac{1}{|\mathcal{L}_{\mathcal{D}}|}
\sum_{\ell\in\mathcal{L}_{\mathcal{D}}}F1_{\ell}.
\label{eq:choice-macro-f1}
\end{equation}

A failed or missing choice response is represented by a dedicated missing prediction label, ensuring that it contributes a false negative for the Gold exact-set class rather than disappearing from the computation.

\paragraph{Open-report Slot-Macro-F1.}
For open reports, Macro-F1 is computed from exact $(\text{slot},\text{value})$ matches and then macro-averaged over structured output slots. Let $\mathcal{S}_{\mathcal{D}}$ be the union of required flattened slots in an open-report evaluation cell, and let $\mathcal{I}_s$ contain the items for which slot $s$ is required. For each slot, an exactly matched value contributes one true positive:

\begin{equation}
\mathrm{TP}_{s}
=
\sum_{i\in\mathcal{I}_s}
\mathbb{1}
[s\in\widehat{\mathcal{S}}_i
\land
\hat{y}_{is}=y_{is}].
\label{eq:slot-tp}
\end{equation}

A present but incorrect value contributes one false positive:

\begin{equation}
\mathrm{FP}_{s}
=
\sum_{i\in\mathcal{I}_s}
\mathbb{1}
[s\in\widehat{\mathcal{S}}_i
\land
\hat{y}_{is}\neq y_{is}].
\label{eq:slot-fp}
\end{equation}

The same incorrect value also misses the Gold value and therefore contributes one false negative; an omitted required slot contributes one false negative but no false positive:

\begin{equation}
\mathrm{FN}_{s}
=
\sum_{i\in\mathcal{I}_s}
\mathbb{1}
[s\notin\widehat{\mathcal{S}}_i
\lor
(s\in\widehat{\mathcal{S}}_i
\land
\hat{y}_{is}\neq y_{is})].
\label{eq:slot-fn}
\end{equation}

The slot-level exact-match F1 is

\begin{equation}
F1_s
=
\frac{2\mathrm{TP}_{s}}
{2\mathrm{TP}_{s}+\mathrm{FP}_{s}+\mathrm{FN}_{s}}.
\label{eq:slot-f1}
\end{equation}

The reported open-report F1 is the unweighted mean over scored slots:

\begin{equation}
\mathrm{Slot\text{-}MacroF1}(\mathcal{D})
=
\frac{1}{|\mathcal{S}_{\mathcal{D}}|}
\sum_{s\in\mathcal{S}_{\mathcal{D}}}
\frac{2\mathrm{TP}_{s}}
{2\mathrm{TP}_{s}+\mathrm{FP}_{s}+\mathrm{FN}_{s}}.
\label{eq:macro-f1}
\end{equation}

This definition assigns equal weight to heterogeneous slots such as \texttt{location}, \texttt{trend}, \texttt{driver}, and \texttt{risk}, irrespective of how frequently each slot occurs or how many categorical values it can take. In particular, it does not macro-average the raw value labels within a slot. Extra paths outside the task-specific scored slot set do not affect this metric.

This equal weighting is particularly important for evaluating physical-world understanding. The scored slots correspond to qualitatively different aspects of a physical state: identifying \emph{where} an event occurs is not interchangeable with identifying \emph{how} a field evolves, \emph{what} drives the change, or \emph{which} region becomes hazardous. Slot-Macro-F1 prevents a model from compensating for systematic failure on one such physical component by performing well on a more frequent component. Together, Acc and Macro-F1 therefore measure both the amount of physical target content recovered and the breadth with which that recovery extends across heterogeneous physical commitments.

\subsection{Evidence F1}

Evidence F1 is a deterministic proxy for whether an open report identifies the Gold-linked spatial regions and physical variables supporting its prediction.

The motivation is that a physically correct conclusion should be grounded in the observations that make that conclusion identifiable. In FireWorldBench, Gold evidence anchors are derived from the same retained physical state used to construct the target. They encode which spatial region and which physical variable provide task-relevant support, rather than generic keywords expected to appear in a fluent fire description. Consequently, Evidence F1 asks whether the model attends to the physically relevant part of the world state rather than merely producing a plausible final answer.

Let $t_i$ be the lowercased concatenation of the generated \texttt{conclusion}, \texttt{evidence}, and \texttt{mechanism} fields:

\begin{equation}
t_i
=
\operatorname{lower}
\left(
\operatorname{concat}
(\texttt{conclusion}_i,\texttt{evidence}_i,\texttt{mechanism}_i)
\right).
\label{eq:evidence-text}
\end{equation}

Each Gold \texttt{evidence\_claim} yields an anchor $a=(r_a,v_a)$, where $r_a$ is its lowercased region identifier and $v_a$ is the first physical variable matched from the frozen vocabulary

\begin{equation}
\mathcal{V}
=
\{
\texttt{temperature},
\texttt{visibility},
\texttt{soot},
\texttt{flow},
\texttt{velocity},
\texttt{risk},
\texttt{fire},
\texttt{smoke},
\texttt{sensor},
\texttt{ventilation}
\}.
\label{eq:evidence-vocabulary}
\end{equation}

Let $\mathcal{A}_i$ be the Gold anchor collection and $K(a)$ the set of nonempty components of anchor $a$. Let $\operatorname{hit}(x,t_i)$ denote the evaluator's deterministic match for component $x$. The anchor-level evidence recall is

\begin{equation}
R_i^{\mathrm{Ev}}
=
\frac{1}{\max(1,|\mathcal{A}_i|)}
\sum_{a\in\mathcal{A}_i}
\frac{1}{\max(1,|K(a)|)}
\sum_{x\in K(a)}
\mathbb{1}[\operatorname{hit}(x,t_i)].
\label{eq:evidence-recall}
\end{equation}

Let $\mathcal{M}_i$ contain all case-insensitive region identifiers matching \texttt{R\textbackslash d+} and all vocabulary terms in $\mathcal{V}$ extracted from $t_i$. Let $\mathcal{G}_i$ collect the corresponding nonempty Gold anchor components:

\begin{equation}
\mathcal{G}_i
=
\bigcup_{a\in\mathcal{A}_i}K(a).
\label{eq:evidence-gold-components}
\end{equation}

Evidence precision penalizes unsupported region or variable mentions:

\begin{equation}
P_i^{\mathrm{Ev}}
=
\frac{|\mathcal{M}_i\cap\mathcal{G}_i|}
{\max(1,|\mathcal{M}_i|)}.
\label{eq:evidence-precision}
\end{equation}

The item-level Evidence F1 is

\begin{equation}
\mathrm{Ev}_i
=
\frac{2P_i^{\mathrm{Ev}}R_i^{\mathrm{Ev}}}
{P_i^{\mathrm{Ev}}+R_i^{\mathrm{Ev}}}.
\label{eq:evidence-f1}
\end{equation}

Evidence F1 rewards recovery of annotated anchors while penalizing unsupported mentions. It is a rule-based region--variable proxy, not semantic tuple matching or model-based judging; mentioning a correct region and a correct variable does not by itself establish that the generated text relates them correctly.

This limitation is deliberate and defines the role of Evi-F1 within the complete protocol. Evi-F1 is not intended to establish causal reasoning by itself; it tests the preceding requirement that the model has located the physical evidence on which such reasoning should be based. A report that mentions the wrong region or the wrong physical field cannot be considered well grounded even if its final answer happens to be correct. Whether the recovered evidence is accompanied by the appropriate mechanism and actually supports the final physical commitment is evaluated separately by Mech and GLS. Thus, Evi-F1 forms the evidence-grounding stage of the overall evaluation chain.

\subsection{Mechanism Alignment}

Mechanism Alignment measures recall of the normalized lexical content in the Gold causal mechanism.

Whereas Evi-F1 asks \emph{what physical evidence is relevant}, Mechanism Alignment asks whether the explanation recovers the physical process used by the Gold construction to connect that evidence to the target state or outcome. The Gold mechanism statements are attached to the task-specific physical target during benchmark construction and describe physically meaningful processes such as transport, buoyancy-driven motion, ventilation effects, field propagation, accumulation, or intervention-induced changes. The resulting score therefore provides a deterministic indication of whether the generated explanation contains the mechanism concepts required by the reference physical interpretation.

The tokenizer lowercases the text and extracts tokens matching \texttt{[a-z][a-z0-9\_]+}. Let $g_i^{\mathrm{mech}}$ be the concatenation of all Gold \texttt{mechanism\_claims.statement} strings and $m_i$ the generated \texttt{mechanism}. The Gold token set is

\begin{equation}
\mathcal{G}_i^{\mathrm{mech}}
=
\operatorname{toks}(g_i^{\mathrm{mech}})
\setminus
\mathcal{W}.
\label{eq:mechanism-gold}
\end{equation}

The generated token set is

\begin{equation}
\mathcal{M}_i^{\mathrm{mech}}
=
\operatorname{toks}(m_i).
\label{eq:mechanism-prediction}
\end{equation}

The frozen Gold-side stop-word set is

\begin{equation}
\mathcal{W}
=
\{
\texttt{the},
\texttt{and},
\texttt{with},
\texttt{that},
\texttt{this},
\texttt{from},
\texttt{into},
\texttt{first}
\}.
\label{eq:mechanism-stopwords}
\end{equation}

The item score is Gold-token recall:

\begin{equation}
\mathrm{Mech}_i
=
\frac{
|\mathcal{G}_i^{\mathrm{mech}}
\cap
\mathcal{M}_i^{\mathrm{mech}}|
}{
|\mathcal{G}_i^{\mathrm{mech}}|
}.
\label{eq:mechanism-alignment}
\end{equation}

Because Mech is recall-only, additional irrelevant mechanism tokens are not directly penalized. It should therefore be interpreted jointly with Evidence F1 and GLS rather than as an independent measure of semantic or causal faithfulness.

Accordingly, we do not treat lexical mechanism recovery alone as sufficient evidence of causal understanding. Its value comes from its position between evidence grounding and Gold-linked support: Evi-F1 verifies that the report recovers the relevant observable variables and regions, Mech tests whether the annotated physical mechanism is represented in the explanation, and GLS subsequently requires a correct physical prediction to be explicitly supported by task-aware evidence or mechanism anchors. This decomposition makes it possible to distinguish three qualitatively different failures: selecting the wrong evidence, describing the wrong mechanism, and reaching a prediction that is not supported by the physical explanation.

\subsection{Brier Score}

Brier measures agreement between reported confidence and exact whole-item completion. Its binary target is not the partial-credit completion ratio but the event that every required answer component is correct:

\begin{equation}
z_i
=
\mathbb{1}[r_i=1].
\label{eq:brier-target}
\end{equation}
The response schema requires a numeric confidence value $c_i\in[0,1]$ for every scored item, and all responses included in the reported Brier results satisfy this validity requirement. The Brier score is therefore computed over the complete evaluation set:

\begin{equation}
\mathrm{Brier}(\mathcal{D})
=
\frac{1}{|\mathcal{D}|}
\sum_{i\in\mathcal{D}}
(c_i-z_i)^2.
\label{eq:brier}
\end{equation} An open report that correctly recovers eight of nine required slots contributes $8/9$ to Acc but has $z_i=0$; assigning high confidence to that incomplete report is therefore strongly penalized. Lower Brier is better. Because the target itself depends on exact completion, this metric measures confidence--completion consistency rather than correctness-independent calibration.

Confidence quality constitutes a complementary aspect of reliable physical-world reasoning. Two models may recover a similar fraction of physical targets while differing substantially in whether they recognize when their complete physical interpretation is correct. This distinction is particularly relevant for partially observed and intervention-sensitive systems, where an incorrect but highly confident physical judgment is qualitatively different from an uncertain one. Brier is therefore not used as another measure of physical-answer accuracy; instead, it measures whether a model's stated confidence is consistent with the strict event that all required physical commitments for an item have been recovered.

\subsection{Gold-Linked Support}

GLS measures whether correct open-report predictions are explicitly supported by task-aware evidence or mechanism anchors.

GLS provides the final link between answer correctness and physical justification. Acc can establish that a model produced the required target, Evi-F1 can establish that relevant evidence appears in the report, and Mech can establish that reference mechanism content is recovered; none of these conditions alone guarantees that the \emph{correct prediction is supported by the corresponding physical evidence}. GLS therefore evaluates their conjunction at the level of required prediction slots.

For each required slot $s\in\mathcal{S}_i$, define a correctness indicator using the frozen equivalence function:

\begin{equation}
\mathrm{corr}_{is}
=
\mathbb{1}[s\in\widehat{\mathcal{S}}_i]
\mathbb{1}
[\operatorname{equiv}(\hat{y}_{is},y_{is})].
\label{eq:gls-correctness}
\end{equation}

Let $g_i$ be the concatenated Gold anchor text and $\phi_s(y_{is},t_i,g_i)$ the frozen task-aware support predicate. The supported-correctness indicator is

\begin{equation}
\mathrm{supp}_{is}
=
\mathrm{corr}_{is}
\,
\phi_s(y_{is},t_i,g_i).
\label{eq:gls-support}
\end{equation}

The item-level GLS score is

\begin{equation}
\mathrm{GLS}_i
=
\frac{1}{|\mathcal{S}_i|}
\sum_{s\in\mathcal{S}_i}
\mathrm{supp}_{is}.
\label{eq:gls}
\end{equation}

For $\operatorname{equiv}$, numeric values are compared under floating-point equality, whereas nonnumeric values are compared after the evaluator removes nonalphanumeric characters. The predicate $\phi_s$ uses word-boundary matching and frozen aliases, such as mapping \texttt{thermal}, \texttt{heat}, and \texttt{hot} to temperature-related support and mapping \texttt{up} to increasing-trend expressions. It applies the following slot-dependent rules:

\begin{enumerate}[leftmargin=1.6em,itemsep=0.2em,topsep=0.25em]
    \item A region-valued slot, or a path containing \texttt{region}, is supported only if its value occurs in both the Gold anchors and the generated explanation.
    \item A path ending in \texttt{\_trend} requires the explanation to mention both the corresponding variable and a compatible trend term. A nested \texttt{10s}, \texttt{30s}, or \texttt{60s} path additionally requires the matching horizon.
    \item Paths containing \texttt{signal}, \texttt{driver}, \texttt{variable}, or \texttt{target} require the predicted value to occur in both the Gold anchors and the explanation.
    \item A numeric Gold value requires the corresponding number to appear in the explanation, or the explanation to contain \texttt{all regions} or \texttt{across regions}.
    \item All remaining slots require the normalized Gold value to occur in both the Gold anchors and the generated explanation.
\end{enumerate}

These slot-dependent predicates are designed to reflect the semantics of the physical commitment being evaluated. A spatial prediction must be supported in the corresponding region; a temporal prediction must identify both the physical variable and a compatible change direction, together with the relevant horizon when applicable; and a driver or target prediction must be explicitly connected to the corresponding physical quantity. The evaluator therefore does not treat all textual overlap as equivalent support. Instead, the required support condition changes according to the physical role of the predicted field.

The denominator includes all required Gold slots, not only correctly predicted slots. GLS is therefore a joint correctness-and-support rate, rather than support conditioned on correctness.

This conjunction is central to the interpretation of GLS. A fluent explanation containing correct physical terminology cannot obtain GLS credit for an incorrect prediction, and a correct prediction cannot obtain GLS credit when its required physical support is absent. GLS thus operationalizes the benchmark's strongest report-level requirement: the model must not only reach the correct physical commitment, but also expose evidence or mechanism information that is consistent with why that commitment is correct under the Gold physical state.

\subsection{Aggregation and Invalid Responses}

Items are partitioned by benchmark split, axis label, observation track, and response interface before scoring. Acc, interface-specific F1, and Brier are reported for both choice-format and open-report cells. The three explanation-specific metrics are arithmetic means over open-report items only:

\begin{equation}
\mathrm{Ev}(\mathcal{D}_{\mathrm{open}})
=
\frac{1}{|\mathcal{D}_{\mathrm{open}}|}
\sum_{i\in\mathcal{D}_{\mathrm{open}}}
\mathrm{Ev}_i.
\label{eq:evidence-aggregation}
\end{equation}

\begin{equation}
\mathrm{Mech}(\mathcal{D}_{\mathrm{open}})
=
\frac{1}{|\mathcal{D}_{\mathrm{open}}|}
\sum_{i\in\mathcal{D}_{\mathrm{open}}}
\mathrm{Mech}_i.
\label{eq:mechanism-aggregation}
\end{equation}

\begin{equation}
\mathrm{GLS}(\mathcal{D}_{\mathrm{open}})
=
\frac{1}{|\mathcal{D}_{\mathrm{open}}|}
\sum_{i\in\mathcal{D}_{\mathrm{open}}}
\mathrm{GLS}_i.
\label{eq:gls-aggregation}
\end{equation}

The evaluator's \texttt{overall} entry recomputes each metric over all eligible items and is therefore instance-weighted. By contrast, the Average columns in the paper first compute cell-level scores and then assign equal weight to the five physical capability dimensions P1--P5 or the three fire scenario task dimensions T1--T3. For any reported axis containing $K$ cells, this aggregation is

\begin{equation}
\mathrm{AvgAxis}(M)
=
\frac{1}{K}
\sum_{k=1}^{K}
M(\mathcal{D}_k).
\label{eq:axis-aggregation}
\end{equation}

Here, $M$ denotes any reported metric and $\mathcal{D}_k$ the evaluation cell for the $k$th axis category under a fixed split, track, and response interface. This cell-level averaging prevents categories with more instances from dominating the axis summary.

Equal weighting across axis categories is also important for the intended physical interpretation of the benchmark. It prevents a capability or fire-task category containing more generated instances from dominating the overall assessment and ensures that forecasting, grounding, coupling, mechanism attribution, intervention reasoning, and their corresponding operational tasks remain visible as distinct sources of model strength or failure.

Failed or missing predictions receive $r_i=0$. For choice-format F1, a failed response is assigned the missing prediction label. For open-report Slot-Macro-F1, every required Gold slot in a failed response contributes one false negative and no true positive. Failed open reports receive zero for Evidence F1, Mechanism Alignment, and GLS. If multiple responses share a \texttt{qa\_id}, a response with \texttt{status=ok} is retained preferentially. All six metrics are reported separately and are not collapsed into a single composite score.

We deliberately retain the six measurements separately because they diagnose different stages of physical-world understanding rather than different estimators of the same quantity. High Acc or F1 with low Evi-F1 indicates that a model can reach physical answers without reliably recovering the annotated evidence; high evidence recovery with low Mech indicates difficulty connecting observations through the intended physical process; high answer and evidence scores with low GLS indicate that the final commitment is not consistently linked to its required support; and Brier distinguishes answer quality from confidence reliability. The complete protocol therefore evaluates physical-world understanding through a set of mutually constraining observations rather than through final-answer correctness alone.

In this sense, the structured fields and deterministic matching rules serve as an auditable interface between the latent physical state of the benchmark and the model's externally observable commitments. Correct field recovery is a necessary condition for claiming that the model has produced the corresponding task-defined physical judgment, while Evi-F1, Mech, and GLS test progressively stronger requirements on how that judgment is grounded and justified. This layered design allows FireWorldBench to separate recognition, physical-state recovery, evidence grounding, mechanism recovery, evidence-supported reasoning, and confidence reliability while avoiding reliance on an opaque model-based evaluator.

\section{FireWorldBench Additional Results}
\label{app:additional_results}

The main paper summarizes FireWorldBench performance by averaging the
multiple-choice and open-report interfaces within each evaluation cell.
In this section, we provide the corresponding interface-specific results to
make the benchmark behavior fully transparent.
We retain the same model grouping, observation tracks, evaluation metrics, and
axis-wise aggregation protocol as in the main paper, but report
multiple-choice and open-report performance separately.

Throughout this appendix, we follow the canonical taxonomy defined in the main
text.
The physical capability axis progresses from Temporal Evolution Forecasting
(P1), Physical Field Perception and Grounding (P2), Cross-Field Coupling
Understanding (P3), and Causal Mechanism Attribution (P4), to Counterfactual
Intervention Reasoning (P5).
The fire scenario task axis consists of Localized Onset (T1), Coupled
Propagation (T2), and Critical Transition (T3).
All fire-axis results below are ordered according to this canonical T1--T3
definition.

For Acc. and F1, higher values indicate better performance, whereas lower
Brier scores indicate better calibration.
The reported Average columns follow the same protocol as the main paper:
scores are first computed independently for each evaluation dimension and are
then averaged with equal weight across P1--P5 for the physical capability axis
or T1--T3 for the fire scenario task axis.
This prevents dimensions containing more evaluation instances from
disproportionately dominating the axis-level summary.

\newcommand{\AppMLlamaSmall}{Llama-3.1-8B-Instruct~\citep{grattafiori2024llama3}}
\newcommand{\AppMQwenSmall}{Qwen3-8B~\citep{yang2025qwen3}}
\newcommand{\AppMGemmaSmall}{Gemma-3-12B-IT~\citep{gemmateam2025gemma3}}
\newcommand{\AppMGemmaMid}{Gemma-3-27B-IT~\citep{gemmateam2025gemma3}}
\newcommand{\AppMQwenMid}{Qwen3-32B~\citep{yang2025qwen3}}
\newcommand{\AppMLlamaLarge}{Llama-3.3-70B-Instruct~\citep{grattafiori2024llama3}}
\newcommand{\AppMQwenNext}{Qwen3-Next-80B-A3B-Instruct~\citep{qwen2025qwen3next}}
\newcommand{\AppMGLM}{GLM-4.5-Air (106B-A12B)~\citep{zeng2025glm45}}
\newcommand{\AppMQwenHuge}{Qwen3-235B-A22B~\citep{yang2025qwen3}}
\newcommand{\AppMDeepSeek}{DeepSeek-V3.2~\citep{liu2025deepseekv32}}
\newcommand{\AppMKimi}{Kimi-K2-Instruct (1T-A32B)~\citep{kimiteam2025kimik2}}
\newcommand{\AppMGPT}{GPT-5.6-Luna~\citep{openai2025gpt5}}
\newcommand{\AppMGemini}{Gemini-2.5-Flash~\citep{comanici2025gemini25}}
\newcommand{\AppMClaude}{Claude-Haiku-4.5~\citep{anthropic2025haiku45}}
\newcommand{\AppMQwenPlus}{Qwen3.5-Plus-02-15~\citep{qwen2026qwen35}}
\newcommand{\AppMInternSmall}{InternVL3-8B~\citep{zhu2025internvl3}}
\newcommand{\AppMQwenVLSmall}{Qwen3-VL-8B-Instruct~\citep{bai2025qwen3vl}}
\newcommand{\AppMInternMid}{InternVL3-38B~\citep{zhu2025internvl3}}
\newcommand{\AppMQwenVLMid}{Qwen3-VL-32B-Instruct~\citep{bai2025qwen3vl}}
\newcommand{\AppMInternLarge}{InternVL3-78B~\citep{zhu2025internvl3}}
\newcommand{\AppMQwenVLHuge}{Qwen3-VL-235B-A22B-Instruct~\citep{bai2025qwen3vl}}

\newcommand{\AppCSHead}{\rowcolor{ttgroup}\multicolumn{19}{c}{S-track: structured sensor and physical records}\\}
\newcommand{\AppCIHead}{\rowcolor{ttgroup}\multicolumn{19}{c}{I-track: multimodal physics-rendered visual evidence}\\}
\newcommand{\AppCSmall}{\rowcolor{fwgroup}\multicolumn{19}{l}{\textit{Open-weight models: fewer than 20B total parameters}}\\}
\newcommand{\AppCMid}{\rowcolor{fwgroup}\multicolumn{19}{l}{\textit{Open-weight models: 20--50B total parameters}}\\}
\newcommand{\AppCLarge}{\rowcolor{fwgroup}\multicolumn{19}{l}{\textit{Open-weight models: 50--200B total parameters}}\\}
\newcommand{\AppCHuge}{\rowcolor{fwgroup}\multicolumn{19}{l}{\textit{Open-weight models: more than 200B total parameters}}\\}
\newcommand{\AppCClosed}{\rowcolor{fwgroup}\multicolumn{19}{l}{\textit{Closed-source models}}\\}
\newcommand{\AppCISmall}{\rowcolor{fwgroup}\multicolumn{19}{l}{\textit{Open-weight multimodal models: 7--8B}}\\}
\newcommand{\AppCIMid}{\rowcolor{fwgroup}\multicolumn{19}{l}{\textit{Open-weight multimodal models: 20--50B}}\\}
\newcommand{\AppCIHuge}{\rowcolor{fwgroup}\multicolumn{19}{l}{\textit{Open-weight multimodal models: more than 50B total parameters}}\\}
\newcommand{\AppCIClosed}{\rowcolor{fwgroup}\multicolumn{19}{l}{\textit{Closed-source multimodal models}}\\}

\newcommand{\AppCPHeader}[3]{%
\toprule
\multirow{2}{*}{\textbf{Model}}
& \multicolumn{3}{c}{\textbf{P1}} & \multicolumn{3}{c}{\textbf{P2}} & \multicolumn{3}{c}{\textbf{P3}}
& \multicolumn{3}{c}{\textbf{P4}} & \multicolumn{3}{c}{\textbf{P5}} & \multicolumn{3}{c}{\textbf{Average}}\\
\cmidrule(lr){2-4}\cmidrule(lr){5-7}\cmidrule(lr){8-10}\cmidrule(lr){11-13}\cmidrule(lr){14-16}\cmidrule(lr){17-19}
& #1 & #2 & #3 & #1 & #2 & #3 & #1 & #2 & #3 & #1 & #2 & #3 & #1 & #2 & #3 & #1 & #2 & #3\\
\midrule}

\newcommand{\AppCTHeader}[3]{%
\toprule
\multirow{2}{*}{\textbf{Model}}
& \multicolumn{3}{c}{\textbf{T1}} & \multicolumn{3}{c}{\textbf{T2}} & \multicolumn{3}{c}{\textbf{T3}}
& \multicolumn{3}{c}{\textbf{T4}} & \multicolumn{3}{c}{\textbf{T5}} & \multicolumn{3}{c}{\textbf{Average}}\\
\cmidrule(lr){2-4}\cmidrule(lr){5-7}\cmidrule(lr){8-10}\cmidrule(lr){11-13}\cmidrule(lr){14-16}\cmidrule(lr){17-19}
& #1 & #2 & #3 & #1 & #2 & #3 & #1 & #2 & #3 & #1 & #2 & #3 & #1 & #2 & #3 & #1 & #2 & #3\\
\midrule}

\newcommand{\AppCTable}[4]{%
\begin{table}[t]
\centering
\caption{#1}
\label{#2}
\scriptsize
\setlength{\tabcolsep}{2.15pt}
\resizebox{\textwidth}{!}{%
\begin{tabular}{l*{18}{c}}
#3
#4
\bottomrule
\end{tabular}}
\end{table}
\FloatBarrier}

\newcommand{\AppFSHead}{\rowcolor{ttgroup}\multicolumn{13}{c}{S-track: structured sensor and physical records}\\}
\newcommand{\AppFIHead}{\rowcolor{ttgroup}\multicolumn{13}{c}{I-track: multimodal physics-rendered visual evidence}\\}
\newcommand{\AppFSmall}{\rowcolor{fwgroup}\multicolumn{13}{l}{\textit{Open-weight models: fewer than 20B total parameters}}\\}
\newcommand{\AppFMid}{\rowcolor{fwgroup}\multicolumn{13}{l}{\textit{Open-weight models: 20--50B total parameters}}\\}
\newcommand{\AppFLarge}{\rowcolor{fwgroup}\multicolumn{13}{l}{\textit{Open-weight models: 50--200B total parameters}}\\}
\newcommand{\AppFHuge}{\rowcolor{fwgroup}\multicolumn{13}{l}{\textit{Open-weight models: more than 200B total parameters}}\\}
\newcommand{\AppFClosed}{\rowcolor{fwgroup}\multicolumn{13}{l}{\textit{Closed-source models}}\\}
\newcommand{\AppFISmall}{\rowcolor{fwgroup}\multicolumn{13}{l}{\textit{Open-weight multimodal models: 7--8B}}\\}
\newcommand{\AppFIMid}{\rowcolor{fwgroup}\multicolumn{13}{l}{\textit{Open-weight multimodal models: 20--50B}}\\}
\newcommand{\AppFIHuge}{\rowcolor{fwgroup}\multicolumn{13}{l}{\textit{Open-weight multimodal models: more than 50B total parameters}}\\}
\newcommand{\AppFIClosed}{\rowcolor{fwgroup}\multicolumn{13}{l}{\textit{Closed-source multimodal models}}\\}

\newcommand{\AppFTHeader}[3]{%
\toprule
\multirow{2}{*}{\textbf{Model}}
& \multicolumn{3}{c}{\textbf{T1}} & \multicolumn{3}{c}{\textbf{T2}} & \multicolumn{3}{c}{\textbf{T3}}
& \multicolumn{3}{c}{\textbf{Average}}\\
\cmidrule(lr){2-4}\cmidrule(lr){5-7}\cmidrule(lr){8-10}\cmidrule(lr){11-13}
& #1 & #2 & #3 & #1 & #2 & #3 & #1 & #2 & #3 & #1 & #2 & #3\\
\midrule}

\newcommand{\AppFTable}[4]{%
\begin{table}[t]
\centering
\caption{#1}
\label{#2}
\scriptsize
\setlength{\tabcolsep}{2.15pt}
\resizebox{\textwidth}{!}{%
\begin{tabular}{l*{12}{c}}
#3
#4
\bottomrule
\end{tabular}}
\end{table}
\FloatBarrier}

\subsection{Controlled Simulation Results}
\label{app:controlled_simulation_results}

We first decompose performance on the controlled simulation subset.
Tables~\ref{tab:main_choice_physical_axis} and
\ref{tab:main_open_physical_axis} report results along the physical capability
axis, while Tables~\ref{tab:main_choice_fire_axis} and
\ref{tab:main_open_fire_axis} provide the complementary decomposition along
the fire scenario task axis.
For each axis, we report the structured-observation S-track and the
multimodal image-observation I-track separately.
This decomposition exposes whether the aggregate trends reported in the main
paper persist when models must either select among candidate hypotheses or
construct a physical-world report without answer candidates.

\AppCTable{\textbf{Choice-format evaluation along the physical capability axis.} Models are evaluated on the controlled simulation subset under the structured-observation track (S-track) and multimodal image-observation track (I-track). Each dimension reports accuracy (Acc), macro-F1 (F1), and Brier score; Average is the equal-weight mean across P1--P5.}
{tab:main_choice_physical_axis}
{\AppCPHeader{Acc}{F1}{Brier}}
{%
\AppCSHead
\AppCSmall
\AppMLlamaSmall & 23.33 & 34.36 & 0.59 & 22.21 & 31.01 & 0.65 & 23.52 & 34.43 & 0.58 & 28.83 & 40.15 & 0.52 & 26.59 & 37.92 & 0.49 & 24.90 & 35.57 & 0.56\\
\AppMQwenSmall & 38.40 & 49.22 & 0.33 & 30.75 & 38.93 & 0.55 & 27.70 & 38.56 & 0.47 & 39.31 & 50.16 & 0.39 & 33.70 & 44.03 & 0.44 & 33.97 & 44.18 & 0.44\\
\AppMGemmaSmall & 22.14 & 31.59 & 0.64 & 32.70 & 43.56 & 0.52 & 32.36 & 45.62 & 0.48 & 31.25 & 42.09 & 0.53 & 28.65 & 38.52 & 0.57 & 29.42 & 40.28 & 0.55\\
\AppCMid
\AppMGemmaMid & \fwsecond{73.23} & \fwsecond{78.88} & 0.17 & \fwbest{57.11} & \fwbest{64.73} & 0.31 & \fwsecond{53.04} & \fwsecond{64.37} & 0.29 & 33.91 & 41.14 & 0.55 & \fwbest{61.92} & \fwbest{68.99} & 0.26 & \fwsecond{55.84} & \fwsecond{63.62} & 0.32\\
\AppMQwenMid & 62.61 & 70.99 & 0.18 & 43.31 & 52.80 & 0.35 & 38.38 & 50.58 & 0.35 & 41.31 & 51.15 & 0.36 & 42.98 & 53.16 & 0.33 & 45.72 & 55.74 & 0.31\\
\AppCLarge
\AppMLlamaLarge & 40.04 & 50.20 & 0.24 & 44.53 & 54.04 & 0.30 & 31.25 & 43.66 & 0.31 & 33.29 & 43.55 & 0.34 & 27.32 & 37.48 & 0.36 & 35.29 & 45.79 & 0.31\\
\AppMQwenNext & 37.80 & 48.37 & 0.40 & 41.01 & 51.79 & 0.42 & 36.92 & 49.27 & 0.42 & 31.48 & 40.23 & 0.54 & 33.41 & 42.44 & 0.49 & 36.12 & 46.42 & 0.45\\
\AppMGLM & 43.24 & 54.48 & 0.21 & 36.74 & 46.40 & 0.33 & 37.71 & 50.39 & 0.27 & 34.19 & 44.61 & 0.35 & 32.57 & 42.72 & 0.30 & 36.89 & 47.72 & 0.29\\
\AppCHuge
\AppMQwenHuge & 57.51 & 66.67 & 0.16 & 43.30 & 51.71 & 0.39 & 43.38 & 55.52 & 0.30 & 39.49 & 48.34 & 0.37 & 36.50 & 45.33 & 0.35 & 44.04 & 53.51 & 0.31\\
\AppMDeepSeek & 64.44 & 71.81 & 0.13 & 48.09 & 58.01 & \fwsecond{0.27} & 45.03 & 56.10 & 0.24 & \fwsecond{45.64} & 53.26 & 0.29 & 46.65 & 55.60 & 0.19 & 49.97 & 58.96 & 0.23\\
\AppMKimi & 63.99 & 72.41 & 0.13 & 50.89 & 59.63 & 0.30 & 43.43 & 54.88 & 0.27 & 42.18 & 50.43 & 0.35 & 41.89 & 49.84 & 0.26 & 48.48 & 57.44 & 0.26\\
\AppCClosed
\AppMGPT & \fwbest{75.06} & \fwbest{80.52} & \fwsecond{0.10} & \fwsecond{54.43} & 59.50 & 0.33 & \fwbest{57.74} & \fwbest{68.16} & \fwsecond{0.20} & \fwbest{57.24} & \fwbest{63.38} & \fwsecond{0.24} & \fwsecond{58.35} & \fwsecond{64.65} & \fwbest{0.18} & \fwbest{60.56} & \fwbest{67.24} & \fwsecond{0.21}\\
\AppMGemini & 59.40 & 67.96 & 0.21 & 50.76 & 59.57 & 0.36 & 41.15 & 53.33 & 0.35 & 40.03 & 48.68 & 0.44 & 42.85 & 51.90 & 0.34 & 46.84 & 56.29 & 0.34\\
\AppMClaude & 56.64 & 66.54 & \fwbest{0.09} & 53.52 & \fwsecond{63.53} & \fwbest{0.19} & 41.52 & 53.59 & \fwbest{0.16} & 45.63 & \fwsecond{54.69} & \fwbest{0.23} & 36.94 & 45.52 & \fwsecond{0.19} & 46.85 & 56.77 & \fwbest{0.17}\\
\AppMQwenPlus & 62.22 & 71.28 & 0.20 & 50.14 & 58.46 & 0.34 & 40.88 & 52.25 & 0.38 & 39.22 & 47.98 & 0.42 & 50.50 & 60.45 & 0.29 & 48.59 & 58.08 & 0.33\\
\midrule
\AppCIHead
\AppCISmall
\AppMInternSmall & 34.07 & 42.03 & 0.37 & 28.20 & 38.72 & \fwbest{0.24} & 33.31 & 45.26 & 0.39 & 26.53 & 36.83 & \fwsecond{0.31} & 29.20 & 40.54 & \fwsecond{0.22} & 30.26 & 40.68 & \fwsecond{0.31}\\
\AppMQwenVLSmall & 31.53 & 38.43 & 0.50 & 26.70 & 35.56 & 0.61 & 27.79 & 38.09 & 0.56 & 22.90 & 30.09 & 0.64 & 29.93 & 40.96 & 0.52 & 27.77 & 36.63 & 0.56\\
\AppCIMid
\AppMInternMid & 44.26 & 51.54 & 0.36 & 33.99 & 42.91 & 0.53 & 40.47 & 51.59 & 0.41 & 30.60 & 36.58 & 0.60 & 36.38 & 47.15 & 0.37 & 37.14 & 45.95 & 0.46\\
\AppMQwenVLMid & \fwsecond{55.42} & \fwsecond{61.18} & \fwsecond{0.23} & 42.99 & 52.06 & 0.38 & 42.29 & 53.02 & 0.33 & 38.61 & \fwsecond{46.01} & 0.43 & 32.47 & 42.29 & 0.31 & 42.36 & 50.91 & 0.34\\
\AppCIHuge
\AppMInternLarge & 47.66 & 54.21 & 0.34 & 38.65 & 48.26 & 0.48 & 40.39 & 51.46 & 0.40 & 34.14 & 42.12 & 0.54 & 32.86 & 43.86 & 0.51 & 38.74 & 47.98 & 0.45\\
\AppMQwenVLHuge & 46.92 & 52.24 & 0.35 & 45.02 & 53.31 & 0.38 & 38.17 & 48.29 & 0.40 & \fwsecond{38.86} & 45.00 & 0.47 & 19.99 & 26.67 & 0.36 & 37.79 & 45.10 & 0.39\\
\AppCIClosed
\AppMGPT & \fwbest{59.46} & \fwbest{63.86} & 0.28 & \fwsecond{45.73} & \fwsecond{53.33} & 0.38 & \fwsecond{45.57} & \fwsecond{55.01} & 0.35 & 38.27 & 44.44 & 0.40 & \fwbest{45.88} & \fwbest{55.64} & 0.27 & \fwsecond{46.98} & \fwsecond{54.46} & 0.34\\
\AppMGemini & 55.36 & 60.10 & 0.29 & \fwbest{48.29} & \fwbest{56.47} & 0.35 & \fwbest{49.05} & \fwbest{59.56} & \fwsecond{0.27} & \fwbest{47.97} & \fwbest{55.60} & 0.34 & \fwsecond{40.67} & \fwsecond{50.92} & 0.31 & \fwbest{48.27} & \fwbest{56.53} & 0.31\\
\AppMClaude & 54.26 & 59.45 & \fwbest{0.22} & 39.06 & 48.13 & \fwsecond{0.31} & 40.56 & 51.03 & \fwbest{0.24} & 37.14 & 45.61 & \fwbest{0.31} & 27.44 & 34.47 & \fwbest{0.13} & 39.69 & 47.74 & \fwbest{0.24}\\
\AppMQwenPlus & 48.35 & 54.28 & 0.36 & 44.33 & 53.25 & 0.38 & 39.95 & 50.13 & 0.40 & 34.44 & 42.51 & 0.47 & 29.40 & 39.29 & 0.39 & 39.29 & 47.89 & 0.40\\
}
\paragraph{Physical capability decomposition.}
The multiple-choice results already reveal a substantial capability gradient.
On the S-track, the model-averaged Acc. decreases from 52.00 on P1 to 38.87
on P4, with only a small recovery to 40.05 on P5.
The trend is therefore not perfectly monotonic for every capability, but
higher-level coupled-field and mechanism-oriented reasoning is consistently
harder than temporal prediction for the model population as a whole.
Strong individual models can partially resist this degradation:
GPT-5.6-Luna maintains comparatively high performance across the axis, while
Gemma-3-27B-IT is particularly competitive on several physical capability
levels.
Nevertheless, no single model dominates every capability and every metric.

The I-track presents a different profile.
Performance is generally more variable across capability levels and models,
indicating that converting physics-rendered visual evidence into the latent
physical state required by the task introduces an additional source of
difficulty.
In particular, models that are competitive when operating on structured
sensor records do not necessarily retain the same relative advantage under
visual observation.
This interface- and modality-dependent behavior supports the use of both
tracks rather than treating multimodal rendering as a direct substitute for
structured physical measurements.
\AppCTable{\textbf{Open-report evaluation along the physical capability axis.} Models are evaluated on the controlled simulation subset under the structured-observation track (S-track) and multimodal image-observation track (I-track). Each dimension reports accuracy (Acc), macro-F1 (F1), and Brier score; Average is the equal-weight mean across P1--P5.}
{tab:main_open_physical_axis}
{\AppCPHeader{Acc}{F1}{Brier}}
{%
\AppCSHead
\AppCSmall
\AppMLlamaSmall & 40.53 & 50.73 & 0.30 & 40.49 & 48.57 & 0.33 & 33.25 & 43.95 & 0.30 & 45.48 & 40.02 & \fwbest{0.32} & 32.17 & 39.71 & 0.36 & 38.38 & 44.60 & 0.32\\
\AppMQwenSmall & 54.71 & 64.14 & 0.29 & 39.94 & 46.13 & 0.42 & 40.05 & 51.53 & 0.29 & 46.64 & 40.99 & 0.35 & 29.46 & 35.76 & 0.43 & 42.16 & 47.71 & 0.35\\
\AppMGemmaSmall & 47.21 & 56.54 & 0.26 & 44.44 & 54.36 & 0.32 & 34.25 & 44.15 & 0.36 & 36.82 & 31.11 & 0.41 & 27.13 & 33.49 & 0.45 & 37.97 & 43.93 & 0.36\\
\AppCMid
\AppMGemmaMid & 44.84 & 56.05 & 0.29 & 49.69 & 56.45 & 0.32 & 40.09 & 50.91 & 0.33 & 40.24 & 34.77 & 0.38 & 31.78 & 39.65 & 0.40 & 41.33 & 47.57 & 0.34\\
\AppMQwenMid & 53.99 & 64.29 & 0.23 & 48.21 & 54.38 & 0.33 & 41.93 & 53.10 & 0.30 & 44.03 & 38.45 & 0.39 & 36.24 & 43.49 & 0.37 & 44.88 & 50.74 & 0.33\\
\AppCLarge
\AppMLlamaLarge & 54.11 & 63.57 & 0.20 & 43.46 & 48.62 & 0.35 & 42.97 & 55.03 & \fwsecond{0.21} & 42.76 & 36.92 & \fwsecond{0.32} & 28.29 & 36.28 & 0.33 & 42.32 & 48.08 & 0.28\\
\AppMQwenNext & 59.01 & \fwbest{69.13} & 0.26 & 44.20 & 51.52 & 0.40 & 42.13 & 52.86 & 0.36 & 34.33 & 29.11 & 0.53 & 34.50 & 42.03 & 0.47 & 42.83 & 48.93 & 0.40\\
\AppMGLM & 55.81 & 65.46 & 0.22 & 42.78 & 48.08 & 0.38 & 41.97 & 52.88 & 0.28 & 40.79 & 35.28 & 0.37 & 26.55 & 33.31 & 0.40 & 41.58 & 47.00 & 0.33\\
\AppCHuge
\AppMQwenHuge & 53.25 & 63.56 & 0.23 & 42.22 & 47.60 & 0.39 & 44.08 & 55.30 & 0.28 & 41.64 & 35.78 & 0.38 & 36.43 & 42.91 & 0.34 & 43.52 & 49.03 & 0.32\\
\AppMDeepSeek & 56.96 & 66.12 & 0.22 & 50.37 & 58.07 & 0.29 & 45.20 & 56.63 & 0.24 & \fwbest{48.18} & \fwsecond{43.45} & 0.33 & \fwbest{42.25} & \fwbest{50.12} & \fwbest{0.24} & \fwsecond{48.59} & \fwsecond{54.88} & \fwbest{0.26}\\
\AppMKimi & 58.56 & 68.18 & 0.20 & 52.72 & 59.97 & \fwsecond{0.28} & \fwsecond{45.64} & \fwsecond{57.07} & 0.25 & 39.79 & 36.98 & 0.40 & 34.11 & 41.98 & 0.30 & 46.16 & 52.84 & 0.29\\
\AppCClosed
\AppMGPT & 58.12 & 68.72 & \fwbest{0.17} & \fwbest{54.44} & \fwsecond{60.47} & 0.32 & \fwbest{53.24} & \fwbest{64.99} & \fwbest{0.21} & \fwsecond{47.03} & \fwbest{44.03} & 0.37 & 37.79 & 45.23 & \fwsecond{0.27} & \fwbest{50.12} & \fwbest{56.69} & \fwsecond{0.27}\\
\AppMGemini & \fwbest{60.45} & \fwsecond{69.07} & 0.25 & 50.86 & 58.46 & 0.33 & 43.88 & 54.98 & 0.33 & 34.61 & 32.56 & 0.51 & 37.98 & 45.12 & 0.34 & 45.56 & 52.04 & 0.35\\
\AppMClaude & 56.48 & 66.06 & \fwsecond{0.18} & \fwsecond{54.38} & \fwbest{61.04} & \fwbest{0.25} & 38.33 & 49.68 & 0.24 & 38.42 & 36.85 & 0.40 & 25.97 & 32.73 & 0.29 & 42.72 & 49.27 & 0.27\\
\AppMQwenPlus & \fwsecond{59.31} & 68.86 & 0.24 & 44.26 & 49.47 & 0.42 & 44.64 & 55.71 & 0.32 & 42.00 & 37.81 & 0.43 & \fwsecond{39.34} & \fwsecond{46.86} & 0.32 & 45.91 & 51.74 & 0.35\\
\midrule
\AppCIHead
\AppCISmall
\AppMInternSmall & 19.65 & 15.70 & 0.54 & 21.96 & 18.92 & 0.58 & 34.84 & 27.83 & 0.47 & 26.64 & 21.09 & 0.53 & 23.33 & 23.33 & 0.68 & 25.28 & 21.37 & 0.56\\
\AppMQwenVLSmall & 32.31 & 42.14 & 0.52 & 16.24 & 19.19 & 0.72 & 37.10 & 41.86 & 0.49 & 26.71 & 27.31 & 0.62 & 42.92 & 42.92 & 0.43 & 31.06 & 34.68 & 0.56\\
\AppCIMid
\AppMInternMid & 39.61 & 28.87 & 0.29 & 19.33 & 17.71 & 0.63 & 35.44 & 31.45 & 0.45 & \fwbest{40.50} & \fwsecond{32.45} & \fwsecond{0.40} & 27.50 & 27.50 & 0.57 & 32.48 & 27.60 & 0.47\\
\AppMQwenVLMid & \fwsecond{61.13} & \fwsecond{71.73} & 0.14 & \fwbest{37.48} & \fwbest{41.42} & \fwsecond{0.45} & 40.72 & 46.20 & 0.39 & \fwsecond{35.20} & 32.15 & 0.42 & 39.17 & 39.17 & 0.44 & \fwbest{42.74} & \fwbest{46.13} & 0.37\\
\AppCIHuge
\AppMInternLarge & 44.26 & 33.11 & 0.26 & 35.70 & 32.93 & 0.51 & 31.30 & 27.21 & 0.51 & 32.71 & 27.01 & 0.49 & \fwbest{49.58} & \fwbest{49.58} & 0.39 & 38.71 & 33.97 & 0.43\\
\AppMQwenVLHuge & 29.11 & 37.41 & 0.33 & 35.83 & 39.61 & 0.46 & 35.14 & 39.82 & 0.41 & 29.91 & 30.63 & \fwbest{0.38} & 21.25 & 21.25 & \fwsecond{0.37} & 30.25 & 33.74 & 0.39\\
\AppCIClosed
\AppMGPT & \fwbest{62.02} & \fwbest{72.50} & \fwbest{0.11} & 36.49 & 40.04 & \fwbest{0.41} & \fwsecond{42.61} & \fwsecond{47.35} & \fwbest{0.34} & 30.53 & 31.82 & 0.41 & 23.33 & 23.33 & 0.37 & 39.00 & 43.01 & \fwbest{0.33}\\
\AppMGemini & 50.33 & 59.79 & 0.27 & 33.40 & 36.57 & 0.56 & \fwbest{42.68} & \fwbest{47.85} & 0.40 & 34.27 & \fwbest{35.37} & 0.51 & \fwsecond{46.25} & \fwsecond{46.25} & 0.40 & \fwsecond{41.39} & \fwsecond{45.17} & 0.43\\
\AppMClaude & 57.82 & 68.39 & \fwsecond{0.11} & 29.98 & 33.31 & 0.49 & 40.42 & 45.63 & \fwsecond{0.36} & 30.92 & 31.05 & 0.45 & 37.50 & 37.50 & \fwbest{0.29} & 39.33 & 43.18 & \fwsecond{0.34}\\
\AppMQwenPlus & 41.49 & 49.84 & 0.38 & \fwsecond{37.21} & \fwsecond{40.20} & 0.50 & 41.48 & 46.33 & 0.44 & 31.07 & 29.16 & 0.50 & \fwsecond{46.25} & \fwsecond{46.25} & 0.40 & 39.50 & 42.36 & 0.44\\
}
Open-report evaluation sharpens this separation.
On the S-track, the model-averaged Acc. declines from 54.22 on P1 to 33.33
on P5, showing that the difficulty of the physical capability hierarchy
becomes more visible when candidate answers are removed.
The strongest models remain substantially better at temporal evolution and
field-level interpretation than at constructing reliable counterfactual
physical commitments.
The I-track is more irregular: its mean performance drops sharply between
P1 and P2 and remains substantially below the strongest S-track results over
most higher-level capabilities.
Thus, open reporting does not simply reproduce the ranking observed under
multiple choice; it exposes additional failures in state recovery and
physical commitment that can be hidden by candidate-conditioned answering.

Taken together, the two response interfaces suggest that FireWorldBench
measures more than answer selection accuracy.
Multiple-choice evaluation captures whether a model can recognize a
physically plausible hypothesis, whereas open-report evaluation additionally
requires the model to recover, organize, and explicitly commit to the
underlying physical state.
The widening gap at the more demanding capability levels is therefore
consistent with the central benchmark observation that recognition of a
correct physical hypothesis is easier than independently reconstructing the
reasoning needed to support it.
\paragraph{Fire scenario task decomposition.}
The fire scenario axis provides a complementary view of the same models.
Unlike the physical capability axis, the three task categories are not
expected to form a strictly monotonic empirical accuracy sequence for every
model, because they represent distinct stages of fire evolution: Localized
Onset (T1), Coupled Propagation (T2), and Critical Transition (T3).
The relevant question is therefore not whether performance decreases at
every adjacent T-level, but whether models exhibit stable competence across
these distinct stages of fire-world evolution.
\AppFTable{\textbf{Choice-format evaluation along the fire scenario task axis.} Models are evaluated on the controlled simulation subset under the structured-observation track (S-track) and multimodal image-observation track (I-track). Each dimension reports accuracy (Acc), macro-F1 (F1), and Brier score; Average is the equal-weight mean across T1--T3.}
{tab:main_choice_fire_axis}
{\AppFTHeader{Acc}{F1}{Brier}}
{%
\AppFSHead
\AppFSmall
\AppMLlamaSmall & 24.67 & -- & 0.59 & 25.86 & -- & 0.56 & 24.09 & -- & 0.57 & 24.87 & -- & 0.57\\
\AppMQwenSmall & 33.76 & -- & 0.50 & 32.74 & -- & 0.42 & 37.31 & -- & 0.35 & 34.60 & -- & 0.43\\
\AppMGemmaSmall & 32.84 & -- & 0.52 & 31.19 & -- & 0.51 & 23.65 & -- & 0.62 & 29.23 & -- & 0.55\\
\AppFMid
\AppMGemmaMid & 47.45 & -- & 0.44 & 45.69 & -- & 0.36 & \fwsecond{70.60} & -- & 0.19 & \fwsecond{54.58} & -- & 0.33\\
\AppMQwenMid & 41.50 & -- & 0.37 & 40.53 & -- & 0.33 & 58.05 & -- & 0.22 & 46.69 & -- & 0.31\\
\AppFLarge
\AppMLlamaLarge & 41.58 & -- & 0.31 & 30.38 & -- & 0.32 & 37.08 & -- & 0.27 & 36.35 & -- & 0.30\\
\AppMQwenNext & 39.53 & -- & 0.45 & 32.35 & -- & 0.48 & 36.78 & -- & 0.42 & 36.22 & -- & 0.45\\
\AppMGLM & 35.71 & -- & 0.36 & 36.32 & -- & 0.28 & 40.76 & -- & 0.23 & 37.60 & -- & 0.29\\
\AppFHuge
\AppMQwenHuge & 40.92 & -- & 0.41 & 42.68 & -- & 0.30 & 52.63 & -- & 0.21 & 45.41 & -- & 0.30\\
\AppMDeepSeek & 45.37 & -- & \fwsecond{0.31} & \fwsecond{46.99} & -- & 0.23 & 60.30 & -- & 0.14 & 50.89 & -- & 0.23\\
\AppMKimi & 47.53 & -- & 0.35 & 42.73 & -- & 0.27 & 58.86 & -- & 0.16 & 49.71 & -- & 0.26\\
\AppFClosed
\AppMGPT & \fwbest{50.78} & -- & 0.34 & \fwbest{62.36} & -- & \fwsecond{0.17} & \fwbest{71.18} & -- & \fwsecond{0.12} & \fwbest{61.44} & -- & \fwsecond{0.21}\\
\AppMGemini & 48.40 & -- & 0.39 & 38.67 & -- & 0.39 & 55.55 & -- & 0.24 & 47.54 & -- & 0.34\\
\AppMClaude & \fwsecond{50.08} & -- & \fwbest{0.23} & 43.33 & -- & \fwbest{0.17} & 52.06 & -- & \fwbest{0.12} & 48.49 & -- & \fwbest{0.17}\\
\AppMQwenPlus & 45.87 & -- & 0.39 & 40.01 & -- & 0.38 & 59.50 & -- & 0.22 & 48.46 & -- & 0.33\\
\midrule
\AppFIHead
\AppFISmall
\AppMInternSmall & -- & -- & -- & -- & -- & -- & -- & -- & -- & -- & -- & --\\
\AppMQwenVLSmall & 26.02 & -- & 0.63 & 25.73 & -- & 0.57 & 30.98 & -- & 0.51 & 27.58 & -- & 0.57\\
\AppFIMid
\AppMInternMid & -- & -- & -- & -- & -- & -- & -- & -- & -- & -- & -- & --\\
\AppMQwenVLMid & 42.68 & -- & 0.41 & 40.16 & -- & 0.35 & 47.45 & -- & \fwsecond{0.26} & 43.43 & -- & 0.34\\
\AppFIHuge
\AppMInternLarge & -- & -- & -- & -- & -- & -- & -- & -- & -- & -- & -- & --\\
\AppMQwenVLHuge & 40.93 & -- & 0.45 & 40.89 & -- & 0.38 & 37.56 & -- & 0.35 & 39.79 & -- & 0.39\\
\AppFIClosed
\AppMGPT & \fwsecond{44.87} & \fwsecond{51.54} & 0.41 & \fwsecond{41.88} & \fwsecond{50.68} & 0.34 & \fwbest{54.75} & \fwbest{61.00} & 0.27 & \fwsecond{47.17} & \fwsecond{54.41} & 0.34\\
\AppMGemini & \fwbest{48.37} & \fwbest{55.72} & \fwsecond{0.37} & \fwbest{48.50} & \fwbest{58.63} & \fwsecond{0.28} & \fwsecond{50.26} & \fwsecond{56.91} & 0.30 & \fwbest{49.04} & \fwbest{57.09} & \fwsecond{0.31}\\
\AppMClaude & 39.11 & 47.56 & \fwbest{0.34} & 38.78 & 48.98 & \fwbest{0.24} & 44.95 & 50.77 & \fwbest{0.19} & 40.95 & 49.10 & \fwbest{0.26}\\
\AppMQwenPlus & 42.56 & 50.70 & 0.40 & 37.20 & 47.16 & 0.42 & 41.77 & 49.07 & 0.37 & 40.51 & 48.98 & 0.40\\

}
The multiple-choice results show pronounced stage-dependent variation.
Performance at Localized Onset does not necessarily transfer to Coupled
Propagation or Critical Transition. In the reported controlled-simulation
results, the model-level means vary across T1--T3 on both observation tracks,
and several models exchange relative positions between the three stages.
This reinforces the motivation for retaining the three fire-evolution
categories rather than collapsing fire reasoning into a single overall task.
\AppFTable{\textbf{Open-report evaluation along the fire scenario task axis.} Models are evaluated on the controlled simulation subset under the structured-observation track (S-track) and multimodal image-observation track (I-track). Each dimension reports accuracy (Acc), macro-F1 (F1), and Brier score; Average is the equal-weight mean across T1--T3.}
{tab:main_open_fire_axis}
{\AppFTHeader{Acc}{F1}{Brier}}
{%
\AppFSHead
\AppFSmall
\AppMLlamaSmall & 40.42 & 46.22 & \fwsecond{0.33} & 33.75 & 41.26 & 0.34 & 41.05 & 48.17 & 0.30 & 38.41 & 45.22 & 0.32\\
\AppMQwenSmall & 41.30 & 45.11 & 0.38 & 38.47 & 46.31 & 0.36 & 51.69 & 57.54 & 0.30 & 43.82 & 49.65 & 0.35\\
\AppMGemmaSmall & 42.79 & 49.67 & 0.35 & 27.35 & 34.18 & 0.44 & 45.50 & 51.18 & 0.29 & 38.55 & 45.01 & 0.36\\
\AppFMid
\AppMGemmaMid & 45.89 & 51.17 & 0.35 & 33.45 & 40.97 & 0.39 & 45.14 & 52.24 & 0.29 & 41.49 & 48.13 & 0.34\\
\AppMQwenMid & 45.74 & 50.47 & 0.34 & 37.13 & 44.73 & 0.39 & 52.97 & 59.46 & 0.24 & 45.28 & 51.55 & 0.33\\
\AppFLarge
\AppMLlamaLarge & 41.92 & 46.04 & 0.33 & 37.76 & 45.72 & 0.30 & 51.22 & 57.23 & 0.21 & 43.63 & 49.66 & 0.28\\
\AppMQwenNext & 42.84 & 47.56 & 0.41 & 31.27 & 38.31 & 0.51 & 56.93 & 62.83 & 0.28 & 43.68 & 49.57 & 0.40\\
\AppMGLM & 41.76 & 45.73 & 0.38 & 35.62 & 43.09 & 0.36 & 52.17 & 57.99 & 0.24 & 43.18 & 48.94 & 0.33\\
\AppFHuge
\AppMQwenHuge & 41.97 & 46.13 & 0.37 & 36.76 & 44.28 & 0.37 & 52.56 & 58.76 & 0.23 & 43.76 & 49.72 & 0.33\\
\AppMDeepSeek & 47.26 & \fwsecond{51.85} & \fwbest{0.32} & \fwsecond{43.87} & \fwsecond{51.70} & 0.30 & 56.63 & 62.40 & 0.21 & \fwsecond{49.25} & \fwsecond{55.32} & \fwsecond{0.27}\\
\AppMKimi & \fwsecond{47.38} & 51.71 & 0.35 & 40.27 & 48.03 & 0.32 & 56.23 & 62.09 & 0.20 & 47.96 & 53.94 & 0.29\\
\AppFClosed
\AppMGPT & \fwbest{48.45} & \fwbest{52.02} & 0.35 & \fwbest{50.66} & \fwbest{58.49} & \fwsecond{0.28} & 56.62 & 63.26 & \fwbest{0.17} & \fwbest{51.91} & \fwbest{57.92} & \fwbest{0.27}\\
\AppMGemini & 44.88 & 49.18 & 0.41 & 37.23 & 44.86 & 0.41 & \fwbest{58.36} & \fwsecond{63.50} & 0.25 & 46.82 & 52.51 & 0.36\\
\AppMClaude & 44.97 & 49.37 & 0.36 & 38.52 & 46.40 & \fwbest{0.27} & 51.85 & 58.32 & \fwsecond{0.19} & 45.11 & 51.36 & 0.27\\
\AppMQwenPlus & 44.04 & 47.01 & 0.41 & 38.71 & 46.37 & 0.41 & \fwsecond{58.17} & \fwbest{63.75} & 0.24 & 46.97 & 52.38 & 0.35\\
\midrule
\AppFIHead
\AppFISmall
\AppMInternSmall & -- & -- & -- & -- & -- & -- & -- & -- & -- & -- & -- & --\\
\AppMQwenVLSmall & 20.66 & 23.00 & 0.69 & 30.56 & 34.86 & 0.55 & 37.71 & 42.41 & 0.47 & 29.64 & 33.42 & 0.57\\
\AppFIMid
\AppMInternMid & -- & -- & -- & -- & -- & -- & -- & -- & -- & -- & -- & --\\
\AppMQwenVLMid & \fwsecond{38.37} & \fwbest{41.62} & \fwbest{0.43} & 33.76 & 38.56 & 0.44 & \fwbest{54.27} & \fwbest{60.42} & 0.23 & \fwbest{42.13} & \fwbest{46.87} & 0.37\\
\AppFIHuge
\AppMInternLarge & -- & -- & -- & -- & -- & -- & -- & -- & -- & -- & -- & --\\
\AppMQwenVLHuge & 36.32 & 38.87 & 0.47 & 30.27 & 34.94 & \fwsecond{0.37} & 28.30 & 31.80 & 0.35 & 31.63 & 35.20 & 0.40\\
\AppFIClosed
\AppMGPT & 36.66 & 38.18 & \fwsecond{0.43} & \fwsecond{36.54} & \fwsecond{41.45} & \fwbest{0.34} & 48.58 & 55.42 & \fwsecond{0.20} & 40.59 & \fwsecond{45.02} & \fwbest{0.32}\\
\AppMGemini & 36.47 & 37.18 & 0.54 & \fwbest{36.83} & \fwbest{42.40} & 0.46 & 48.91 & 55.09 & 0.31 & \fwsecond{40.74} & 44.89 & 0.43\\
\AppMClaude & 34.12 & 35.01 & 0.47 & 33.14 & 38.09 & 0.40 & \fwsecond{50.77} & \fwsecond{57.66} & \fwbest{0.17} & 39.34 & 43.59 & \fwsecond{0.35}\\
\AppMQwenPlus & \fwbest{41.41} & \fwsecond{40.91} & 0.46 & 32.04 & 36.61 & 0.51 & 43.14 & 48.59 & 0.38 & 38.86 & 42.04 & 0.45\\

}
The open-report results further amplify this stage dependence.
Removing candidate answers changes both absolute performance and the relative
strengths of the models across Localized Onset, Coupled Propagation, and
Critical Transition. In particular, the reported model-level mean Acc. is
lowest at T2 and highest at T3 on both observation tracks, showing that
independent reconstruction of coupled propagation is especially challenging
in the controlled subset. Closed-source systems remain among the strongest
models overall, but their advantages are not uniform across all three fire
stages or observation tracks. Likewise, several open-weight models are
competitive at individual stages while remaining substantially weaker at
others.

These results provide a more detailed interpretation of the averaged tables
in the main paper.
The aggregate score is useful for summarizing overall benchmark competence,
but it hides important structure: a model may identify a localized onset yet
fail to reconstruct coupled propagation or a critical transition, or may
perform well when evidence is supplied as structured records but fail to
reconstruct the same physical state from rendered observations.
The two-axis design therefore exposes qualitatively different failure modes
that cannot be recovered from a single task-level average.

\subsection{Real-World-Aligned Results}
\label{app:real_world_aligned_results}

We next provide the same decomposition on the real-world-aligned subset.
Tables~\ref{tab:rwa_choice_physical_axis} and
\ref{tab:rwa_open_physical_axis} report the physical capability axis, and
Tables~\ref{tab:rwa_choice_fire_axis} and
\ref{tab:rwa_open_fire_axis} report the fire scenario task axis.
The objective of this evaluation is not to reproduce the controlled
simulation setting exactly, but to test whether the capability boundaries
observed under controlled conditions remain visible when the benchmark is
grounded in real-world-aligned fire evidence.

As in the controlled simulation evaluation, choice and open-report results are
kept separate.
This distinction is particularly important in the real-world-aligned setting,
where observation ambiguity and incomplete evidence can make open-ended
physical-state reconstruction considerably more difficult than selecting
among predefined candidate hypotheses.

\AppCTable{\textbf{Choice-format evaluation along the physical capability axis.} Models are evaluated on the real-world-aligned subset under the structured-observation track (S-track) and multimodal image-observation track (I-track). Each dimension reports accuracy (Acc), macro-F1 (F1), and Brier score; Average is the equal-weight mean across P1--P5.}
{tab:rwa_choice_physical_axis}
{\AppCPHeader{Acc}{F1}{Brier}}
{%
\AppCSHead
\AppCSmall
\AppMLlamaSmall & 20.89 & 30.74 & 0.66 & 22.16 & 30.93 & 0.65 & 28.03 & 40.00 & 0.51 & 28.50 & 40.17 & 0.52 & 33.80 & 46.67 & 0.41 & 26.68 & 37.70 & 0.55\\
\AppMQwenSmall & 23.10 & 33.78 & 0.63 & 32.40 & 40.87 & 0.54 & 37.12 & 48.18 & 0.41 & 30.62 & 43.00 & 0.52 & 37.50 & 51.11 & 0.43 & 32.15 & 43.39 & 0.50\\
\AppMGemmaSmall & 42.08 & 52.41 & 0.42 & 43.92 & 55.69 & 0.39 & 37.95 & 49.00 & 0.43 & 42.75 & 55.25 & 0.39 & 43.06 & 53.89 & 0.39 & 41.95 & 53.25 & 0.40\\
\AppCMid
\AppMGemmaMid & 43.83 & 55.08 & 0.38 & 43.80 & 54.73 & 0.37 & 38.26 & 44.24 & 0.54 & 42.29 & 53.33 & 0.39 & 44.81 & 55.50 & 0.35 & 42.60 & 52.58 & 0.40\\
\AppMQwenMid & 53.69 & 62.33 & 0.23 & 47.29 & 56.81 & 0.33 & 48.18 & \fwsecond{57.88} & 0.28 & 38.29 & 47.26 & 0.41 & 40.28 & 50.74 & 0.38 & 45.55 & 55.00 & 0.33\\
\AppCLarge
\AppMLlamaLarge & 56.38 & 66.30 & 0.17 & 44.34 & 55.85 & 0.26 & 47.73 & 54.85 & \fwsecond{0.26} & 39.71 & 51.58 & 0.28 & 47.41 & 58.52 & 0.22 & 47.11 & 57.42 & 0.24\\
\AppMQwenNext & 38.44 & 48.25 & 0.41 & 53.21 & 63.75 & 0.28 & 36.82 & 46.06 & 0.44 & 44.79 & 57.58 & 0.33 & 45.37 & 55.93 & 0.35 & 43.73 & 54.31 & 0.36\\
\AppMGLM & 61.14 & 69.27 & 0.17 & 57.08 & 65.79 & 0.21 & \fwbest{51.14} & 57.27 & 0.27 & 43.71 & 55.07 & 0.27 & 40.19 & 52.96 & 0.21 & 50.65 & 60.07 & 0.22\\
\AppCHuge
\AppMQwenHuge & 52.88 & 62.18 & 0.23 & 46.24 & 55.23 & 0.36 & 41.29 & 50.15 & 0.37 & 37.04 & 46.83 & 0.43 & 36.30 & 48.89 & 0.37 & 42.75 & 52.66 & 0.35\\
\AppMDeepSeek & 63.45 & 71.18 & \fwbest{0.11} & 54.68 & 65.47 & \fwsecond{0.18} & 49.02 & 55.15 & \fwbest{0.21} & 37.71 & 44.83 & 0.35 & 41.57 & 51.67 & 0.22 & 49.29 & 57.66 & 0.21\\
\AppMKimi & 55.15 & 60.47 & 0.16 & 58.41 & 67.33 & 0.18 & 42.42 & 48.79 & 0.38 & 45.42 & 52.93 & 0.32 & 40.19 & 49.07 & 0.25 & 48.32 & 55.72 & 0.26\\
\AppCClosed
\AppMGPT & \fwbest{74.93} & \fwbest{79.26} & \fwbest{0.11} & \fwsecond{63.66} & 70.12 & 0.19 & 48.18 & 56.36 & 0.29 & \fwbest{65.67} & \fwbest{69.67} & \fwbest{0.21} & \fwsecond{57.31} & \fwsecond{65.93} & \fwbest{0.13} & \fwbest{61.95} & \fwbest{68.27} & \fwbest{0.19}\\
\AppMGemini & \fwsecond{67.40} & \fwsecond{73.80} & \fwsecond{0.15} & 57.61 & 66.74 & 0.26 & \fwsecond{50.00} & 57.42 & 0.34 & 49.75 & 58.75 & 0.35 & \fwbest{61.30} & \fwbest{68.15} & 0.22 & \fwsecond{57.21} & \fwsecond{64.97} & 0.26\\
\AppMClaude & 57.46 & 66.09 & 0.16 & \fwbest{64.80} & \fwbest{72.03} & \fwbest{0.15} & 48.03 & 55.61 & 0.33 & \fwsecond{58.04} & \fwsecond{65.17} & \fwsecond{0.21} & 51.11 & 61.11 & \fwsecond{0.20} & 55.89 & 64.00 & \fwsecond{0.21}\\
\AppMQwenPlus & 57.64 & 65.57 & 0.26 & 61.72 & \fwsecond{70.61} & 0.22 & \fwsecond{50.00} & \fwbest{58.79} & 0.33 & 51.87 & 60.08 & 0.31 & 51.11 & 63.33 & 0.26 & 54.47 & 63.68 & 0.28\\
\midrule
\AppCIHead
\AppCISmall
\AppMInternSmall & 24.23 & 27.32 & 0.49 & 29.44 & 38.81 & \fwbest{0.15} & 28.58 & 40.17 & 0.29 & 33.52 & 45.76 & \fwbest{0.15} & 30.83 & 42.50 & 0.23 & 29.32 & 38.91 & 0.26\\
\AppMQwenVLSmall & 29.67 & 34.31 & 0.51 & 37.59 & 46.67 & 0.49 & 22.58 & 30.25 & 0.65 & 29.17 & 36.82 & 0.57 & 29.58 & 40.50 & 0.51 & 29.72 & 37.71 & 0.55\\
\AppCIMid
\AppMInternMid & 40.73 & 44.72 & 0.44 & 48.52 & 56.67 & 0.39 & 34.83 & 43.68 & 0.51 & 44.77 & 52.12 & 0.44 & 37.33 & 48.00 & 0.37 & 41.24 & 49.04 & 0.43\\
\AppMQwenVLMid & 33.74 & 36.59 & 0.42 & 57.93 & 65.41 & 0.27 & 35.92 & 43.67 & 0.43 & 48.14 & 54.24 & 0.36 & 52.25 & 61.33 & 0.21 & 45.60 & 52.25 & 0.34\\
\AppCIHuge
\AppMInternLarge & 24.47 & 26.67 & 0.64 & 42.37 & 53.56 & 0.42 & 36.96 & 45.50 & 0.48 & 40.91 & 48.79 & 0.47 & 28.25 & 39.33 & 0.56 & 34.59 & 42.77 & 0.52\\
\AppMQwenVLHuge & 60.16 & \fwsecond{63.25} & 0.28 & 51.15 & 57.85 & 0.35 & 29.88 & 39.00 & 0.48 & 48.64 & 53.33 & 0.39 & 30.92 & 41.67 & 0.37 & 44.15 & 51.02 & 0.37\\
\AppCIClosed
\AppMGPT & 60.24 & 62.60 & \fwsecond{0.21} & \fwbest{65.44} & \fwbest{70.52} & 0.20 & \fwbest{45.92} & \fwbest{52.83} & \fwsecond{0.28} & \fwbest{59.24} & \fwbest{64.09} & 0.24 & \fwbest{55.83} & \fwsecond{63.00} & \fwsecond{0.16} & \fwbest{57.33} & \fwbest{62.61} & \fwsecond{0.22}\\
\AppMGemini & \fwbest{60.77} & 62.28 & 0.32 & 55.19 & 60.59 & 0.37 & 38.29 & \fwsecond{48.33} & 0.41 & 43.71 & 50.91 & 0.46 & \fwsecond{54.08} & \fwbest{66.33} & 0.20 & 50.41 & 57.69 & 0.35\\
\AppMClaude & \fwsecond{60.37} & \fwbest{63.74} & \fwbest{0.19} & \fwsecond{61.44} & \fwsecond{66.96} & \fwsecond{0.20} & \fwsecond{38.75} & 47.17 & \fwbest{0.27} & \fwsecond{50.04} & \fwsecond{57.32} & \fwsecond{0.23} & 49.83 & 60.00 & \fwbest{0.12} & \fwsecond{52.09} & \fwsecond{59.04} & \fwbest{0.20}\\
\AppMQwenPlus & 50.77 & 53.50 & 0.39 & 54.81 & 62.37 & 0.30 & 36.79 & 43.67 & 0.46 & 36.70 & 43.94 & 0.45 & 40.00 & 52.33 & 0.28 & 43.81 & 51.16 & 0.38\\
}
\paragraph{Physical capability decomposition.}
The real-world-aligned results preserve the broad capability separation found
in controlled simulations, although individual cells are more variable.
For the S-track choice setting, the model-averaged Acc. is highest at P1 and
remains lower on the more demanding coupled-field and mechanism-oriented
capabilities.
The I-track shows an even less uniform profile, reflecting the additional
difficulty introduced by multimodal observation and the smaller,
heterogeneous real-world-aligned evaluation set.

The important pattern is therefore not a perfectly monotonic decrease for
every cell, but the absence of a model that is uniformly strong across all
five physical capabilities and both observation modalities.
Several models that perform competitively on early physical interpretation
lose substantial performance on coupling, attribution, or counterfactual
reasoning, while others exhibit the opposite trade-off on particular cells.
\AppCTable{\textbf{Open-report evaluation along the physical capability axis.} Models are evaluated on the real-world-aligned subset under the structured-observation track (S-track) and multimodal image-observation track (I-track). Each dimension reports accuracy (Acc), macro-F1 (F1), and Brier score; Average is the equal-weight mean across P1--P5.}
{tab:rwa_open_physical_axis}
{\AppCPHeader{Acc}{F1}{Brier}}
{%
\AppCSHead
\AppCSmall
\AppMLlamaSmall & 49.42 & 57.22 & 0.26 & 39.55 & 48.61 & 0.32 & 37.88 & 42.73 & 0.36 & 30.00 & 35.50 & 0.40 & 35.19 & 42.78 & 0.32 & 38.41 & 45.37 & 0.33\\
\AppMQwenSmall & 52.79 & 57.04 & 0.35 & 37.73 & 45.72 & 0.43 & \fwbest{60.61} & \fwbest{67.73} & \fwsecond{0.20} & 47.50 & 53.50 & 0.33 & 31.48 & 38.89 & 0.39 & 46.02 & 52.58 & 0.34\\
\AppMGemmaSmall & 45.39 & 51.86 & 0.29 & 44.77 & 56.30 & 0.29 & 40.91 & 45.45 & 0.34 & 38.33 & 43.00 & 0.44 & 33.33 & 41.11 & 0.40 & 40.55 & 47.54 & 0.35\\
\AppCMid
\AppMGemmaMid & 70.04 & 75.11 & 0.17 & 42.50 & 53.78 & 0.33 & 50.00 & 55.00 & 0.33 & 45.00 & 53.50 & 0.35 & 27.78 & 37.22 & 0.45 & 47.06 & 54.92 & 0.33\\
\AppMQwenMid & 71.73 & 76.60 & 0.15 & 43.86 & 52.60 & 0.36 & 56.06 & 63.18 & 0.25 & 40.00 & 44.00 & 0.45 & 37.04 & 47.78 & 0.34 & 49.74 & 56.83 & 0.31\\
\AppCLarge
\AppMLlamaLarge & \fwbest{75.86} & \fwbest{79.59} & \fwbest{0.13} & 52.73 & 61.77 & 0.20 & 46.97 & 53.18 & 0.26 & \fwbest{58.33} & \fwbest{65.75} & \fwbest{0.19} & \fwsecond{42.59} & \fwsecond{51.67} & \fwbest{0.23} & 55.30 & 62.39 & \fwbest{0.20}\\
\AppMQwenNext & 67.52 & 71.84 & 0.21 & 59.32 & 68.17 & 0.24 & 56.06 & \fwsecond{63.64} & 0.27 & 47.50 & 52.25 & 0.41 & 40.74 & 50.00 & 0.41 & 54.23 & 61.18 & 0.31\\
\AppMGLM & 72.18 & 75.56 & 0.17 & 49.55 & 57.44 & 0.29 & 51.52 & 57.73 & 0.29 & 47.50 & 51.50 & 0.36 & 31.48 & 41.67 & 0.35 & 50.45 & 56.78 & 0.29\\
\AppCHuge
\AppMQwenHuge & 54.73 & 59.11 & 0.25 & 48.41 & 58.89 & 0.29 & 51.52 & 58.64 & 0.28 & 46.67 & 54.00 & \fwsecond{0.30} & 29.63 & 38.89 & 0.34 & 46.19 & 53.91 & 0.29\\
\AppMDeepSeek & 63.27 & 66.70 & 0.22 & \fwsecond{64.32} & \fwsecond{73.28} & \fwsecond{0.16} & 45.45 & 50.91 & 0.30 & 43.33 & 48.00 & 0.42 & 29.63 & 37.78 & 0.40 & 49.20 & 55.33 & 0.30\\
\AppMKimi & 67.19 & 71.57 & 0.14 & 62.05 & 70.79 & 0.17 & 50.00 & 56.36 & 0.27 & 37.50 & 41.50 & 0.43 & 37.04 & 46.11 & 0.34 & 50.76 & 57.27 & 0.27\\
\AppCClosed
\AppMGPT & \fwsecond{75.65} & \fwsecond{78.69} & \fwsecond{0.14} & 59.55 & 68.19 & 0.25 & \fwsecond{59.09} & \fwbest{67.73} & 0.21 & \fwsecond{55.00} & 57.00 & 0.36 & \fwsecond{42.59} & \fwsecond{51.67} & \fwsecond{0.29} & \fwbest{58.38} & \fwbest{64.66} & 0.25\\
\AppMGemini & 72.18 & 76.49 & 0.16 & 57.50 & 67.55 & 0.20 & 46.97 & 52.73 & 0.34 & 47.50 & 53.75 & 0.35 & 33.33 & 44.44 & 0.36 & 51.50 & 58.99 & 0.28\\
\AppMClaude & 70.49 & 74.71 & 0.15 & \fwbest{65.45} & \fwbest{74.60} & \fwbest{0.14} & 53.03 & 60.45 & \fwbest{0.19} & 52.50 & 57.25 & 0.33 & 29.63 & 35.56 & 0.38 & 54.22 & 60.51 & \fwsecond{0.24}\\
\AppMQwenPlus & 71.43 & 75.42 & 0.18 & 53.18 & 62.43 & 0.28 & 51.52 & 57.73 & 0.34 & \fwsecond{55.00} & \fwsecond{60.50} & 0.32 & \fwbest{50.00} & \fwbest{58.33} & 0.31 & \fwsecond{56.23} & \fwsecond{62.88} & 0.29\\
\midrule
\AppCIHead
\AppCISmall
\AppMInternSmall & 14.18 & 12.53 & 0.71 & 28.15 & 20.59 & 0.48 & 24.58 & 16.64 & 0.51 & 25.76 & 19.55 & 0.49 & 30.00 & 30.00 & 0.60 & 24.53 & 19.86 & 0.56\\
\AppMQwenVLSmall & 55.04 & 57.46 & 0.37 & 45.93 & 51.19 & 0.41 & 35.42 & 40.92 & 0.50 & 31.82 & 35.45 & 0.57 & \fwsecond{50.00} & \fwsecond{50.00} & 0.44 & 43.64 & 47.00 & 0.46\\
\AppCIMid
\AppMInternMid & 21.33 & 14.83 & 0.47 & 28.89 & 22.30 & 0.46 & 37.08 & 30.44 & 0.41 & 47.73 & 39.77 & 0.36 & 25.00 & 25.00 & 0.59 & 32.01 & 26.47 & 0.46\\
\AppMQwenVLMid & 55.72 & \fwbest{61.57} & 0.24 & 49.26 & 54.44 & \fwsecond{0.35} & 42.08 & 48.50 & 0.38 & \fwbest{53.79} & \fwbest{61.82} & \fwbest{0.26} & 45.00 & 45.00 & 0.42 & \fwbest{49.17} & \fwbest{54.27} & 0.33\\
\AppCIHuge
\AppMInternLarge & 30.66 & 23.35 & 0.44 & 33.33 & 25.85 & 0.44 & 31.25 & 23.75 & 0.45 & \fwsecond{51.52} & 43.18 & \fwsecond{0.33} & \fwbest{65.00} & \fwbest{65.00} & \fwbest{0.27} & 42.35 & 36.23 & 0.38\\
\AppMQwenVLHuge & 55.24 & 56.36 & 0.27 & 31.85 & 37.93 & 0.51 & 31.67 & 38.67 & 0.47 & 40.15 & 45.23 & 0.45 & 35.00 & 35.00 & 0.45 & 38.78 & 42.64 & 0.43\\
\AppCIClosed
\AppMGPT & \fwbest{58.50} & \fwsecond{61.27} & \fwsecond{0.19} & \fwbest{55.56} & \fwbest{61.33} & \fwbest{0.20} & 36.67 & 41.83 & \fwsecond{0.34} & 47.73 & 52.27 & 0.36 & 30.00 & 30.00 & 0.38 & \fwsecond{45.69} & \fwsecond{49.34} & \fwbest{0.29}\\
\AppMGemini & 57.15 & 59.13 & 0.28 & 44.44 & 49.41 & 0.40 & 37.08 & 45.42 & 0.39 & 50.76 & \fwsecond{56.82} & 0.33 & 35.00 & 35.00 & 0.51 & 44.89 & 49.16 & 0.38\\
\AppMClaude & 56.33 & 59.62 & \fwbest{0.19} & 41.85 & 47.56 & 0.36 & \fwbest{43.75} & \fwsecond{50.42} & \fwbest{0.29} & 37.12 & 41.36 & 0.43 & 20.00 & 20.00 & \fwsecond{0.36} & 39.81 & 43.79 & \fwsecond{0.32}\\
\AppMQwenPlus & \fwsecond{58.00} & 59.19 & 0.30 & \fwsecond{50.00} & \fwsecond{55.33} & 0.37 & \fwsecond{42.92} & \fwbest{50.50} & 0.39 & 44.70 & 50.23 & 0.39 & 20.00 & 20.00 & 0.54 & 43.12 & 47.05 & 0.40\\
}
The open-report setting produces a substantially clearer difficulty gradient.
For the S-track, the model-averaged Acc. decreases from 65.32 on P1 to 52.03
on P2, 50.51 on P3, 46.11 on P4, and 35.43 on P5.
This progression is especially informative because models cannot rely on
candidate answers: they must directly recover the relevant physical state
from the available evidence and express the required commitment.
The large decline toward counterfactual intervention reasoning indicates that
successful perception or forecasting does not automatically translate into
reliable reasoning about how the system would respond under a hypothetical
change.

The I-track again exhibits stronger variance, but the same high-level
conclusion holds.
Visual access to a physical field does not guarantee that the model can
translate that field into a consistent latent representation suitable for
coupled reasoning, mechanism attribution, or intervention.
The real-world-aligned results therefore support rather than replace the
controlled simulation analysis: the latter provides controlled physical
coverage, while the former tests whether the same capability boundaries
remain observable under more realistic evidence conditions.
\paragraph{Fire scenario task decomposition.}
The fire scenario decomposition reveals substantial heterogeneity across
Localized Onset, Coupled Propagation, and Critical Transition.
In particular, strong performance at one stage of fire evolution is not a
reliable indicator of strong performance at the remaining stages.
This is visible for both structured and multimodal observations and is
consistent with the stage-specific trends observed on the controlled
simulation subset.
\AppFTable{\textbf{Choice-format evaluation along the fire scenario task axis.} Models are evaluated on the real-world-aligned subset under the structured-observation track (S-track) and multimodal image-observation track (I-track). Each dimension reports accuracy (Acc), macro-F1 (F1), and Brier score; Average is the equal-weight mean across T1--T3.}
{tab:rwa_choice_fire_axis}
{\AppFTHeader{Acc}{F1}{Brier}}
{%
\AppFSHead
\AppFSmall
\AppMLlamaSmall & 24.49 & 34.01 & 0.60 & 27.80 & 39.92 & 0.53 & 24.70 & 35.44 & 0.58 & 25.66 & 36.46 & 0.57\\
\AppMQwenSmall & 32.70 & 42.44 & 0.53 & 32.72 & 44.39 & 0.47 & 27.35 & 38.89 & 0.57 & 30.92 & 41.91 & 0.52\\
\AppMGemmaSmall & 43.37 & 54.98 & 0.39 & 40.45 & 52.80 & 0.40 & 42.37 & 52.84 & 0.41 & 42.06 & 53.54 & 0.40\\
\AppFMid
\AppMGemmaMid & 43.75 & 54.38 & 0.38 & 39.43 & 48.29 & 0.46 & 44.12 & 55.20 & 0.37 & 42.43 & 52.62 & 0.40\\
\AppMQwenMid & 40.86 & 49.07 & 0.40 & 49.19 & 60.34 & 0.27 & 49.73 & 58.91 & 0.28 & 46.59 & 56.11 & 0.32\\
\AppFLarge
\AppMLlamaLarge & 43.32 & 54.98 & 0.27 & 43.25 & 52.52 & 0.27 & 53.74 & 64.01 & 0.19 & 46.77 & 57.17 & 0.24\\
\AppMQwenNext & 50.38 & 61.26 & 0.30 & 40.69 & 52.20 & 0.38 & 40.49 & 50.52 & 0.39 & 43.85 & 54.66 & 0.36\\
\AppMGLM & 51.46 & 61.00 & 0.25 & 49.76 & 58.36 & 0.24 & 54.95 & 64.46 & 0.18 & 52.06 & 61.27 & 0.22\\
\AppFHuge
\AppMQwenHuge & 40.12 & 48.56 & 0.43 & 44.31 & 54.88 & 0.33 & 47.98 & 58.26 & 0.28 & 44.14 & 53.90 & 0.34\\
\AppMDeepSeek & 44.70 & 53.19 & 0.29 & 50.89 & 59.27 & \fwbest{0.19} & 56.99 & 65.42 & \fwsecond{0.14} & 50.86 & 59.29 & 0.21\\
\AppMKimi & 51.18 & 58.61 & 0.29 & 48.62 & 57.17 & 0.26 & 50.73 & 57.11 & 0.18 & 50.18 & 57.63 & 0.25\\
\AppFClosed
\AppMGPT & \fwbest{60.38} & \fwsecond{66.34} & \fwsecond{0.23} & \fwbest{62.52} & \fwbest{68.29} & \fwsecond{0.20} & \fwbest{69.73} & \fwbest{75.33} & \fwbest{0.12} & \fwbest{64.21} & \fwbest{69.99} & \fwbest{0.18}\\
\AppMGemini & 52.69 & 61.18 & 0.32 & 53.66 & 62.76 & 0.30 & \fwsecond{65.60} & \fwsecond{72.13} & 0.17 & 57.32 & 65.36 & 0.26\\
\AppMClaude & \fwsecond{60.17} & \fwbest{66.71} & \fwbest{0.20} & 56.54 & 64.96 & 0.22 & 55.59 & 64.62 & 0.17 & \fwsecond{57.43} & \fwsecond{65.43} & \fwsecond{0.20}\\
\AppMQwenPlus & 54.86 & 62.67 & 0.30 & \fwsecond{56.71} & \fwsecond{66.59} & 0.24 & 55.71 & 64.91 & 0.26 & 55.76 & 64.72 & 0.27\\
\midrule
\AppFIHead
\AppFISmall
\AppMInternSmall & -- & -- & -- & -- & -- & -- & -- & -- & -- & -- & -- & --\\
\AppMQwenVLSmall & 35.13 & 43.54 & 0.52 & 24.92 & 32.81 & 0.61 & 29.64 & 36.34 & 0.51 & 29.90 & 37.56 & 0.55\\
\AppFIMid
\AppMInternMid & -- & -- & -- & -- & -- & -- & -- & -- & -- & -- & -- & --\\
\AppMQwenVLMid & 53.03 & 59.13 & 0.34 & 42.42 & 50.52 & 0.36 & 39.81 & 44.70 & 0.35 & 45.09 & 51.45 & 0.35\\
\AppFIHuge
\AppMInternLarge & -- & -- & -- & -- & -- & -- & -- & -- & -- & -- & -- & --\\
\AppMQwenVLHuge & 44.13 & 48.97 & 0.44 & \fwsecond{43.26} & 51.98 & 0.37 & 50.57 & 56.17 & 0.31 & 45.99 & 52.37 & 0.37\\
\AppFIClosed
\AppMGPT & \fwbest{61.77} & \fwbest{66.26} & \fwsecond{0.25} & \fwbest{52.71} & \fwbest{59.38} & \fwbest{0.23} & \fwbest{58.80} & \fwsecond{62.73} & \fwsecond{0.19} & \fwbest{57.76} & \fwbest{62.79} & \fwsecond{0.22}\\
\AppMGemini & 53.21 & 58.56 & 0.39 & 38.75 & 48.33 & 0.44 & \fwsecond{58.58} & \fwbest{63.61} & 0.28 & 50.18 & 56.83 & 0.37\\
\AppMClaude & \fwsecond{58.03} & \fwsecond{62.97} & \fwbest{0.23} & 42.89 & \fwsecond{52.01} & \fwsecond{0.23} & 56.91 & 62.51 & \fwbest{0.17} & \fwsecond{52.61} & \fwsecond{59.16} & \fwbest{0.21}\\
\AppMQwenPlus & 48.44 & 54.77 & 0.38 & 37.58 & 45.73 & 0.43 & 47.24 & 53.11 & 0.35 & 44.42 & 51.20 & 0.39\\

}
Under the choice interface, several models attain high scores at individual
fire-evolution stages while exhibiting much weaker performance elsewhere.
The relative ranking also changes across T1--T3, which indicates that model
scale or overall benchmark accuracy alone does not determine competence at a
specific fire stage. In the reported I-track results, Coupled Propagation has
a lower model-level mean Acc. than Localized Onset and Critical Transition,
while the S-track is more even across the three categories.
\AppFTable{\textbf{Open-report evaluation along the fire scenario task axis.} Models are evaluated on the real-world-aligned subset under the structured-observation track (S-track) and multimodal image-observation track (I-track). Each dimension reports accuracy (Acc), macro-F1 (F1), and Brier score; Average is the equal-weight mean across T1--T3.}
{tab:rwa_open_fire_axis}
{\AppFTHeader{Acc}{F1}{Brier}}
{%
\AppFSHead
\AppFSmall
\AppMLlamaSmall & 43.18 & 34.09 & 0.28 & 23.58 & 20.16 & 0.47 & 45.22 & 40.07 & 0.28 & 37.33 & 31.44 & 0.34\\
\AppMQwenSmall & 45.54 & 37.86 & 0.36 & 47.15 & 39.07 & 0.32 & 46.50 & 43.15 & 0.36 & 46.40 & 40.03 & 0.35\\
\AppMGemmaSmall & 48.26 & 37.67 & 0.29 & 30.89 & 26.61 & 0.46 & 41.83 & 34.20 & 0.32 & 40.33 & 32.83 & 0.36\\
\AppFMid
\AppMGemmaMid & 44.15 & 33.71 & 0.33 & 46.34 & 39.31 & 0.34 & 57.57 & 50.75 & 0.25 & 49.35 & 41.26 & 0.31\\
\AppMQwenMid & 43.03 & 36.64 & 0.40 & 47.97 & 39.50 & 0.33 & 61.49 & 54.51 & 0.21 & 50.83 & 43.55 & 0.31\\
\AppFLarge
\AppMLlamaLarge & 56.21 & 48.50 & \fwbest{0.20} & 49.59 & 41.03 & 0.22 & \fwbest{66.05} & 59.36 & \fwbest{0.16} & 57.28 & 49.63 & \fwbest{0.19}\\
\AppMQwenNext & 51.95 & 45.28 & 0.35 & 57.72 & 51.23 & 0.26 & 59.62 & 52.78 & 0.27 & 56.43 & 49.76 & 0.29\\
\AppMGLM & 46.87 & 40.72 & 0.35 & 52.85 & 47.43 & 0.27 & 60.17 & 53.93 & 0.22 & 53.30 & 47.36 & 0.28\\
\AppFHuge
\AppMQwenHuge & 50.72 & 41.20 & 0.27 & 44.72 & 38.62 & 0.33 & 47.33 & 40.64 & 0.28 & 47.59 & 40.15 & 0.29\\
\AppMDeepSeek & \fwbest{57.90} & \fwsecond{50.83} & \fwsecond{0.25} & 43.90 & 37.05 & 0.34 & 53.34 & 47.94 & 0.27 & 51.71 & 45.27 & 0.29\\
\AppMKimi & 52.77 & 47.37 & 0.30 & 46.34 & 40.00 & 0.28 & 58.29 & 51.80 & 0.20 & 52.47 & 46.39 & 0.26\\
\AppFClosed
\AppMGPT & 52.10 & 46.54 & 0.35 & \fwbest{66.67} & \fwbest{58.14} & \fwsecond{0.17} & \fwsecond{65.90} & \fwbest{60.47} & \fwsecond{0.18} & \fwbest{61.56} & \fwbest{55.05} & 0.24\\
\AppMGemini & 52.26 & 44.74 & 0.31 & 50.41 & 42.45 & 0.25 & 60.71 & 53.44 & 0.22 & 54.46 & 46.88 & 0.26\\
\AppMClaude & \fwsecond{57.13} & \fwbest{51.10} & 0.27 & 59.35 & 50.16 & \fwbest{0.16} & 58.43 & 53.95 & 0.22 & \fwsecond{58.30} & 51.74 & \fwsecond{0.22}\\
\AppMQwenPlus & 49.33 & 42.53 & 0.35 & \fwsecond{60.16} & \fwsecond{53.41} & 0.23 & 65.11 & \fwsecond{59.67} & 0.22 & 58.20 & \fwsecond{51.87} & 0.27\\
\midrule
\AppFIHead
\AppFISmall
\AppMInternSmall & -- & -- & -- & -- & -- & -- & -- & -- & -- & -- & -- & --\\
\AppMQwenVLSmall & 39.49 & 35.69 & 0.51 & 36.20 & 29.40 & 0.48 & \fwbest{53.39} & \fwbest{51.05} & 0.39 & 43.03 & 38.71 & 0.46\\
\AppFIMid
\AppMInternMid & -- & -- & -- & -- & -- & -- & -- & -- & -- & -- & -- & --\\
\AppMQwenVLMid & \fwbest{53.08} & \fwbest{46.63} & \fwbest{0.31} & \fwsecond{44.01} & \fwsecond{35.97} & 0.35 & \fwsecond{52.20} & \fwsecond{47.64} & 0.30 & \fwbest{49.76} & \fwbest{43.41} & \fwsecond{0.32}\\
\AppFIHuge
\AppMInternLarge & -- & -- & -- & -- & -- & -- & -- & -- & -- & -- & -- & --\\
\AppMQwenVLHuge & 33.33 & 28.67 & 0.53 & 35.94 & 29.13 & 0.42 & 48.60 & 45.50 & 0.33 & 39.29 & 34.43 & 0.43\\
\AppFIClosed
\AppMGPT & 49.23 & 44.46 & \fwsecond{0.31} & \fwbest{44.79} & \fwbest{38.54} & \fwsecond{0.29} & 49.16 & 44.22 & \fwsecond{0.25} & \fwsecond{47.73} & \fwsecond{42.41} & \fwbest{0.28}\\
\AppMGemini & 45.64 & 41.18 & 0.41 & 42.97 & 32.93 & 0.34 & 49.89 & 45.31 & 0.35 & 46.17 & 39.81 & 0.37\\
\AppMClaude & 37.69 & 33.54 & 0.44 & 44.01 & 35.77 & \fwbest{0.28} & 44.42 & 39.32 & \fwbest{0.24} & 42.04 & 36.21 & 0.32\\
\AppMQwenPlus & \fwsecond{51.03} & \fwsecond{45.54} & 0.37 & 40.89 & 33.03 & 0.40 & 45.54 & 43.02 & 0.38 & 45.82 & 40.53 & 0.38\\

}
Open reporting makes these stage-specific differences more pronounced.
The strongest systems can still achieve high performance at particular
fire-evolution stages, yet the reported model-level means differ across
Localized Onset, Coupled Propagation, and Critical Transition. Critical
Transition attains the highest mean Acc. on both observation tracks, while
Coupled Propagation remains lower on the I-track. The multimodal track further
changes the relative ranking of several models, showing that visual grounding
and stage-level fire reasoning remain only partially coupled in current
systems.

Overall, the real-world-aligned evaluation reaches the same qualitative
conclusion as the controlled simulation study from a complementary source of
evidence.
Current foundation models exhibit localized strengths rather than a single
general fire-world reasoning capability.
Performance depends jointly on the physical capability being tested, the
fire scenario task, the observation modality, and whether the answer is
candidate-conditioned or generated openly.

\subsection{Open-Report Explanation Metrics}
\label{app:open_report_explanation_metrics}

Answer correctness alone does not determine whether an open report is
physically well supported.
We therefore provide the three explanation-oriented metrics that are defined
only for the open-report interface: Evidence-F1 (Evi-F1), Mechanism Alignment
(Mech), and Gold-Linked Support (GLS).
Evi-F1 measures whether the generated report recovers the evidence elements
required by the Gold annotation, Mech evaluates agreement with the target
physical mechanism, and GLS measures whether the final report is supported by
Gold-linked evidence.
Higher values are better for all three metrics.

Tables~\ref{tab:sim_open_physical_explanation} and
\ref{tab:sim_open_fire_explanation} report these metrics for controlled
simulations, while Tables~\ref{tab:real_open_physical_explanation} and
\ref{tab:real_open_fire_explanation} provide the corresponding
real-world-aligned results.
These diagnostics are reported separately from Acc., F1, and Brier because
they characterize different properties of an open physical-world report
rather than another form of answer accuracy.

\AppCTable{\textbf{Open-report explanation metrics along the physical capability axis.} Models are evaluated on the controlled simulation subset. Each dimension reports Evidence-F1 (Evi-F1), Mechanism Alignment (Mech), and Gold-Linked Support (GLS); Average is the equal-weight mean across P1--P5. Higher is better for all three metrics.}
{tab:sim_open_physical_explanation}
{\AppCPHeader{Evi-F1}{Mech}{GLS}}
{%
\AppCSHead
\AppCSmall
\AppMLlamaSmall & 57.16 & 32.34 & 16.73 & 52.59 & 13.55 & 27.53 & 64.15 & 31.07 & 24.54 & 31.04 & 11.76 & 30.00 & 12.03 & 15.59 & 20.35 & 43.39 & 20.86 & 23.83\\
\AppMQwenSmall & \fwbest{68.70} & 39.75 & 20.06 & 49.63 & 19.74 & 28.70 & 65.96 & 29.61 & 35.09 & 35.95 & 10.34 & \fwbest{41.88} & 10.50 & 16.08 & 23.64 & 46.15 & 23.10 & 29.87\\
\AppMGemmaSmall & 64.09 & 33.68 & 18.38 & 51.43 & 16.68 & 36.79 & 69.85 & 32.71 & 26.86 & \fwsecond{43.84} & 13.66 & 26.00 & 14.40 & 15.88 & 19.57 & 48.72 & 22.52 & 25.52\\
\AppCMid
\AppMGemmaMid & 65.27 & 35.23 & 21.23 & 51.40 & 16.26 & 36.30 & 65.23 & 26.04 & 29.94 & 39.84 & 12.93 & 29.39 & 12.96 & 15.46 & 24.42 & 46.94 & 21.18 & 28.26\\
\AppMQwenMid & 61.59 & 36.69 & 21.21 & 51.65 & 23.52 & 35.00 & 61.75 & 31.70 & 32.53 & 39.81 & 16.81 & 39.09 & 13.11 & 18.86 & 28.29 & 45.58 & 25.52 & 31.22\\
\AppCLarge
\AppMLlamaLarge & 62.77 & 36.26 & 16.12 & \fwsecond{55.68} & 13.50 & 29.94 & 68.46 & 33.64 & 34.17 & 39.26 & 14.53 & 31.21 & 13.30 & 18.91 & 21.71 & 47.89 & 23.37 & 26.63\\
\AppMQwenNext & 61.89 & 40.85 & 20.61 & 49.12 & 21.17 & 31.42 & 60.05 & 29.82 & 30.30 & 41.00 & 14.47 & 30.45 & 12.20 & 20.80 & 23.06 & 44.85 & 25.42 & 27.17\\
\AppMGLM & 62.63 & 40.16 & 18.57 & 50.53 & 22.66 & 30.56 & 61.02 & 32.29 & 31.89 & 40.28 & 19.61 & 38.24 & 9.55 & 19.12 & 25.39 & 44.80 & 26.77 & 28.93\\
\AppCHuge
\AppMQwenHuge & 63.70 & 38.33 & 21.47 & 50.49 & 20.81 & 29.57 & 61.15 & 29.96 & 32.89 & 33.68 & 15.05 & 38.67 & 12.02 & 18.40 & 31.20 & 44.21 & 24.51 & 30.76\\
\AppMDeepSeek & 63.81 & \fwsecond{42.73} & 19.66 & \fwbest{55.78} & 26.60 & 38.02 & 60.59 & 35.60 & \fwsecond{36.13} & 42.24 & 20.41 & \fwsecond{40.48} & 15.16 & 20.18 & \fwbest{34.88} & 47.52 & 29.10 & \fwbest{33.83}\\
\AppMKimi & 57.37 & 36.98 & 20.38 & 49.28 & 24.79 & \fwsecond{39.44} & 53.53 & 30.54 & 29.62 & 39.68 & 19.75 & 29.18 & 8.03 & 21.18 & 27.33 & 41.58 & 26.65 & 29.19\\
\AppCClosed
\AppMGPT & 65.26 & 32.93 & \fwsecond{23.79} & 52.28 & 22.63 & 38.46 & \fwbest{71.96} & 32.79 & 29.62 & 41.60 & \fwsecond{21.19} & 37.24 & \fwbest{19.27} & 17.43 & \fwsecond{33.14} & \fwsecond{50.07} & 25.39 & \fwsecond{32.45}\\
\AppMGemini & \fwsecond{65.48} & \fwbest{45.20} & 21.17 & 55.37 & \fwsecond{26.75} & 38.40 & \fwsecond{70.37} & \fwbest{39.71} & \fwbest{36.57} & \fwbest{45.34} & 20.79 & 27.52 & 16.16 & \fwsecond{21.46} & 29.07 & \fwbest{50.54} & \fwsecond{30.78} & 30.55\\
\AppMClaude & 58.53 & 40.34 & \fwbest{25.66} & 48.90 & \fwbest{30.58} & \fwbest{40.74} & 60.83 & \fwsecond{37.28} & 28.74 & 40.22 & \fwbest{26.35} & 31.55 & 11.36 & 21.45 & 23.06 & 43.97 & \fwbest{31.20} & 29.95\\
\AppMQwenPlus & 61.83 & 40.18 & 20.48 & 51.10 & 23.79 & 31.73 & 62.04 & 33.42 & 35.69 & 41.31 & 18.60 & 37.94 & \fwsecond{17.02} & \fwbest{21.79} & 27.91 & 46.66 & 27.56 & 30.75\\
\midrule
\AppCIHead
\AppCISmall
\AppMInternSmall & 47.11 & 29.19 & 13.85 & 23.02 & 14.55 & 15.19 & 43.44 & 19.45 & 31.90 & 15.25 & 1.88 & 17.52 & 0.00 & 13.15 & 12.50 & 25.76 & 15.64 & 18.19\\
\AppMQwenVLSmall & 53.36 & 31.23 & 20.87 & 24.31 & 19.52 & 12.16 & 41.77 & 16.78 & 34.01 & 15.30 & 4.44 & 15.11 & 0.00 & 15.99 & 34.17 & 26.95 & 17.59 & 23.26\\
\AppCIMid
\AppMInternMid & 56.51 & 37.03 & 27.96 & 23.77 & 22.37 & 18.08 & 43.58 & 19.69 & 33.41 & 15.68 & 13.71 & \fwbest{33.72} & 0.00 & \fwsecond{20.97} & 19.58 & 27.91 & 22.75 & 26.55\\
\AppMQwenVLMid & 54.77 & \fwbest{42.98} & \fwsecond{48.02} & \fwsecond{27.13} & \fwbest{31.71} & \fwbest{29.65} & \fwsecond{51.86} & \fwsecond{28.84} & \fwsecond{36.27} & 14.96 & \fwbest{22.16} & \fwsecond{27.10} & 0.09 & 20.55 & 33.75 & 29.76 & \fwbest{29.25} & \fwbest{34.96}\\
\AppCIHuge
\AppMInternLarge & \fwbest{59.65} & 35.65 & 25.40 & 23.44 & 21.11 & 26.04 & 43.57 & 17.11 & 25.94 & 16.26 & 13.03 & 23.36 & 0.00 & 18.62 & \fwsecond{38.33} & 28.58 & 21.10 & 27.81\\
\AppMQwenVLHuge & 31.03 & 20.48 & 13.16 & 18.50 & 18.36 & 18.47 & 36.20 & 16.38 & 22.02 & 9.30 & 9.44 & 15.03 & 0.00 & 13.79 & 12.50 & 19.01 & 15.69 & 16.24\\
\AppCIClosed
\AppMGPT & \fwsecond{57.97} & 35.13 & 34.10 & 24.62 & 22.83 & 27.35 & 49.84 & 20.20 & \fwsecond{36.27} & \fwbest{20.28} & 16.87 & 20.40 & 7.04 & 14.77 & 17.50 & \fwsecond{31.95} & 21.96 & 27.12\\
\AppMGemini & 48.65 & 23.50 & 27.30 & 26.57 & 21.46 & 25.71 & 36.92 & 20.14 & 35.52 & \fwsecond{19.45} & 13.32 & 20.40 & 10.17 & 19.21 & 37.50 & 28.35 & 19.53 & 29.29\\
\AppMClaude & 51.01 & \fwsecond{38.37} & \fwbest{49.45} & \fwbest{29.06} & \fwsecond{29.15} & 24.52 & \fwbest{53.06} & \fwbest{31.15} & 35.67 & 17.97 & \fwsecond{22.11} & 25.23 & \fwbest{14.41} & 18.96 & 33.75 & \fwbest{33.10} & \fwsecond{27.95} & \fwsecond{33.72}\\
\AppMQwenPlus & 51.25 & 35.95 & 27.59 & 26.06 & 25.59 & \fwsecond{27.81} & 51.13 & 28.07 & \fwbest{37.33} & 18.62 & 11.95 & 26.95 & \fwsecond{11.92} & \fwbest{21.83} & \fwbest{45.42} & 31.80 & 24.68 & 33.02\\
}
\paragraph{Controlled-simulation explanation quality.}
The physical-axis results reveal a substantial separation between evidence
recovery and mechanism reconstruction.
For many models and capability levels, Evi-F1 is considerably higher than
Mech, indicating that identifying relevant physical observations is easier
than correctly connecting them through the underlying causal mechanism.
This gap is visible even for strong models and becomes particularly important
at the higher-level capability dimensions.

GLS provides a complementary view.
A model may recover individual Gold evidence items without consistently using
them to support the final physical commitment, and conversely a report with
reasonable answer accuracy may still have weak explicit evidence linkage.
The three explanation metrics should therefore not be interpreted as
redundant measures of the same property.
\AppFTable{\textbf{Open-report explanation metrics along the fire scenario task axis.} Models are evaluated on the controlled simulation subset. Each dimension reports Evidence-F1 (Evi-F1), Mechanism Alignment (Mech), and Gold-Linked Support (GLS); Average is the equal-weight mean across T1--T3. Higher is better for all three metrics.}
{tab:sim_open_fire_explanation}
{\AppFTHeader{Evi-F1}{Mech}{GLS}}
{%
\AppFSHead
\AppFSmall
\AppMLlamaSmall & 46.56 & 12.70 & 19.10 & 47.66 & 27.58 & 25.80 & 40.98 & 17.85 & 14.30 & 45.07 & 19.38 & 19.73\\
\AppMQwenSmall & 46.53 & 17.17 & 22.60 & 51.38 & 23.35 & \fwsecond{35.00} & 50.69 & 24.86 & 19.50 & 49.53 & 21.79 & 25.70\\
\AppMGemmaSmall & 52.96 & 14.91 & 23.80 & 54.63 & 26.63 & 22.00 & 49.54 & 23.67 & 18.20 & 52.38 & 21.74 & 21.33\\
\AppFMid
\AppMGemmaMid & 49.68 & 14.29 & 24.00 & 51.71 & 23.09 & 26.30 & 49.82 & 22.02 & 21.40 & 50.40 & 19.80 & 23.90\\
\AppMQwenMid & 48.90 & 21.68 & 26.00 & 50.71 & 28.47 & 30.50 & 46.69 & 27.02 & 21.40 & 48.77 & 25.72 & 25.97\\
\AppFLarge
\AppMLlamaLarge & \fwsecond{54.24} & 14.90 & 20.50 & 51.26 & 30.58 & 32.30 & 47.72 & 23.60 & 14.40 & 51.07 & 23.03 & 22.40\\
\AppMQwenNext & 48.17 & 18.71 & 23.50 & 49.79 & 25.74 & 23.50 & 46.13 & 28.06 & 20.60 & 48.03 & 24.17 & 22.53\\
\AppMGLM & 49.54 & 20.00 & 23.90 & 49.74 & 32.81 & 28.50 & 45.92 & 25.27 & 20.20 & 48.40 & 26.03 & 24.20\\
\AppFHuge
\AppMQwenHuge & 44.67 & 19.77 & 23.10 & 49.47 & 26.44 & 29.60 & 47.62 & 25.42 & \fwsecond{23.10} & 47.25 & 23.88 & 25.27\\
\AppMDeepSeek & 52.46 & \fwsecond{23.76} & \fwbest{27.00} & 50.99 & 32.32 & \fwbest{36.90} & 49.22 & 29.16 & 19.80 & 50.89 & 28.41 & \fwbest{27.90}\\
\AppMKimi & 47.65 & 20.27 & \fwsecond{26.90} & 45.75 & 28.36 & 26.80 & 42.65 & 27.62 & 21.70 & 45.35 & 25.42 & 25.13\\
\AppFClosed
\AppMGPT & 50.66 & 18.18 & 26.00 & \fwbest{57.21} & 29.95 & 33.80 & \fwbest{54.11} & 28.06 & 17.20 & \fwsecond{53.99} & 25.40 & 25.67\\
\AppMGemini & \fwbest{54.74} & 22.98 & 25.70 & \fwsecond{56.98} & \fwsecond{34.45} & 32.50 & \fwsecond{51.36} & \fwbest{32.88} & 22.90 & \fwbest{54.36} & \fwsecond{30.10} & 27.03\\
\AppMClaude & 47.45 & \fwbest{27.13} & 25.90 & 50.55 & \fwbest{35.89} & 32.30 & 43.92 & 29.40 & \fwbest{24.50} & 47.31 & \fwbest{30.81} & \fwsecond{27.57}\\
\AppMQwenPlus & 49.27 & 21.68 & 24.90 & 51.25 & 29.91 & 32.40 & 48.34 & \fwsecond{30.98} & 22.20 & 49.62 & 27.52 & 26.50\\
\midrule
\AppFIHead
\AppFISmall
\AppMInternSmall & -- & -- & -- & -- & -- & -- & -- & -- & -- & -- & -- & --\\
\AppMQwenVLSmall & 23.84 & 16.70 & 9.50 & 31.36 & 11.34 & 26.90 & 34.29 & 21.23 & 25.40 & 29.83 & 16.42 & 20.60\\
\AppFIMid
\AppMInternMid & -- & -- & -- & -- & -- & -- & -- & -- & -- & -- & -- & --\\
\AppMQwenVLMid & 25.67 & \fwbest{28.11} & 23.30 & \fwsecond{37.73} & \fwsecond{29.96} & \fwbest{30.80} & 37.45 & 29.00 & \fwsecond{43.10} & 33.62 & \fwbest{29.02} & \fwsecond{32.40}\\
\AppFIHuge
\AppMInternLarge & -- & -- & -- & -- & -- & -- & -- & -- & -- & -- & -- & --\\
\AppMQwenVLHuge & 18.47 & 17.17 & 14.80 & 24.88 & 14.49 & 16.60 & 22.03 & 15.42 & 12.90 & 21.79 & 15.69 & 14.77\\
\AppFIClosed
\AppMGPT & 25.98 & 20.01 & \fwsecond{26.60} & 36.64 & 20.25 & 29.50 & \fwbest{40.28} & 28.06 & 28.30 & \fwsecond{34.30} & 22.77 & 28.13\\
\AppMGemini & \fwsecond{27.08} & 17.74 & 24.90 & 28.27 & 19.25 & 29.50 & 35.29 & 22.01 & 30.80 & 30.21 & 19.67 & 28.40\\
\AppMClaude & \fwbest{28.12} & \fwsecond{24.81} & 26.60 & \fwbest{38.40} & \fwbest{30.34} & \fwsecond{30.00} & \fwsecond{38.30} & \fwbest{31.63} & \fwbest{44.00} & \fwbest{34.94} & \fwsecond{28.93} & \fwbest{33.53}\\
\AppMQwenPlus & 26.19 & 21.60 & \fwbest{32.60} & 37.28 & 22.68 & 28.70 & 37.59 & \fwsecond{31.05} & 33.80 & 33.69 & 25.11 & 31.70\\

}
The fire-task decomposition exhibits the same general phenomenon.
Evidence grounding, mechanism alignment, and Gold-linked support vary
differently across Localized Onset, Coupled Propagation, and Critical
Transition. In particular, mechanism alignment remains substantially more
difficult than recovering surface-level evidence for many model--stage
combinations. This helps explain why improvements in open-report answer
accuracy do not always imply an equivalent improvement in physically
accountable reasoning.

These results complement the main benchmark metrics by showing where an
apparently correct report obtains its support.
FireWorldBench therefore distinguishes between producing the right final
commitment, citing relevant evidence, recovering the correct mechanism, and
linking the final claim to evidence sanctioned by the benchmark annotation.
\paragraph{Real-world-aligned explanation quality.}
The same diagnostics are evaluated on the real-world-aligned subset to test
whether explanation failures persist outside the controlled simulation
setting.
The results remain highly heterogeneous across models and capability levels,
but the distinction between evidence identification and mechanism
understanding is again clear.
\AppCTable{\textbf{Open-report explanation metrics along the physical capability axis.} Models are evaluated on the real-world-aligned subset. Each dimension reports Evidence-F1 (Evi-F1), Mechanism Alignment (Mech), and Gold-Linked Support (GLS); Average is the equal-weight mean across P1--P5. Higher is better for all three metrics.}
{tab:real_open_physical_explanation}
{\AppCPHeader{Evi-F1}{Mech}{GLS}}
{%
\AppCSHead
\AppCSmall
\AppMLlamaSmall & 50.70 & 34.30 & 23.26 & 54.36 & 13.86 & 28.64 & 85.66 & 30.91 & 27.27 & 41.83 & 4.78 & 16.67 & 18.19 & 15.37 & 16.67 & 50.15 & 19.84 & 22.50\\
\AppMQwenSmall & \fwsecond{65.34} & 43.53 & 10.85 & 56.10 & 21.61 & 28.64 & \fwsecond{86.51} & 36.36 & \fwbest{60.61} & 45.97 & 8.75 & 38.33 & 25.94 & 14.86 & 22.22 & 55.97 & 25.02 & 32.13\\
\AppMGemmaSmall & 63.52 & 38.90 & 28.68 & 57.30 & 18.58 & 32.05 & 84.87 & 33.03 & 36.36 & 48.31 & 13.49 & 23.33 & 26.63 & 16.77 & 20.37 & 56.13 & 24.15 & 28.16\\
\AppCMid
\AppMGemmaMid & \fwbest{68.03} & 43.88 & 26.87 & 60.78 & 18.32 & 34.32 & 79.35 & 32.12 & 45.45 & 46.02 & 11.78 & 37.50 & \fwsecond{27.83} & 15.17 & 16.67 & \fwsecond{56.40} & 24.25 & 32.16\\
\AppMQwenMid & 63.76 & 36.37 & 28.42 & 54.44 & 25.18 & 31.59 & 81.03 & \fwbest{41.52} & \fwsecond{56.06} & 49.14 & 12.19 & 30.83 & 24.88 & 17.87 & 20.37 & 54.65 & 26.63 & 33.45\\
\AppCLarge
\AppMLlamaLarge & 63.19 & 38.58 & 23.77 & \fwbest{65.35} & 15.84 & 36.36 & \fwbest{90.35} & 35.45 & 39.39 & 53.30 & 8.67 & \fwsecond{42.50} & 26.80 & 17.12 & 25.93 & \fwbest{59.80} & 23.13 & 33.59\\
\AppMQwenNext & 63.00 & 41.28 & 20.93 & 52.14 & 27.46 & 43.41 & 71.79 & 32.73 & 50.00 & 47.09 & 17.14 & 38.33 & 17.59 & 20.62 & 25.93 & 50.32 & 27.85 & 35.72\\
\AppMGLM & 59.66 & 44.80 & 27.13 & 56.24 & 23.44 & 33.64 & 77.96 & 33.03 & 39.39 & 51.59 & 22.45 & 35.00 & \fwbest{29.71} & 19.78 & \fwsecond{31.48} & 55.03 & 28.70 & 33.33\\
\AppCHuge
\AppMQwenHuge & 58.36 & 38.95 & 29.20 & 60.57 & 24.03 & 37.05 & 75.69 & 32.12 & 46.97 & 35.21 & 12.43 & 41.67 & 16.81 & 19.20 & 29.63 & 49.33 & 25.35 & 36.90\\
\AppMDeepSeek & 53.15 & \fwsecond{45.57} & 14.73 & 56.05 & 23.73 & 44.09 & 64.32 & \fwsecond{38.48} & 42.42 & 50.60 & 21.65 & 30.83 & 20.74 & 22.61 & 22.22 & 48.97 & \fwsecond{30.41} & 30.86\\
\AppMKimi & 52.33 & 37.18 & 19.64 & 52.92 & 26.66 & 44.32 & 57.50 & 28.48 & 39.39 & 46.18 & \fwsecond{27.15} & 24.17 & 16.87 & \fwbest{28.38} & 25.93 & 45.16 & 29.57 & 30.69\\
\AppCClosed
\AppMGPT & 56.95 & 36.51 & 29.46 & 59.68 & 26.81 & \fwbest{45.91} & 83.02 & 29.39 & 50.00 & \fwbest{60.82} & 24.12 & 41.67 & 20.10 & 18.31 & \fwbest{42.59} & 56.11 & 27.03 & \fwbest{41.93}\\
\AppMGemini & 61.14 & \fwbest{47.38} & \fwsecond{29.97} & \fwsecond{62.23} & 25.08 & 42.50 & 79.34 & 37.27 & 46.97 & \fwsecond{53.77} & 16.49 & 35.83 & 22.62 & 18.82 & 27.78 & 55.82 & 29.01 & 36.61\\
\AppMClaude & 53.45 & 42.87 & \fwbest{34.11} & 48.70 & \fwbest{30.22} & \fwsecond{45.45} & 65.17 & 37.27 & 50.00 & 45.00 & \fwbest{27.71} & 41.67 & 18.04 & \fwsecond{23.42} & 22.22 & 46.07 & \fwbest{32.30} & 38.69\\
\AppMQwenPlus & 61.50 & 40.34 & 28.68 & 56.49 & \fwsecond{27.47} & 39.55 & 72.64 & 32.12 & 50.00 & 47.46 & 18.43 & \fwbest{45.00} & 19.78 & 19.89 & \fwsecond{31.48} & 51.57 & 27.65 & \fwsecond{38.94}\\
\midrule
\AppCIHead
\AppCISmall
\AppMInternSmall & 47.28 & 26.93 & 7.32 & 19.56 & 15.37 & 20.00 & 43.57 & 18.78 & 23.33 & 19.78 & 3.48 & 18.18 & 0.00 & 13.89 & 15.00 & 26.04 & 15.69 & 16.77\\
\AppMQwenVLSmall & 54.34 & 28.73 & 16.53 & 25.73 & 17.29 & \fwbest{28.15} & 43.33 & 22.37 & 33.33 & \fwsecond{25.62} & 13.15 & 17.42 & 0.00 & 19.13 & 25.00 & 29.80 & 20.13 & 24.09\\
\AppCIMid
\AppMInternMid & 55.42 & 30.18 & 11.11 & 27.34 & 22.84 & 24.81 & 45.66 & 23.58 & 35.83 & 19.38 & 16.82 & \fwsecond{40.15} & 0.00 & 23.37 & 15.00 & 29.56 & 23.36 & 25.38\\
\AppMQwenVLMid & 55.37 & \fwbest{37.92} & \fwbest{28.18} & \fwbest{32.92} & \fwsecond{31.45} & 26.30 & \fwsecond{53.96} & \fwsecond{30.47} & \fwsecond{40.42} & 25.15 & \fwsecond{21.45} & \fwbest{44.70} & 4.76 & 24.49 & \fwsecond{35.00} & \fwsecond{34.43} & \fwsecond{29.16} & \fwbest{34.92}\\
\AppCIHuge
\AppMInternLarge & \fwbest{61.84} & 33.48 & 20.87 & 28.59 & 15.36 & 24.44 & 48.80 & 20.10 & 25.83 & 21.65 & 13.24 & \fwsecond{40.15} & 3.33 & 22.83 & \fwbest{40.00} & 32.84 & 21.00 & 30.26\\
\AppMQwenVLHuge & 47.82 & 33.11 & 22.49 & 23.25 & 21.78 & 21.11 & 34.00 & 18.66 & 16.25 & 23.86 & 16.66 & 21.21 & 3.10 & 22.04 & 25.00 & 26.41 & 22.45 & 21.21\\
\AppCIClosed
\AppMGPT & 53.86 & \fwsecond{35.14} & 20.05 & 23.66 & 21.60 & 26.30 & 47.88 & 20.60 & 31.67 & 25.52 & 15.59 & 27.27 & 3.43 & 19.46 & 30.00 & 30.87 & 22.48 & 27.06\\
\AppMGemini & 53.06 & 34.52 & 15.45 & 25.01 & 25.09 & 25.93 & 40.75 & 24.73 & 36.25 & \fwbest{27.31} & 15.89 & \fwsecond{40.15} & \fwsecond{7.60} & \fwbest{26.99} & \fwsecond{35.00} & 30.75 & 25.44 & \fwsecond{30.56}\\
\AppMClaude & 52.99 & 35.13 & \fwsecond{23.04} & \fwsecond{30.58} & \fwbest{31.67} & 22.59 & \fwbest{54.84} & \fwbest{35.27} & \fwbest{42.92} & 22.83 & \fwbest{22.21} & 31.06 & \fwbest{16.17} & 24.22 & 20.00 & \fwbest{35.48} & \fwbest{29.70} & 27.92\\
\AppMQwenPlus & \fwsecond{55.43} & 32.44 & 18.43 & 28.24 & 25.94 & \fwsecond{27.04} & 49.46 & 27.47 & 37.50 & 24.94 & 12.64 & 32.58 & 5.67 & \fwsecond{25.08} & 20.00 & 32.75 & 24.71 & 27.11\\
}
For the physical capability axis, several models achieve relatively strong
Evi-F1 while obtaining much lower Mechanism Alignment.
The gap is particularly informative at the more demanding capability levels:
a model can identify observations associated with the target phenomenon
without recovering the causal interpretation required to explain that
phenomenon correctly.
GLS further shows that correct evidence extraction does not guarantee that
the generated answer is explicitly grounded in the appropriate Gold-linked
support.
\AppFTable{\textbf{Open-report explanation metrics along the fire scenario task axis.} Models are evaluated on the real-world-aligned subset. Each dimension reports Evidence-F1 (Evi-F1), Mechanism Alignment (Mech), and Gold-Linked Support (GLS); Average is the equal-weight mean across T1--T3. Higher is better for all three metrics.}
{tab:real_open_fire_explanation}
{\AppFTHeader{Evi-F1}{Mech}{GLS}}
{%
\AppFSHead
\AppFSmall
\AppMLlamaSmall & 49.24 & 9.93 & 27.60 & 67.04 & 20.38 & 17.90 & 41.11 & 28.72 & 21.30 & 52.46 & 19.68 & 22.27\\
\AppMQwenSmall & 53.64 & 17.45 & 33.70 & 66.44 & 23.58 & 47.20 & \fwsecond{53.71} & 35.07 & 14.20 & 57.93 & 25.37 & 31.70\\
\AppMGemmaSmall & 56.03 & 16.09 & 31.90 & 65.34 & 25.31 & 26.00 & 52.64 & 32.37 & 26.20 & 58.00 & 24.59 & 28.03\\
\AppFMid
\AppMGemmaMid & 54.78 & 15.04 & 36.10 & 65.86 & 24.55 & 40.70 & \fwbest{56.17} & 35.41 & 23.90 & 58.94 & 25.00 & 33.57\\
\AppMQwenMid & 52.59 & 19.76 & 29.10 & 66.47 & 29.86 & 48.00 & 52.29 & 30.91 & 26.00 & 57.12 & 26.84 & 34.37\\
\AppFLarge
\AppMLlamaLarge & \fwbest{62.82} & 13.15 & 36.40 & \fwsecond{71.01} & 23.63 & 43.90 & 52.46 & 32.25 & 24.40 & \fwbest{62.10} & 23.01 & 34.90\\
\AppMQwenNext & 50.54 & 23.49 & 37.60 & 60.28 & 26.50 & 51.20 & 49.60 & 35.18 & 22.40 & 53.47 & 28.39 & 37.07\\
\AppMGLM & 54.31 & 21.13 & 29.40 & 66.43 & 31.27 & 44.70 & 50.82 & 37.42 & 28.40 & 57.19 & 29.94 & 34.17\\
\AppFHuge
\AppMQwenHuge & 50.05 & 19.64 & \fwbest{40.50} & 60.61 & 24.01 & 41.50 & 46.10 & 33.12 & 29.30 & 52.25 & 25.59 & 37.10\\
\AppMDeepSeek & 56.03 & 21.01 & 37.00 & 55.21 & 33.93 & 41.50 & 43.59 & \fwsecond{38.80} & 16.90 & 51.61 & \fwsecond{31.25} & 31.80\\
\AppMKimi & 51.93 & 22.79 & 34.10 & 50.38 & \fwsecond{34.25} & 38.20 & 41.86 & 34.58 & 21.50 & 48.06 & 30.54 & 31.27\\
\AppFClosed
\AppMGPT & 57.48 & \fwsecond{24.44} & \fwsecond{37.70} & \fwbest{76.80} & 29.32 & \fwbest{56.90} & 46.08 & 31.14 & \fwbest{33.30} & \fwsecond{60.12} & 28.30 & \fwbest{42.63}\\
\AppMGemini & \fwsecond{60.70} & 22.68 & 35.90 & 65.58 & 27.05 & 48.80 & 49.78 & \fwbest{38.95} & 29.30 & 58.69 & 29.56 & 38.00\\
\AppMClaude & 48.30 & \fwbest{26.74} & 37.40 & 54.57 & \fwbest{37.07} & \fwsecond{56.90} & 43.00 & 37.13 & \fwsecond{30.60} & 48.62 & \fwbest{33.65} & \fwsecond{41.63}\\
\AppMQwenPlus & 53.26 & 22.58 & 37.00 & 61.46 & 28.89 & 54.50 & 49.19 & 34.31 & 29.50 & 54.64 & 28.59 & 40.33\\
\midrule
\AppFIHead
\AppFISmall
\AppMInternSmall & -- & -- & -- & -- & -- & -- & -- & -- & -- & -- & -- & --\\
\AppMQwenVLSmall & 25.94 & 14.03 & 22.10 & 36.44 & 20.93 & 30.20 & 36.52 & 25.58 & 19.30 & 32.97 & 20.18 & 23.87\\
\AppFIMid
\AppMInternMid & -- & -- & -- & -- & -- & -- & -- & -- & -- & -- & -- & --\\
\AppMQwenVLMid & \fwbest{30.87} & \fwsecond{24.84} & \fwbest{31.00} & \fwsecond{42.81} & \fwsecond{30.67} & \fwsecond{43.00} & 38.78 & \fwbest{33.52} & \fwbest{30.40} & \fwbest{37.49} & \fwsecond{29.68} & \fwbest{34.80}\\
\AppFIHuge
\AppMInternLarge & -- & -- & -- & -- & -- & -- & -- & -- & -- & -- & -- & --\\
\AppMQwenVLHuge & 23.21 & 18.15 & 18.70 & 30.43 & 20.00 & 20.60 & 33.15 & 29.48 & \fwsecond{23.30} & 28.93 & 22.54 & 20.87\\
\AppFIClosed
\AppMGPT & 24.99 & 17.81 & 24.40 & 38.73 & 20.69 & 32.30 & 37.32 & 30.00 & 23.30 & 33.68 & 22.83 & 26.67\\
\AppMGemini & 26.91 & 21.29 & 25.60 & 34.49 & 22.40 & 42.40 & 38.15 & \fwsecond{32.05} & 21.90 & 33.18 & 25.25 & \fwsecond{29.97}\\
\AppMClaude & \fwsecond{28.10} & \fwbest{25.34} & 20.30 & \fwbest{42.94} & \fwbest{33.85} & \fwbest{43.50} & \fwbest{40.92} & 31.55 & 22.00 & \fwsecond{37.32} & \fwbest{30.25} & 28.60\\
\AppMQwenPlus & 27.40 & 21.04 & \fwsecond{30.50} & 40.09 & 22.73 & 33.90 & \fwsecond{39.11} & 30.03 & 18.90 & 35.53 & 24.60 & 27.77\\

}
The fire scenario task axis leads to the same conclusion from a
fire-evolution perspective. Explanation quality varies substantially across
Localized Onset, Coupled Propagation, and Critical Transition, and the model
ranking under Evi-F1 does not necessarily coincide with the ranking under
Mechanism Alignment or GLS. This divergence is important because a fire-world
model that reaches a plausible stage-specific prediction for the wrong
physical reason may behave unreliably when the environment, observation
modality, or physical state changes.

Taken together, the additional results clarify the main findings of
FireWorldBench.
First, response format matters: multiple-choice competence systematically
overstates some abilities that become substantially harder under open
reporting.
Second, structured and multimodal observations expose different model
strengths, and strong performance on sensor records does not guarantee robust
reasoning from physics-rendered visual evidence.
Third, both the physical capability axis and the fire scenario task axis
reveal pronounced internal heterogeneity that is obscured by a single
aggregate score.
Finally, open-report diagnostics show that answer correctness, evidence
recovery, causal mechanism understanding, and evidence-grounded support are
related but distinct dimensions of physical-world intelligence.
These observations motivate evaluating fire-world models through the complete
two-axis, multi-interface protocol rather than through final-answer accuracy
alone.
\subsection{Environment-family analysis}

Beyond the capability- and task-level decompositions, we further examine
whether open-report performance varies across the seven environment families
represented in the controlled simulation subset.
Table~\ref{tab:sim_environment_family_results} reports the six evaluation
metrics after grouping the results by their source environment family.
This analysis provides a complementary view of FireWorldBench by revealing
whether answer quality, calibration, and physical explanation remain stable
across different classes of fire environments.

\begin{table*}[t]
\centering
\caption{\textbf{Open-report performance across environment families on the
controlled simulation subset.}
Results are aggregated separately over the seven environment families defined
in FireWorldBench.
Acc., F1, Evi-F1, Mech, and GLS are reported as percentages and are
higher-is-better, whereas Brier is reported on the $[0,1]$ scale and
lower is better. The best result in each metric is shown in bold.}
\label{tab:sim_environment_family_results}
\vspace{2pt}

\small
\setlength{\tabcolsep}{6.0pt}
\renewcommand{\arraystretch}{1.12}

\begin{tabular}{lcccccc}
\toprule
\textbf{Environment Family}
& \textbf{Acc.}
& \textbf{F1}
& \textbf{Brier}
& \textbf{Evi-F1}
& \textbf{Mech}
& \textbf{GLS} \\
\midrule

Transportation and Infrastructure
& 41.6
& \textbf{51.0}
& 0.42
& 36.4
& \textbf{45.9}
& 24.6 \\

Office and Digital Facilities
& 41.0
& 45.8
& 0.37
& 41.6
& 23.9
& 23.7 \\

Healthcare and Education
& \textbf{46.8}
& 49.5
& 0.33
& \textbf{46.9}
& 24.9
& 28.3 \\

Commercial and Public Spaces
& 44.4
& 49.1
& 0.35
& 42.0
& 24.5
& 27.1 \\

Residential and Care Settings
& 45.5
& 48.0
& \textbf{0.32}
& 39.3
& 24.0
& \textbf{28.8} \\

Industry, Energy, and Logistics
& 40.5
& 48.1
& 0.37
& 44.1
& 24.2
& 23.6 \\

Wildland and Wildland--Urban Interface (WUI)
& 41.8
& 41.1
& 0.37
& 37.6
& 21.6
& 21.1 \\

\bottomrule
\end{tabular}
\end{table*}

The results reveal clear environment-dependent differences that are not fully
captured by answer accuracy alone.
Healthcare and Education achieves the highest Acc. (46.8) and Evi-F1 (46.9),
indicating comparatively strong answer completion and evidence recovery.
Transportation and Infrastructure obtains the highest F1 (51.0) and, more
notably, the highest Mechanism Alignment (45.9), suggesting substantially
stronger recovery of the underlying physical mechanisms.
Residential and Care Settings yields the best calibration
(Brier $=0.32$) and the highest GLS (28.8).
In contrast, Wildland and Wildland--Urban Interface (WUI) records the lowest
F1 (41.1), Mech (21.6), and GLS (21.1), while Industry, Energy, and Logistics
has the lowest Acc. (40.5).

A particularly notable observation is that cross-family variation is much
larger for mechanism-level understanding than for final-answer accuracy.
Acc. varies within a relatively narrow range of 40.5--46.8, whereas Mech
ranges from 21.6 to 45.9.
This gap suggests that models can maintain superficially similar answer-level
performance across different fire environments while differing substantially
in whether they recover the correct underlying physical mechanism.
Together with the preceding capability- and task-level results, this
environment-family analysis further supports evaluating physical-world
understanding through answer quality, calibration, evidence grounding, and
mechanism recovery jointly rather than relying on accuracy alone.

\section{FireWorldGPT Additional Results}
\label{app:fireworldgpt}

We provide the complete controlled-simulation results of the FireWorldGPT
method study.
While the main paper reports the equal average of the multiple-choice and
open-report interfaces, we separate the two response formats here to expose
their interface-specific behavior.
All methods are evaluated on the same frozen test instances as their
corresponding Vanilla configuration.
The Vanilla rows in Tables~\ref{tab:fireworldgpt_choice} and
\ref{tab:fireworldgpt_open} serve as the paired reference runs for the
FireWorldGPT intervention study rather than as reproductions of the
benchmark-wide baseline summaries in Appendix~\ref{app:additional_results}.
Accordingly, all reported method differences are interpreted relative to the
matched Vanilla run within the same intervention-study configuration.
For Prompt, CoT, and SFT, the colored annotation following each absolute score
reports its signed difference from Vanilla.
Green upward arrows denote numerical increases and red downward arrows denote
numerical decreases.
Acc. and F1 are reported in percentage points, whereas Brier and its change
are reported directly on the $[0,1]$ scale.
Higher Acc. and F1 are better, while lower Brier is better.

\subsection{Multiple-Choice Results}
\label{app:fireworldgpt_choice}

\begin{table*}[t]
\centering
\caption{\textbf{FireWorldGPT results under multiple-choice evaluation.}
Rows compare Vanilla, physics-aware prompting, chain-of-thought reasoning, and
supervised fine-tuning across four foundation models.
Colored annotations report changes relative to Vanilla:
green denotes performance improvement and red denotes degradation, while arrows
indicate the numerical direction of change.
Acc. and F1 are higher-is-better, whereas lower Brier is better.}
\label{tab:fireworldgpt_choice}
\vspace{2pt}

\scriptsize
\setlength{\tabcolsep}{3.2pt}
\renewcommand{\arraystretch}{1.12}

\resizebox{\textwidth}{!}{%
\begin{tabular}{l*{12}{c}}
\toprule
\multirow{2}{*}{\textbf{Method}}
& \multicolumn{3}{c}{\textbf{Llama-3.1-8B-Instruct}}
& \multicolumn{3}{c}{\textbf{Qwen3-8B}}
& \multicolumn{3}{c}{\textbf{InternVL3-8B}}
& \multicolumn{3}{c}{\textbf{Qwen3-VL-8B}} \\
\cmidrule(lr){2-4}
\cmidrule(lr){5-7}
\cmidrule(lr){8-10}
\cmidrule(lr){11-13}
& Acc & F1 & Brier
& Acc & F1 & Brier
& Acc & F1 & Brier
& Acc & F1 & Brier \\
\midrule

\textbf{Vanilla}
& 24.4 & 34.9 & 0.90
& 30.5 & 40.9 & 0.86
& 22.9 & 31.8 & 0.52
& 28.2 & 37.5 & 0.84 \\

\textbf{+Prompt}
& 28.2 {\scriptsize\textcolor{green!50!black}{$\uparrow$3.8}}
& 39.0 {\scriptsize\textcolor{green!50!black}{$\uparrow$4.1}}
& 0.97 {\scriptsize\textcolor{red}{$\uparrow$0.07}}
& 31.3 {\scriptsize\textcolor{green!50!black}{$\uparrow$0.8}}
& 41.3 {\scriptsize\textcolor{green!50!black}{$\uparrow$0.4}}
& 0.89 {\scriptsize\textcolor{red}{$\uparrow$0.03}}
& 23.8 {\scriptsize\textcolor{green!50!black}{$\uparrow$0.9}}
& 34.1 {\scriptsize\textcolor{green!50!black}{$\uparrow$2.3}}
& 0.48 {\scriptsize\textcolor{green!50!black}{$\downarrow$0.04}}
& 33.3 {\scriptsize\textcolor{green!50!black}{$\uparrow$5.1}}
& 43.1 {\scriptsize\textcolor{green!50!black}{$\uparrow$5.6}}
& 0.89 {\scriptsize\textcolor{red}{$\uparrow$0.05}} \\

\textbf{+CoT}
& 24.0 {\scriptsize\textcolor{red}{$\downarrow$0.4}}
& 34.2 {\scriptsize\textcolor{red}{$\downarrow$0.7}}
& \textbf{0.67} {\scriptsize\textcolor{green!50!black}{$\downarrow$0.23}}
& 32.7 {\scriptsize\textcolor{green!50!black}{$\uparrow$2.2}}
& 43.2 {\scriptsize\textcolor{green!50!black}{$\uparrow$2.3}}
& 0.82 {\scriptsize\textcolor{green!50!black}{$\downarrow$0.04}}
& 26.4 {\scriptsize\textcolor{green!50!black}{$\uparrow$3.5}}
& 35.2 {\scriptsize\textcolor{green!50!black}{$\uparrow$3.4}}
& \textbf{0.21} {\scriptsize\textcolor{green!50!black}{$\downarrow$0.31}}
& 30.3 {\scriptsize\textcolor{green!50!black}{$\uparrow$2.1}}
& 40.5 {\scriptsize\textcolor{green!50!black}{$\uparrow$3.0}}
& 0.87 {\scriptsize\textcolor{red}{$\uparrow$0.03}} \\

\textbf{+SFT}
& \textbf{42.6} {\scriptsize\textcolor{green!50!black}{$\uparrow$18.2}}
& \textbf{55.9} {\scriptsize\textcolor{green!50!black}{$\uparrow$21.0}}
& 0.95 {\scriptsize\textcolor{red}{$\uparrow$0.05}}
& \textbf{42.6} {\scriptsize\textcolor{green!50!black}{$\uparrow$12.1}}
& \textbf{52.2} {\scriptsize\textcolor{green!50!black}{$\uparrow$11.3}}
& \textbf{0.81} {\scriptsize\textcolor{green!50!black}{$\downarrow$0.05}}
& \textbf{40.2} {\scriptsize\textcolor{green!50!black}{$\uparrow$17.3}}
& \textbf{50.4} {\scriptsize\textcolor{green!50!black}{$\uparrow$18.6}}
& 0.87 {\scriptsize\textcolor{red}{$\uparrow$0.35}}
& \textbf{41.3} {\scriptsize\textcolor{green!50!black}{$\uparrow$13.1}}
& \textbf{51.2} {\scriptsize\textcolor{green!50!black}{$\uparrow$13.7}}
& \textbf{0.83} {\scriptsize\textcolor{green!50!black}{$\downarrow$0.01}} \\

\bottomrule
\end{tabular}}
\end{table*}

Multiple-choice evaluation shows a consistent advantage for SFT in answer
quality across all four backbones.
For Llama-3.1-8B-Instruct, SFT raises Acc. from 24.4 to 42.6 and F1 from
34.9 to 55.9.
InternVL3-8B exhibits similarly large gains, increasing from 22.9 to 40.2
in Acc. and from 31.8 to 50.4 in F1.
Qwen3-VL-8B follows the same trend, with SFT improving Acc. from 28.2 to
41.3 and F1 from 37.5 to 51.2.
Thus, under candidate-conditioned evaluation, supervised adaptation produces
the strongest answer-quality gains for every evaluated backbone.

Confidence calibration exhibits a less uniform pattern.
CoT achieves the lowest Brier for Llama-3.1-8B-Instruct and InternVL3-8B,
reducing the score from 0.90 to 0.67 and from 0.52 to 0.21, respectively.
SFT instead achieves the lowest Brier for Qwen3-8B and Qwen3-VL-8B, at
0.81 and 0.83.
In contrast, the strong Acc./F1 gains of InternVL3-8B under SFT are
accompanied by a substantial Brier increase from 0.52 to 0.87.
These results indicate that improvements in candidate-conditioned answer
quality do not necessarily translate into uniformly better calibration.

\subsection{Open-Report Results}
\label{app:fireworldgpt_open}

\begin{table*}[t]
\centering
\caption{\textbf{FireWorldGPT results under open-report evaluation.}
Rows compare Vanilla, physics-aware prompting, chain-of-thought reasoning, and
supervised fine-tuning across four foundation models.
Colored annotations report changes relative to Vanilla:
green denotes performance improvement and red denotes degradation, while arrows
indicate the numerical direction of change.
Acc. and F1 are higher-is-better, whereas lower Brier is better.}
\label{tab:fireworldgpt_open}
\vspace{2pt}

\scriptsize
\setlength{\tabcolsep}{3.2pt}
\renewcommand{\arraystretch}{1.12}

\resizebox{\textwidth}{!}{%
\begin{tabular}{l*{12}{c}}
\toprule
\multirow{2}{*}{\textbf{Method}}
& \multicolumn{3}{c}{\textbf{Llama-3.1-8B-Instruct}}
& \multicolumn{3}{c}{\textbf{Qwen3-8B}}
& \multicolumn{3}{c}{\textbf{InternVL3-8B}}
& \multicolumn{3}{c}{\textbf{Qwen3-VL-8B}} \\
\cmidrule(lr){2-4}
\cmidrule(lr){5-7}
\cmidrule(lr){8-10}
\cmidrule(lr){11-13}
& Acc & F1 & Brier
& Acc & F1 & Brier
& Acc & F1 & Brier
& Acc & F1 & Brier \\
\midrule

\textbf{Vanilla}
& 37.8 & 30.8 & 0.63
& 36.2 & 30.5 & \textbf{0.74}
& 27.2 & 21.0 & 0.65
& 30.2 & 21.5 & 0.74 \\

\textbf{+Prompt}
& 38.5 {\scriptsize\textcolor{green!50!black}{$\uparrow$0.7}}
& 29.9 {\scriptsize\textcolor{red}{$\downarrow$0.9}}
& \textbf{0.62} {\scriptsize\textcolor{green!50!black}{$\downarrow$0.01}}
& \textbf{39.7} {\scriptsize\textcolor{green!50!black}{$\uparrow$3.5}}
& \textbf{32.7} {\scriptsize\textcolor{green!50!black}{$\uparrow$2.2}}
& 0.75 {\scriptsize\textcolor{red}{$\uparrow$0.01}}
& \textbf{35.0} {\scriptsize\textcolor{green!50!black}{$\uparrow$7.8}}
& 24.7 {\scriptsize\textcolor{green!50!black}{$\uparrow$3.7}}
& \textbf{0.64} {\scriptsize\textcolor{green!50!black}{$\downarrow$0.01}}
& 35.7 {\scriptsize\textcolor{green!50!black}{$\uparrow$5.5}}
& 24.2 {\scriptsize\textcolor{green!50!black}{$\uparrow$2.7}}
& \textbf{0.66} {\scriptsize\textcolor{green!50!black}{$\downarrow$0.08}} \\

\textbf{+CoT}
& 40.3 {\scriptsize\textcolor{green!50!black}{$\uparrow$2.5}}
& 32.9 {\scriptsize\textcolor{green!50!black}{$\uparrow$2.1}}
& 0.72 {\scriptsize\textcolor{red}{$\uparrow$0.09}}
& 38.6 {\scriptsize\textcolor{green!50!black}{$\uparrow$2.4}}
& 30.9 {\scriptsize\textcolor{green!50!black}{$\uparrow$0.4}}
& 0.75 {\scriptsize\textcolor{red}{$\uparrow$0.01}}
& 29.9 {\scriptsize\textcolor{green!50!black}{$\uparrow$2.7}}
& \textbf{24.8} {\scriptsize\textcolor{green!50!black}{$\uparrow$3.8}}
& 0.69 {\scriptsize\textcolor{red}{$\uparrow$0.04}}
& 33.2 {\scriptsize\textcolor{green!50!black}{$\uparrow$3.0}}
& 27.2 {\scriptsize\textcolor{green!50!black}{$\uparrow$5.7}}
& 0.76 {\scriptsize\textcolor{red}{$\uparrow$0.02}} \\

\textbf{+SFT}
& \textbf{43.5} {\scriptsize\textcolor{green!50!black}{$\uparrow$5.7}}
& \textbf{33.2} {\scriptsize\textcolor{green!50!black}{$\uparrow$2.4}}
& 0.83 {\scriptsize\textcolor{red}{$\uparrow$0.20}}
& 38.9 {\scriptsize\textcolor{green!50!black}{$\uparrow$2.7}}
& 29.4 {\scriptsize\textcolor{red}{$\downarrow$1.1}}
& 0.87 {\scriptsize\textcolor{red}{$\uparrow$0.13}}
& 28.0 {\scriptsize\textcolor{green!50!black}{$\uparrow$0.8}}
& 24.2 {\scriptsize\textcolor{green!50!black}{$\uparrow$3.2}}
& 0.86 {\scriptsize\textcolor{red}{$\uparrow$0.21}}
& \textbf{40.6} {\scriptsize\textcolor{green!50!black}{$\uparrow$10.4}}
& \textbf{27.4} {\scriptsize\textcolor{green!50!black}{$\uparrow$5.9}}
& 0.81 {\scriptsize\textcolor{red}{$\uparrow$0.07}} \\

\bottomrule
\end{tabular}}
\end{table*}

Open-report evaluation reveals a more heterogeneous interaction between
adaptation strategy and model backbone.
SFT remains highly effective for several models even after candidate answers
are removed.
For Llama-3.1-8B-Instruct, SFT improves Acc. from 37.8 to 43.5 and F1 from
30.8 to 33.2.
The strongest open-report SFT effect appears on Qwen3-VL-8B, where Acc.
increases from 30.2 to 40.6 and F1 from 21.5 to 27.4.
These gains show that parameter-level adaptation can improve open-ended
physical-state reconstruction rather than only candidate-conditioned answer
selection.

Prompting is more competitive for the other backbones.
It achieves the highest Acc. and F1 for Qwen3-8B, and the highest Acc. for
InternVL3-8B, while CoT obtains the highest InternVL3-8B F1.
Calibration again behaves differently from answer quality.
Prompting produces the lowest Brier for Llama-3.1-8B-Instruct,
InternVL3-8B, and Qwen3-VL-8B, including a reduction from 0.74 to 0.66 for
Qwen3-VL-8B.
For Qwen3-8B, the Vanilla configuration retains the lowest Brier at 0.74.
Overall, no single intervention dominates every open-report metric:
supervised adaptation provides the strongest open-ended gains for some
backbones, whereas prompting and CoT remain competitive on others, and
confidence calibration can move independently of answer accuracy.
\section{FireWorldGPT Implementation Details}
\label{app:fireworldgpt_impl}

\subsection{Controlled Enhancement Protocol}

The FireWorldGPT study compares three complementary enhancement strategies against a shared Vanilla baseline: physics-aware prompting, chain-of-thought reasoning, and supervised fine-tuning. Vanilla uses the standard benchmark instruction together with the task-admissible observations and the prescribed response schema. The three enhancements are evaluated as mutually exclusive conditions rather than being composed: \(+\)Prompt modifies the presentation and organization of the available physical evidence, \(+\)CoT modifies the inference process while keeping the evidence fixed, and \(+\)SFT updates the model parameters using training-split supervision. FireWorldGPT specifically denotes the model obtained through the \(+\)SFT condition.

Within each model comparison, the benchmark split, logical items, observation track, temporal cutoff, response schema, decoding configuration, and deterministic evaluator are held fixed. The methods therefore differ only in the enhancement being tested. Prompting and CoT do not update the model parameters, whereas SFT uses no test-time reasoning scaffold beyond the standard Vanilla task instruction. No condition is allowed to access hidden simulator states, post-cutoff observations, Gold evidence, or fields excluded by the task-specific observation contract.

\subsection{Physics-Aware Prompting}

Physics-aware prompting provides an explicit protocol for reading task-admissible fire-world evidence. The prompt directs the model to identify the relevant physical variable, spatial region, observation time, legend or measurement scale, and task-specific output fields before producing its prediction. For the structured-observation track, this protocol is applied to the supplied sensor traces and physical records. For the multimodal observation track, the input is supplemented with available observation-aligned visualizations of temperature, soot density, visibility, or velocity, restricted to timestamps permitted by the task cutoff. The model is explicitly instructed not to infer numerical values that cannot be reliably resolved from the input.

The augmentation is constructed only from public, task-admissible observations. It does not introduce derived evidence tables, region rankings, risk summaries, propagation annotations, answer-conditioned statistics, hidden thresholds, or future frames. Each added visualization retains its source world, physical variable, timestamp, observation cutoff, rendering configuration, and artifact hash. If no nonredundant and admissible field representation can be constructed for an item, the augmentation reduces to a no-op and the item retains its Vanilla evidence. Consequently, \(+\)Prompt changes neither the physical target nor the model parameters; it tests whether an explicit field-reading interface is sufficient to improve physical grounding.

\subsection{Chain-of-Thought Reasoning}

The \(+\)CoT condition uses exactly the same observations as Vanilla and adds a single instruction requesting an explicit four-stage inference process~\citep{wei2022chain}. The model is asked to first identify the relevant observations, then infer the latent physical state, connect that state to a coupled-field mechanism, and finally derive the task-specific consequence. This ordering follows the observation--state--mechanism--consequence formulation introduced in Section~\ref{sec:chain_of_thought} and is applied consistently to grounding, forecasting, diagnosis, and intervention questions.

The complete intermediate rationale is retained for qualitative auditing, but only the final answer is scored. A deterministic extractor projects the final JSON object into the same choice-format or open-report schema used by Vanilla. The method does not employ an external physical rule base, retrieval system, simulator call, verifier, correction stage, self-consistency voting, multi-turn refinement, or LLM judge. It therefore measures the effect of exposing a model's internal inference path while holding its parameters and observable evidence fixed.

\subsection{Supervised Fine-Tuning}

The \(+\)SFT condition adapts the base model using publicly available examples from the controlled-simulation training split of FireWorldBench~\citep{ouyang2022instructgpt}. Training examples are deduplicated at the logical-item level before serialization, ensuring that paired response formats or repeated renderings do not unintentionally multiply the supervision associated with a single physical target. All temporal variants and related instances from the same source event remain subject to the event-group split, and no evaluation or real-world-aligned test item is used for optimization.

Each training input contains the standard task instruction, question, and task-admissible observations from the corresponding observation track. Choice-format targets contain the correct option set. Open-report targets contain the structured \texttt{prediction}, a concise \texttt{conclusion}, Gold-linked \texttt{evidence}, and the corresponding physical \texttt{mechanism}. Visual-model training uses the authorized multimodal observations associated with the item, whereas the structured-observation counterpart uses only the corresponding textual sensor and physical records. Evidence is never transferred across tracks when it is unavailable under the original task contract.

Let \(\mathcal{D}_{\mathrm{train}}\) denote the resulting training set, \(x\) an input sequence, and \(y=(y_1,\ldots,y_{|y|})\) its serialized target. The model is optimized using the standard autoregressive next-token objective:

\begin{equation}
\mathcal{L}_{\mathrm{SFT}}(\theta)
=
-\sum_{(x,y)\in\mathcal{D}_{\mathrm{train}}}
\sum_{t=1}^{|y|}
m_t
\log p_{\theta}(y_t\mid x,y_{<t}).
\label{eq:fireworldgpt_sft}
\end{equation}

Here, \(\theta\) denotes the trainable model parameters and \(m_t\) is a token-level supervision mask. Because the released training examples do not provide authoritative confidence annotations, tokens corresponding to the confidence value are excluded from the supervised objective:

\begin{equation}
m_t
=
\begin{cases}
0, & t\in\mathcal{T}_{\mathrm{conf}},\\
1, & t\notin\mathcal{T}_{\mathrm{conf}},
\end{cases}
\label{eq:fireworldgpt_confidence_mask}
\end{equation}

where \(\mathcal{T}_{\mathrm{conf}}\) is the set of target positions occupied by the confidence value. The remaining response structure is fully supervised. We introduce no auxiliary classification objective, contrastive loss, pairwise preference loss, reward model, synthetic negative examples, or additional self-training stage. Thus, improvements under \(+\)SFT can be attributed to learning the mapping from admissible observations to structured physical predictions, evidence, and mechanisms rather than to an auxiliary training signal.

At inference time, FireWorldGPT receives the same standard task prompt and response schema as Vanilla. Confidence is generated by the adapted model but is not learned from an authoritative confidence target. Brier scores can therefore still be computed by the common evaluator, but they should not be interpreted as evidence that SFT directly optimizes calibration.

\subsection{Paired Evaluation and Reproducibility}

All four conditions---Vanilla, \(+\)Prompt, \(+\)CoT, and \(+\)SFT---are evaluated on matched test instances. For a metric \(M\), method \(a\), and a fixed evaluation cell \(\mathcal{D}_k\), the reported improvement is computed relative to the corresponding Vanilla run:

\begin{equation}
\Delta M_{a,k}
=
M(f_a;\mathcal{D}_k)
-
M(f_{\mathrm{Vanilla}};\mathcal{D}_k).
\label{eq:fireworldgpt_paired_delta}
\end{equation}

Here, \(f_a\) denotes the prediction function induced by enhancement \(a\), while \(\mathcal{D}_k\) fixes the model, benchmark subset, capability group, observation track, and response interface. Deltas are computed from the unrounded scores before being displayed in Tables~\ref{tab:fireworldgpt_choice} and~\ref{tab:fireworldgpt_open}. This paired construction prevents an apparent improvement from being caused by a different test subset, missing predictions, or a changed observation policy.

Every run records the base checkpoint revision, enhancement condition, benchmark split, prompt and template hashes, observation configuration, temporal cutoff policy, decoding parameters, parser version, and evaluator version. Prompt runs additionally record the provenance of every attached field visualization; CoT runs retain the unmodified rationale and the deterministically extracted final response; SFT runs record the training-split manifest and target-serialization version. All predictions are scored using the deterministic protocol in Appendix~\ref{app:protocol}, and invalid responses remain in the applicable metric denominators. These controls make the three enhancement strategies directly comparable while preserving the distinction between input-level evidence structuring, inference-time reasoning, and parameter-level domain adaptation.
\section{Qualitative Benchmark Cases and Multimodal Fire-World Visualizations}
\label{app:qualitative_examples}

\begin{figure}[p]
    \centering
    \includegraphics[
        width=\linewidth,
        height=0.84\textheight,
        keepaspectratio
    ]{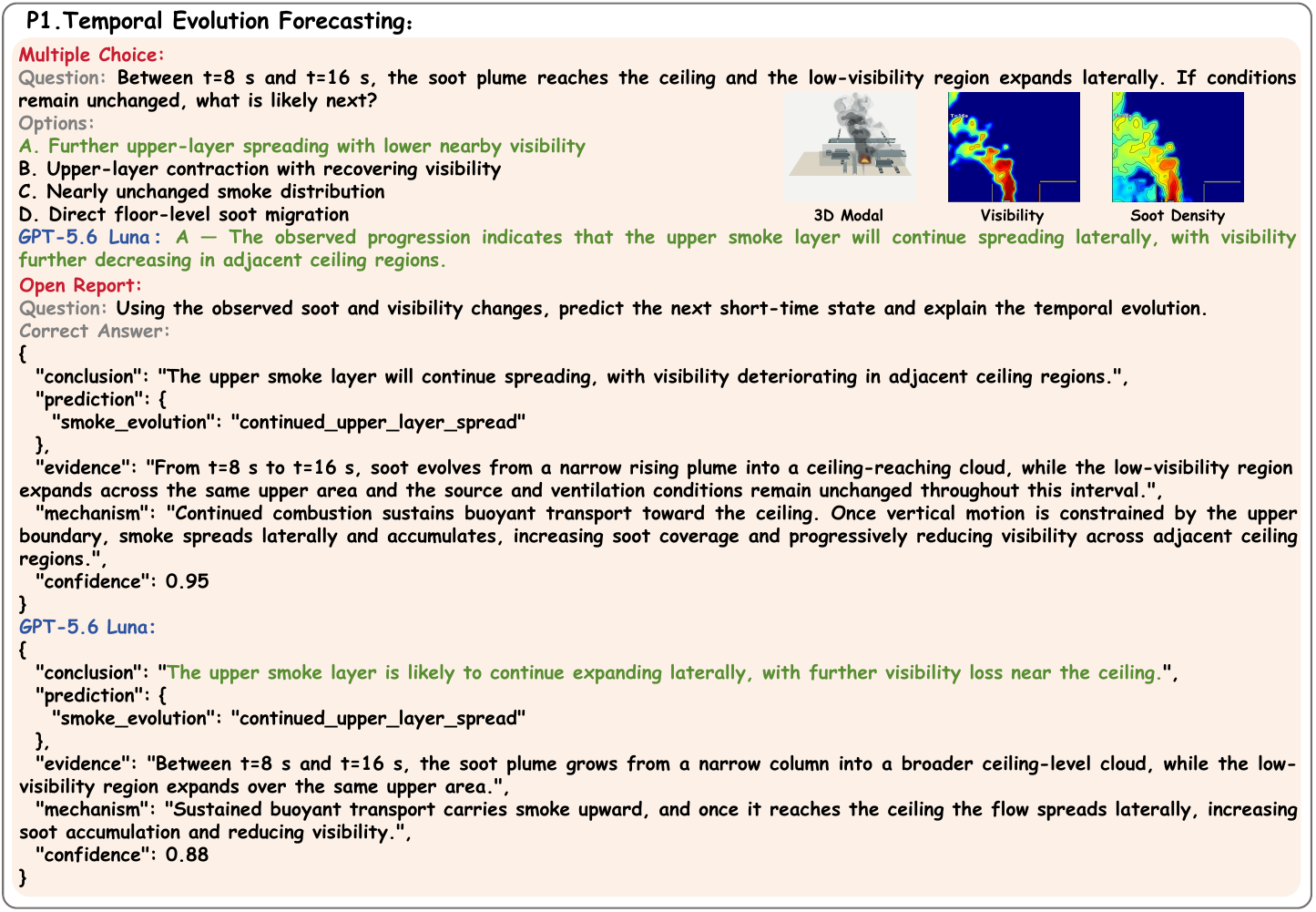}
    \caption{\textbf{P1: Temporal Evolution Forecasting.}
A matched choice-format and open-report case asks how the smoke state will
evolve after the soot plume reaches the ceiling and the low-visibility region
expands laterally from $t=8$\,s to $t=16$\,s.
The Gold response predicts continued upper-layer spreading with further
visibility degradation in adjacent ceiling regions.
The representative GPT-5.6-Luna response reaches the same forecast and
grounds it in the observed ceiling interaction, lateral soot expansion, and
progressive visibility loss.}
    \label{fig:qualitative_p1_temporal}
\end{figure}

\begin{figure}[p]
    \centering
    \includegraphics[
        width=\linewidth,
        height=0.84\textheight,
        keepaspectratio
    ]{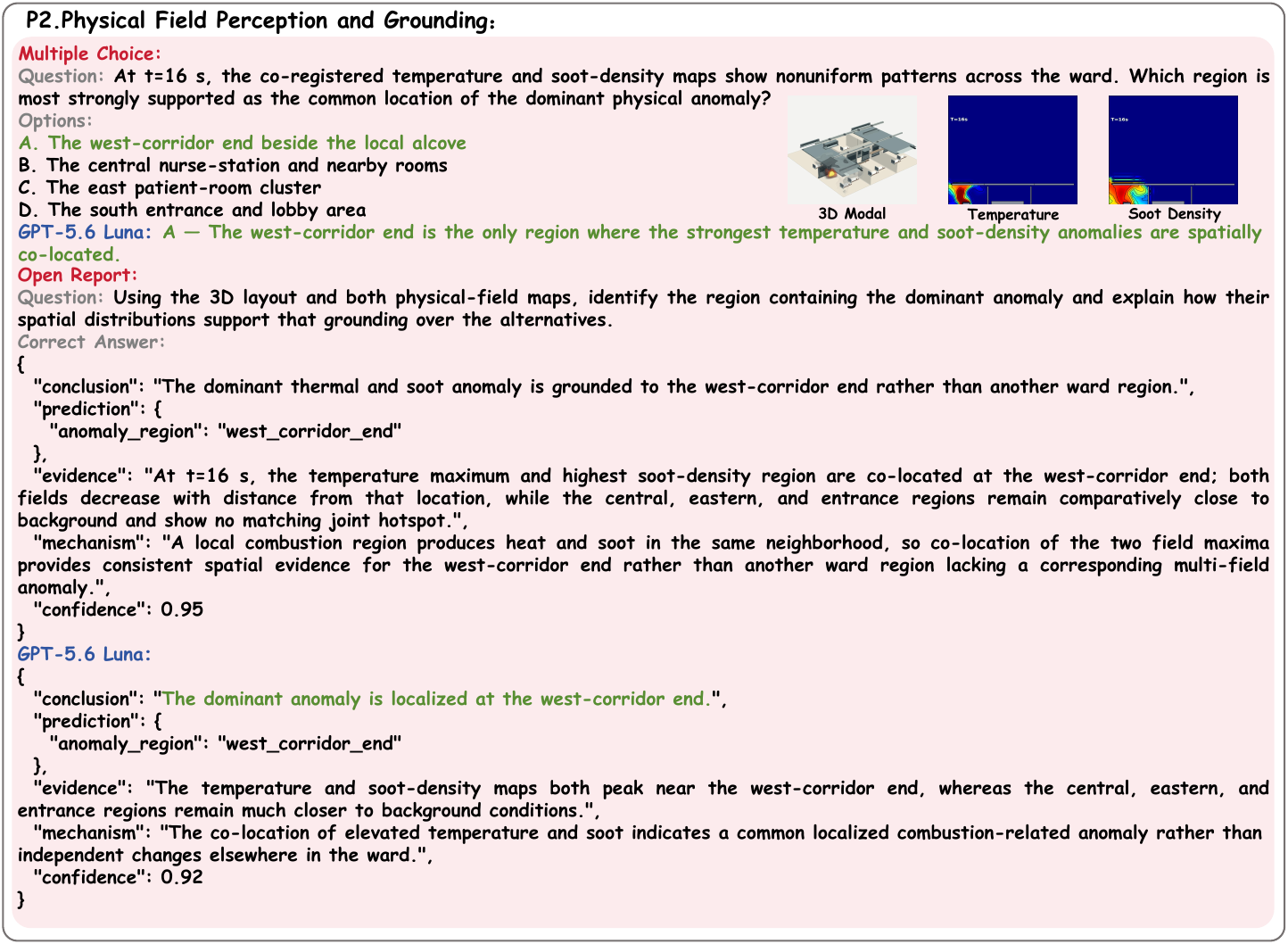}
    \caption{\textbf{P2: Physical Field Perception and Grounding.}
    A localized anomaly in a care-ward corridor is presented through a 3D scene together with temperature and soot-density fields. The task requires distinguishing a nearby linen-cart or combustible-material source from spatially broader alternatives such as ventilation or system-level failure. The representative GPT-5.6-Luna response localizes the source to the west corridor and supports this attribution using the spatial co-occurrence and rapid decay of the thermal and soot anomalies.}
    \label{fig:qualitative_p2_grounding}
\end{figure}

\begin{figure}[p]
    \centering
    \includegraphics[
        width=\linewidth,
        height=0.84\textheight,
        keepaspectratio
    ]{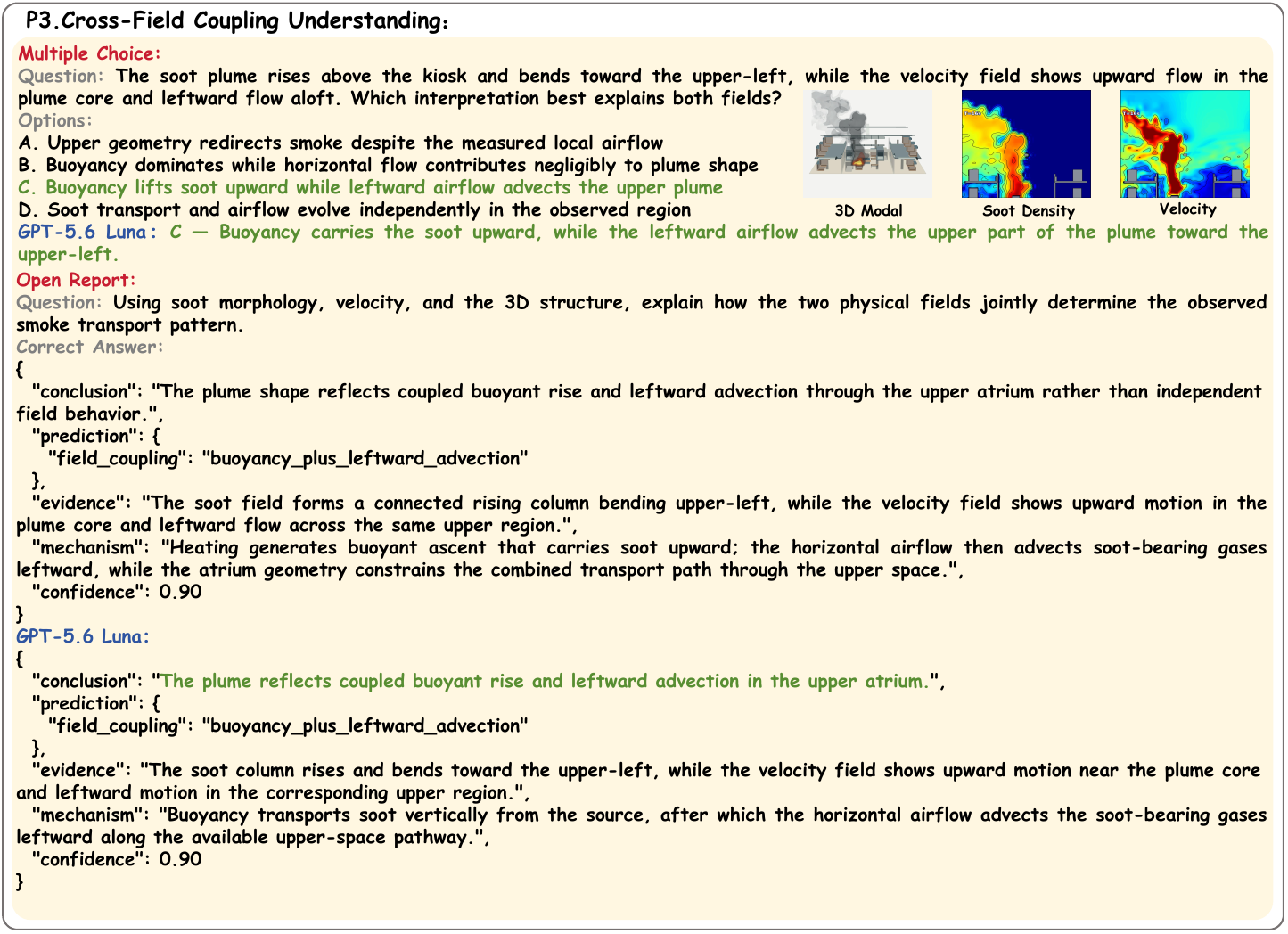}
    \caption{\textbf{P3: Cross-Field Coupling Understanding.}
    A kiosk fire in a public atrium is queried through its 3D structure, soot-density distribution, and velocity field. The task requires combining plume morphology with the flow state to forecast smoke transport under an unchanged source and an open upper exhaust route. The representative GPT-5.6-Luna response predicts continued plume rise and flow-guided lateral transport, linking buoyant ascent, ambient advection, and geometric confinement in a single coupled-field explanation.}
    \label{fig:qualitative_p3_coupling}
\end{figure}

\begin{figure}[p]
    \centering
    \includegraphics[
        width=\linewidth,
        height=0.84\textheight,
        keepaspectratio
    ]{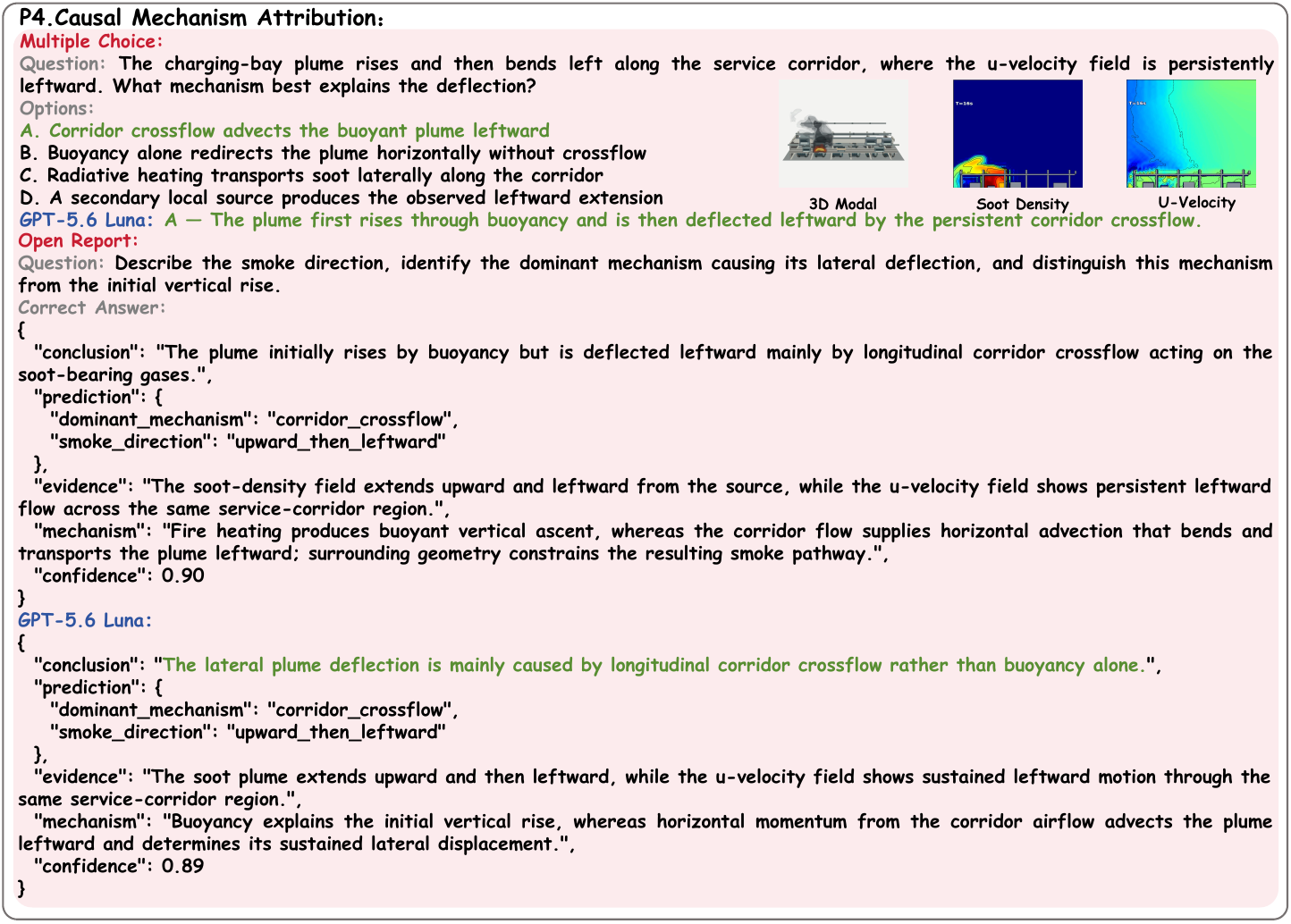}
    \caption{\textbf{P4: Causal Mechanism Attribution.}
A charging-bay fire produces a buoyant smoke plume that subsequently bends
leftward along a service corridor under a persistent leftward airflow.
The Gold response attributes the lateral deflection primarily to corridor
crossflow, while distinguishing this mechanism from the initial buoyant rise.
The representative GPT-5.6-Luna response reaches the same causal attribution,
linking the upward plume development to buoyancy and the sustained lateral
transport to corridor crossflow.}
    \label{fig:qualitative_p4_mechanism}
\end{figure}

\begin{figure}[p]
    \centering
    \includegraphics[
        width=\linewidth,
        height=0.84\textheight,
        keepaspectratio
    ]{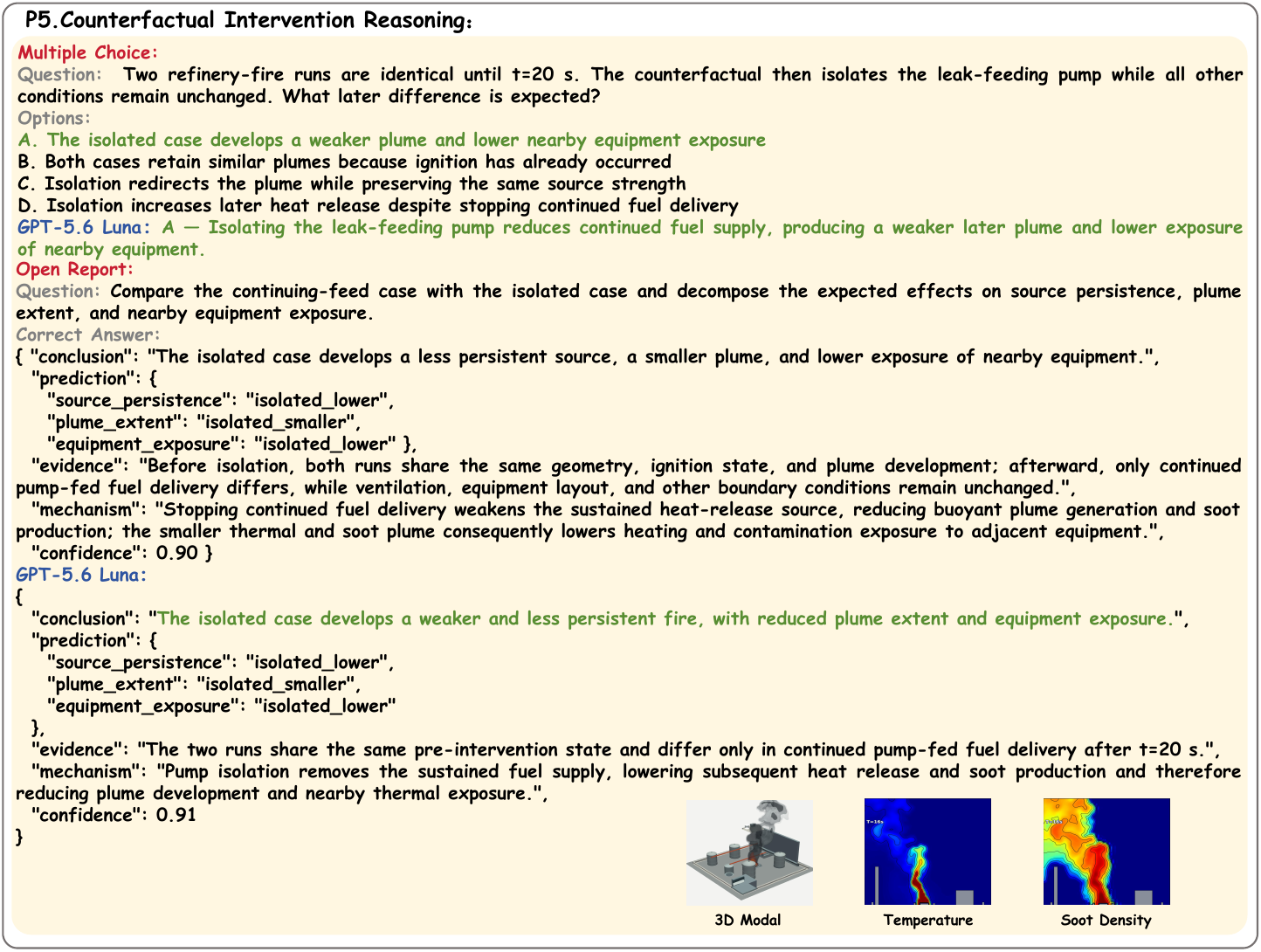}
    \caption{\textbf{P5: Counterfactual Intervention Reasoning.}
    A refinery pipe-rack fire is evaluated under the counterfactual intervention of earlier pump isolation. The task requires decomposing the intervention effect into changes in source strength, plume spread, and equipment exposure. Using the 3D scene, temperature field, and soot-density field, the representative GPT-5.6-Luna response predicts that earlier isolation would reduce continued fuel delivery, suppress subsequent plume development, and lower the thermal and contamination risks to adjacent equipment.}
    \label{fig:qualitative_p5_counterfactual}
\end{figure}

Figures~\ref{fig:qualitative_p1_temporal}--\ref{fig:qualitative_p5_counterfactual} present one representative benchmark case for each dimension of the physical capability axis, ordered from Temporal Evolution Forecasting (P1) to Counterfactual Intervention Reasoning (P5). Each case pairs a choice-format question with its matched open-report counterpart under the same physical target and multimodal evidence, and displays the Gold response alongside a representative answer from GPT-5.6-Luna. These examples illustrate how the benchmark progresses from identifying temporal order and grounding localized field observations to integrating coupled fields, attributing causal mechanisms, and predicting the consequences of interventions. They also expose the distinction between recognizing a plausible answer and producing a structured report whose conclusion is explicitly connected to physical evidence and mechanism.

\begin{figure}[p]
    \centering
    \includegraphics[
        width=0.96\linewidth,
        height=0.82\textheight,
        keepaspectratio
    ]{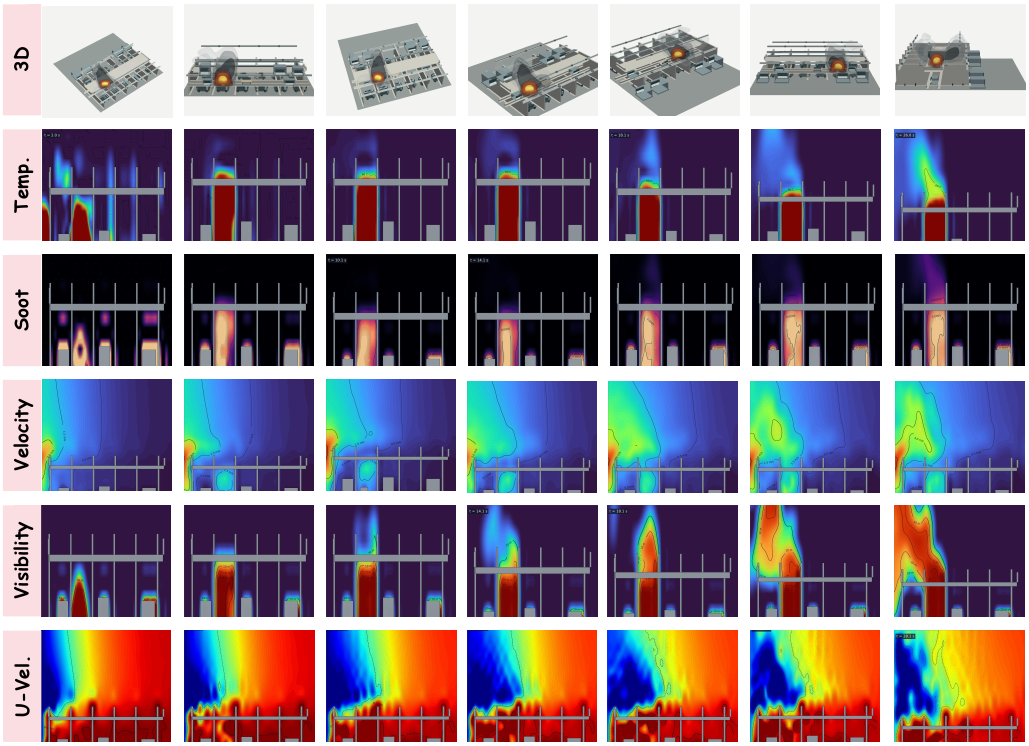}
    \caption{\textbf{Transportation and Infrastructure: Electric Bus Terminal Charging-bay Fire.}
    This controlled fire world shows the time-resolved development of a charging-bay fire through co-registered 3D scenes and temperature, soot-density, velocity, visibility, and streamwise-velocity fields. The terminal geometry makes the interaction between buoyant plume rise, bay-scale confinement, and longitudinal smoke transport directly observable.}
    \label{fig:world_transportation}
\end{figure}

\begin{figure}[p]
    \centering
    \includegraphics[
        width=0.96\linewidth,
        height=0.82\textheight,
        keepaspectratio
    ]{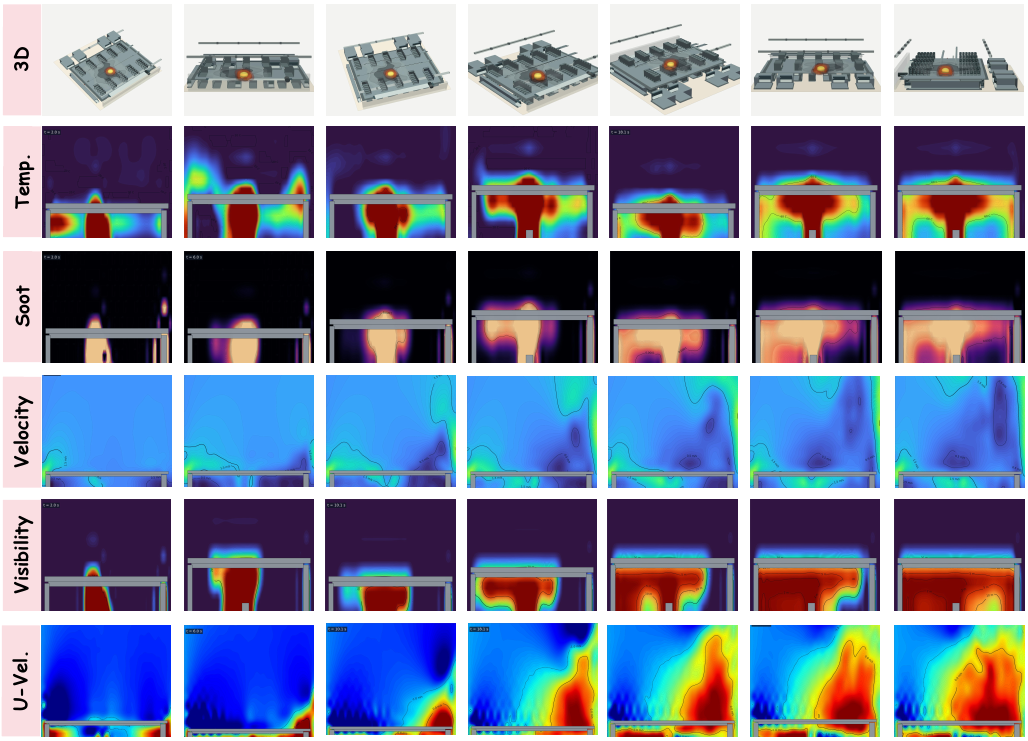}
    \caption{\textbf{Office and Digital Facilities: Data-centre Cold-aisle Cable-tray Fire.}
    This controlled sequence depicts the growth and transport of heat and smoke from a cable-tray fire within a confined cold-aisle layout. The co-registered fields reveal how overhead structures and aisle geometry constrain the thermal plume, redistribute soot, modify local flow, and progressively degrade visibility.}
    \label{fig:world_office_digital}
\end{figure}

\begin{figure}[p]
    \centering
    \includegraphics[
        width=0.96\linewidth,
        height=0.82\textheight,
        keepaspectratio
    ]{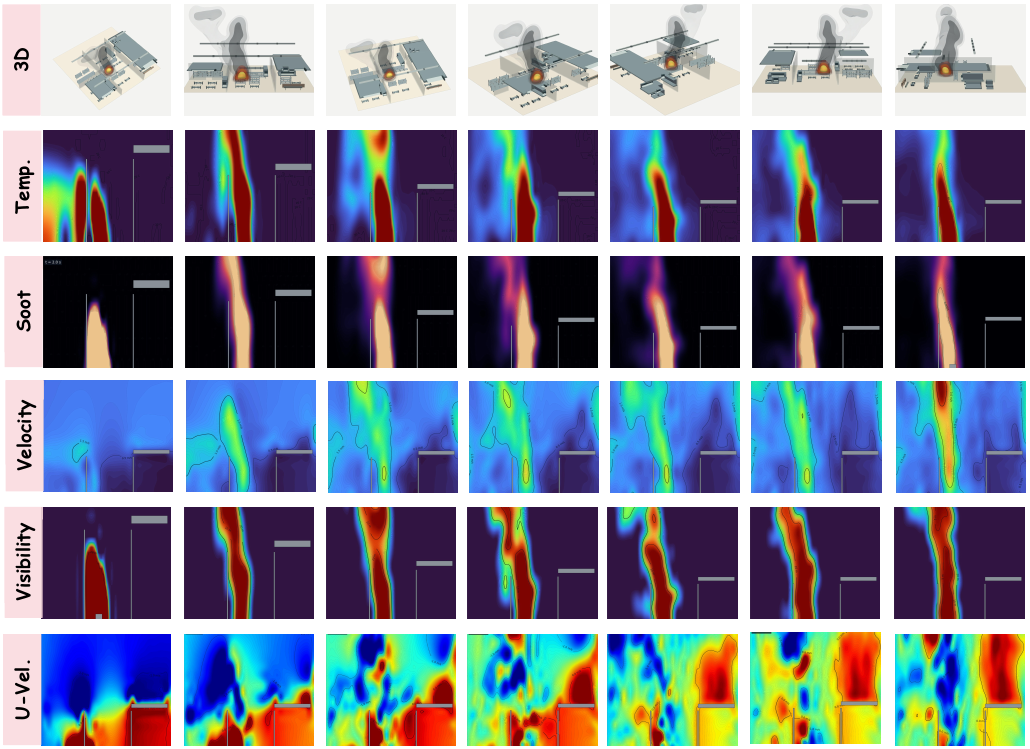}
    \caption{\textbf{Healthcare and Education: Teaching Medical Laboratory Solvent-bench Fire.}
    A localized solvent-bench source develops into a vertically extended thermal and soot plume inside a laboratory compartment. The synchronized physical fields expose the resulting changes in plume velocity, streamwise transport, and visibility, while the 3D views preserve the spatial relationship between the source, laboratory fixtures, and surrounding enclosure.}
    \label{fig:world_healthcare_education}
\end{figure}

\begin{figure}[p]
    \centering
    \includegraphics[
        width=0.96\linewidth,
        height=0.82\textheight,
        keepaspectratio
    ]{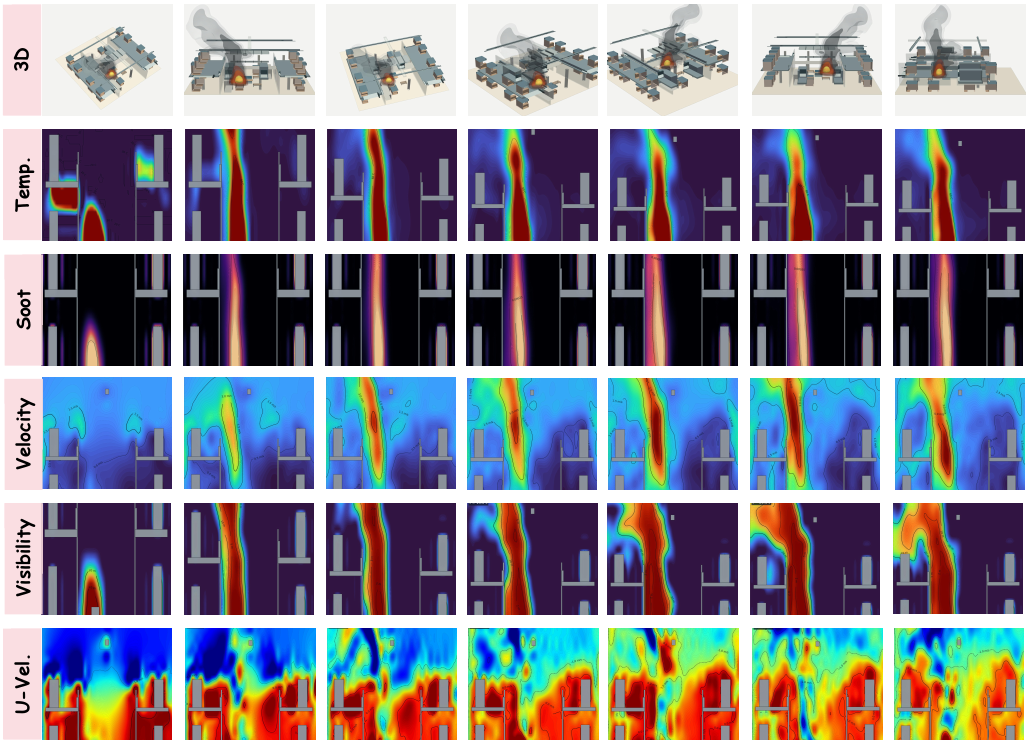}
    \caption{\textbf{Commercial and Public Spaces: Public Atrium Kiosk Fire.}
    This controlled fire world captures plume growth from a kiosk source within a tall, structurally open atrium. The temporal sequence illustrates how buoyancy, upper-level geometry, and the surrounding flow field jointly determine plume rise, lateral smoke spreading, and the spatial progression of visibility loss.}
    \label{fig:world_commercial_public}
\end{figure}

\begin{figure}[p]
    \centering
    \includegraphics[
        width=0.96\linewidth,
        height=0.82\textheight,
        keepaspectratio
    ]{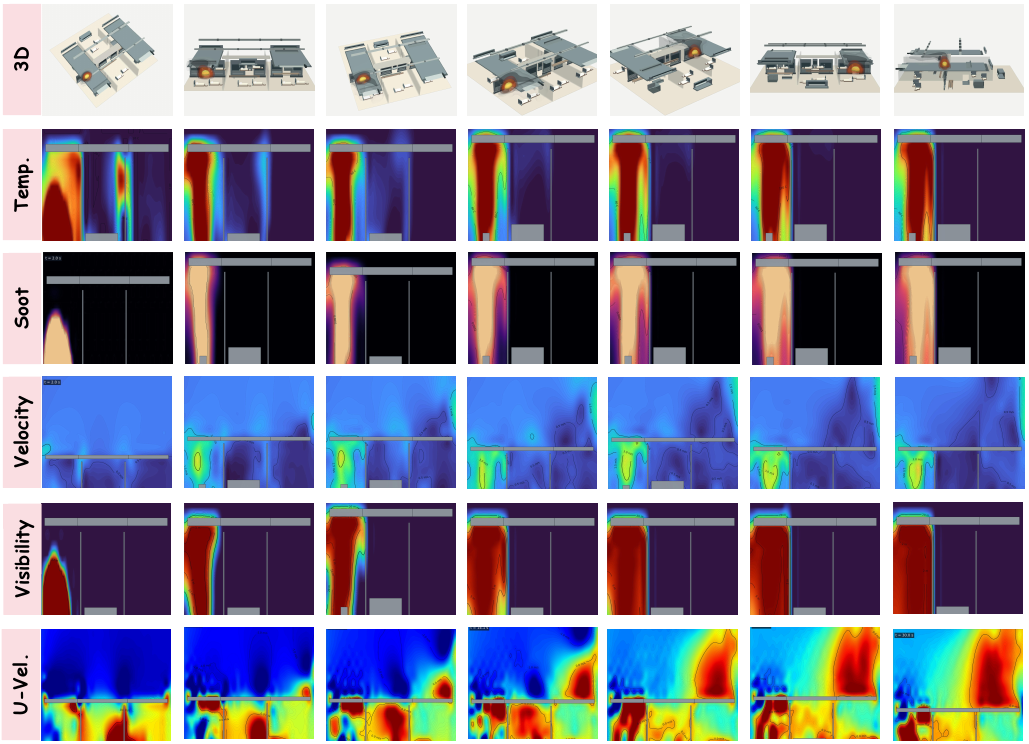}
    \caption{\textbf{Residential and Care Settings: Night-shift Care Ward Linen-cart Fire.}
    A corridor-adjacent linen-cart fire is represented through its evolving 3D scene and five co-registered physical fields. The sequence highlights the localized origin of the thermal and soot anomalies, their subsequent extension through the connected ward geometry, and the associated deterioration of visibility along the transport path.}
    \label{fig:world_residential_care}
\end{figure}

\begin{figure}[p]
    \centering
    \includegraphics[
        width=0.96\linewidth,
        height=0.82\textheight,
        keepaspectratio
    ]{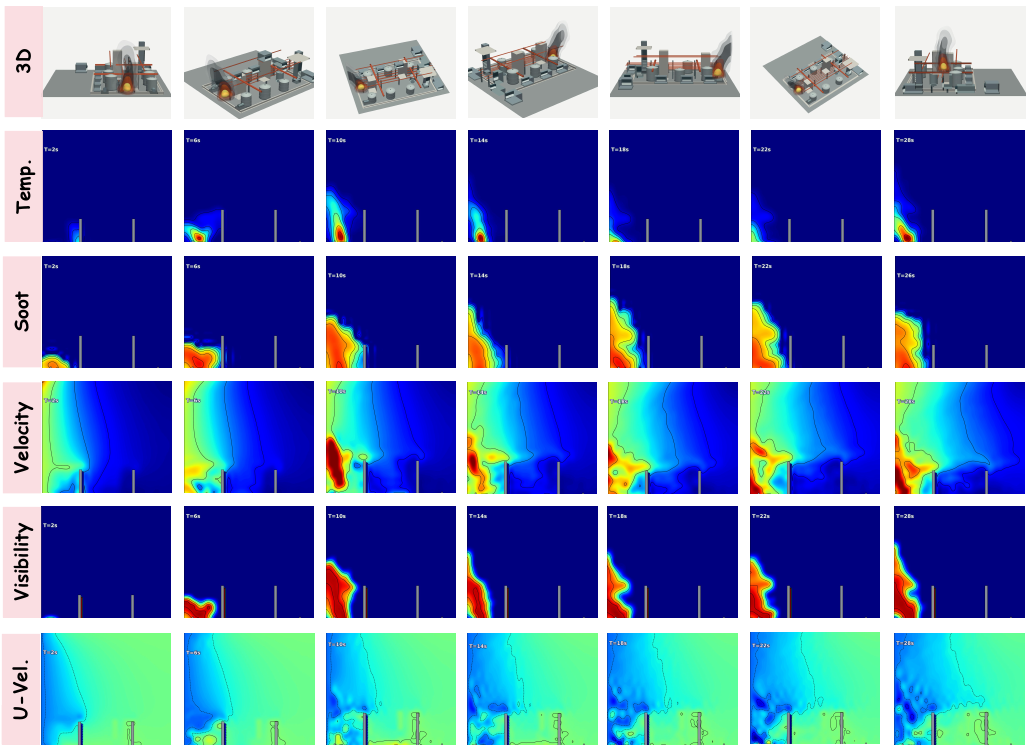}
    \caption{\textbf{Industry, Energy, and Logistics: Integrated Refinery Pipe-rack Energy Fire.}
    This controlled outdoor fire world places the source among pipe racks, processing units, and adjacent equipment. The multimodal sequence exposes the growth of the buoyant plume, the transport of heat and soot through the equipment layout, and the spatially varying velocity and visibility conditions that determine downstream exposure.}
    \label{fig:world_industry_energy}
\end{figure}

\begin{figure}[p]
    \centering
    \includegraphics[
        width=0.96\linewidth,
        height=0.82\textheight,
        keepaspectratio
    ]{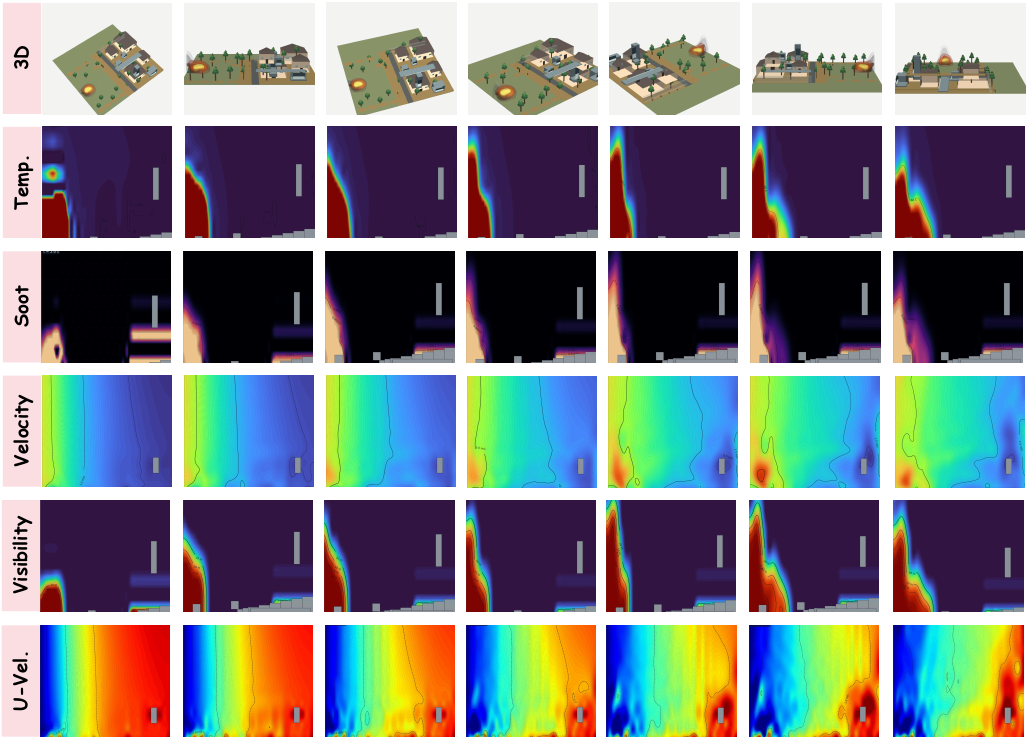}
    \caption{\textbf{Wildland and Wildland--Urban Interface: Hillside WUI Evacuation-lane Fire.}
    This controlled scene represents fire development near vegetation, structures, and a hillside evacuation route. The co-registered field sequence illustrates how terrain, built obstacles, and ambient flow shape the propagation of heat and smoke, producing spatially heterogeneous visibility and transport conditions along the wildland--urban interface.}
    \label{fig:world_wui}
\end{figure}

\begin{figure}[p]
    \centering
    \includegraphics[
        width=0.96\linewidth,
        height=0.82\textheight,
        keepaspectratio
    ]{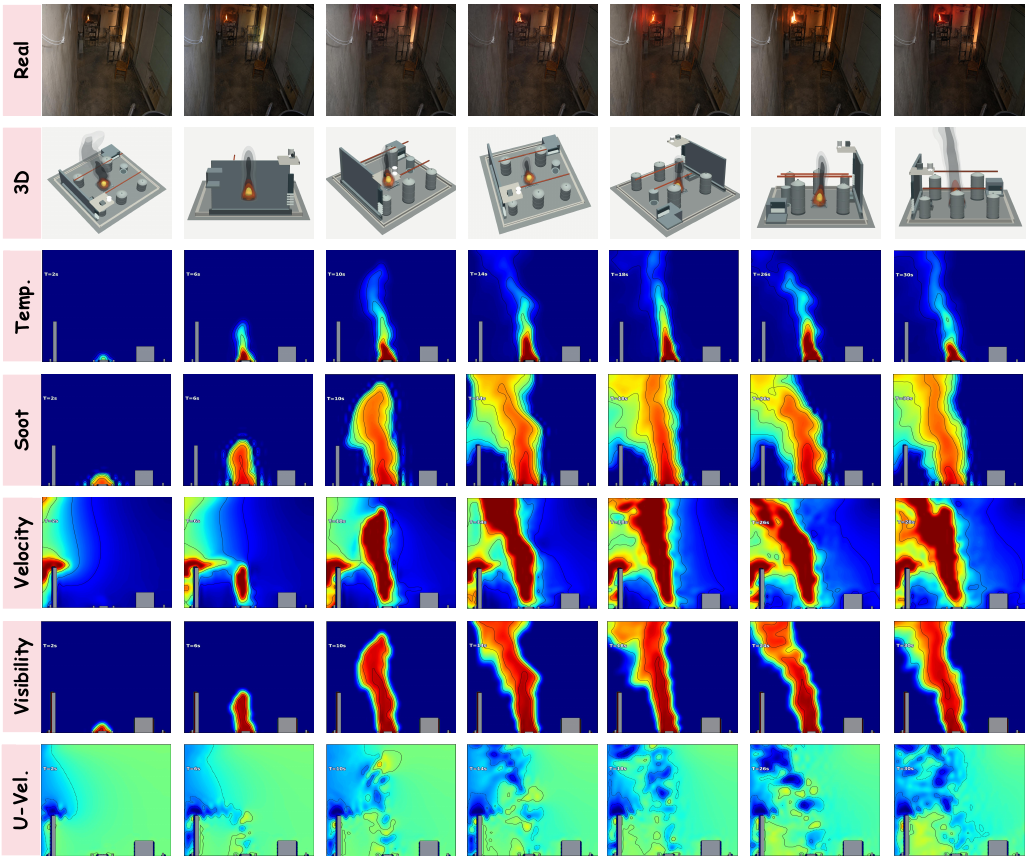}
    \caption{\textbf{Real-world-aligned event: Heptane Pool-fire Experiment.}
    Unlike the preceding controlled worlds, this example begins with an observed experimental sequence. Recorded video frames anchor the event chronology and visible fire development, while the aligned 3D reconstruction and FDS realization provide simulation-completed temperature, soot-density, velocity, visibility, and streamwise-velocity fields at matched time points. The separate rows preserve the distinction between observed evidence and reconstructed latent physical state.}
    \label{fig:world_rwa_heptane}
\end{figure}

Figures~\ref{fig:world_transportation}--\ref{fig:world_wui} complement the question-level cases with one controlled fire world from each of the seven environment families, following the ordering used in the main paper. For every world, time-indexed three-dimensional scene renderings are co-registered with temperature, soot-density, velocity, visibility, and streamwise-velocity fields, making both the semantic environment and its latent physical evolution inspectable. Figure~\ref{fig:world_rwa_heptane} then presents a real-world-aligned heptane pool-fire event: recorded video frames establish the observed chronology, whereas the aligned three-dimensional reconstruction and FDS-derived fields provide the simulation-completed latent state. The two provenance sources remain visually separated to avoid presenting reconstructed physical fields as direct measurements.

\clearpage

\clearpage

\clearpage

\clearpage

\clearpage

\clearpage

\clearpage

\clearpage

\clearpage

\clearpage

\clearpage

\clearpage

\clearpage

\clearpage
\end{document}